\pdfoutput=1

\documentclass[11pt]{article}
\usepackage{longtable} %
\usepackage{array} %
\usepackage{caption} %

\usepackage[dvipsnames]{xcolor}
\usepackage{authblk}
\usepackage{graphicx}
\usepackage{xurl}
\usepackage{enumitem}
\usepackage{xspace}
\usepackage{xparse}
\usepackage{float}

\DeclareDocumentCommand\newstep{o}{%
\item\IfNoValueTF{#1}{}{#1 \textendash\xspace}}
\newlist{steps}{enumerate}{1}
\setlist[steps]{label=\textit{RQ \arabic*:},leftmargin=*}

\usepackage[final]{acl}

\usepackage{times}
\usepackage{latexsym}
\usepackage{colortbl}
\usepackage{amssymb}
\usepackage{tablefootnote}
\usepackage{amsmath}
\usepackage{hyperref}
\usepackage{cleveref}
\crefrangelabelformat{section}{#3#1#4--#5\crefstripprefix{#1}{#2}#6}
\usepackage{subcaption}

\usepackage{ragged2e}

\usepackage[T1]{fontenc}

\usepackage[utf8]{inputenc}

\usepackage{booktabs}
\usepackage{multirow}
\usepackage{makecell} 

\usepackage{microtype}

\usepackage{cuted}

\usepackage{inconsolata}

\newcommand{\temprageval}{\textsc{TempRAGEval}\xspace}
\newcommand{\situatedqa}{\textsc{SituatedQA}\xspace}
\newcommand{\timeqa}{\textsc{TimeQA}\xspace}

\newcommand{\mrag}{MRAG\xspace}

\newcommand{\mini}{MiniLM\xspace}
\newcommand{\bgereranker}{BGE\xspace}
\newcommand{\bgegemma}{\textsc{Gemma}\xspace}

\newcommand{\contriever}{Contriever\xspace}

\newcommand{\eg}{\hbox{\emph{e.g.,}}\xspace}
\newcommand{\ie}{\hbox{\emph{i.e.,}}\xspace}

\newcolumntype{L}[1]{>{\raggedright\let\newline\\\arraybackslash\hspace{0pt}}m{#1}}
\newcolumntype{C}[1]{>{\centering\let\newline\\\arraybackslash\hspace{0pt}}m{#1}}
\newcolumntype{R}[1]{>{\raggedleft\let\newline\\\arraybackslash\hspace{0pt}}m{#1}}

\title{\mrag: A Modular Retrieval Framework for \\ Time-Sensitive Question Answering}

\author{
\textbf{Siyue Zhang}$^{\diamondsuit \heartsuit}$ \quad
\textbf{Yuxiang Xue}$^{\spadesuit}$ \quad
\textbf{Yiming Zhang}$^{\clubsuit}$ \quad
\textbf{Xiaobao Wu}$^{\diamondsuit}$ \\
\textbf{Anh Tuan Luu}$^{\diamondsuit}$ \quad
\textbf{Chen Zhao}$^{\spadesuit\bigstar}$}
\affil{
    $^{\diamondsuit}$Nanyang Technological University \quad $^{\heartsuit}$Alibaba-NTU JRI \quad $^{\spadesuit}$NYU Shanghai  \\
    $^{\clubsuit}$Zhejiang University \quad $^{\bigstar}$Center for Data Science, New York University \\
    \texttt{$\{$siyue001, xiaobao002$\}$@e.ntu.edu.sg} \qquad
    \texttt{anhtuan.luu@ntu.edu.sg} \\
    \texttt{$\{$yx3044, cz1285$\}$@nyu.edu} \qquad \texttt{yimingz@zju.edu.cn}}

\begin{document}

\maketitle

\begin{abstract}

Understanding temporal concepts and answering time-sensitive questions is crucial yet a challenging task for question-answering systems powered by large language models (LLMs). Existing approaches either update the parametric knowledge of LLMs with new facts, which is resource-intensive and often impractical, or integrate LLMs with external knowledge retrieval (\ie retrieval-augmented generation). However, off-the-shelf retrievers often struggle to identify relevant documents that require intensive temporal reasoning. To systematically study time-sensitive question answering, we introduce the \temprageval benchmark, which repurposes existing datasets by incorporating complex temporal perturbations and gold evidence labels. As anticipated, all existing retrieval methods struggle with these temporal reasoning-intensive questions. We further propose Modular Retrieval (\mrag), a trainless framework that includes three modules: (1) \emph{Question Processing} that decomposes question into a main content and a temporal constraint; (2) \emph{Retrieval and Summarization} that retrieves, splits, and summarize evidence passages based on the main content; (3) \emph{Semantic-Temporal Hybrid Ranking} that scores semantic and temporal relevance separately for each fine-grained evidence. On \temprageval, \mrag significantly outperforms baseline retrievers in retrieval performance, leading to further improvements in final answer accuracy.\footnote{Our code and data are available at \url{https://github/anonymous}.}

\end{abstract}

\section{Introduction}

\begin{figure}[t!]
    \centering
    \includegraphics[width=0.49\textwidth]{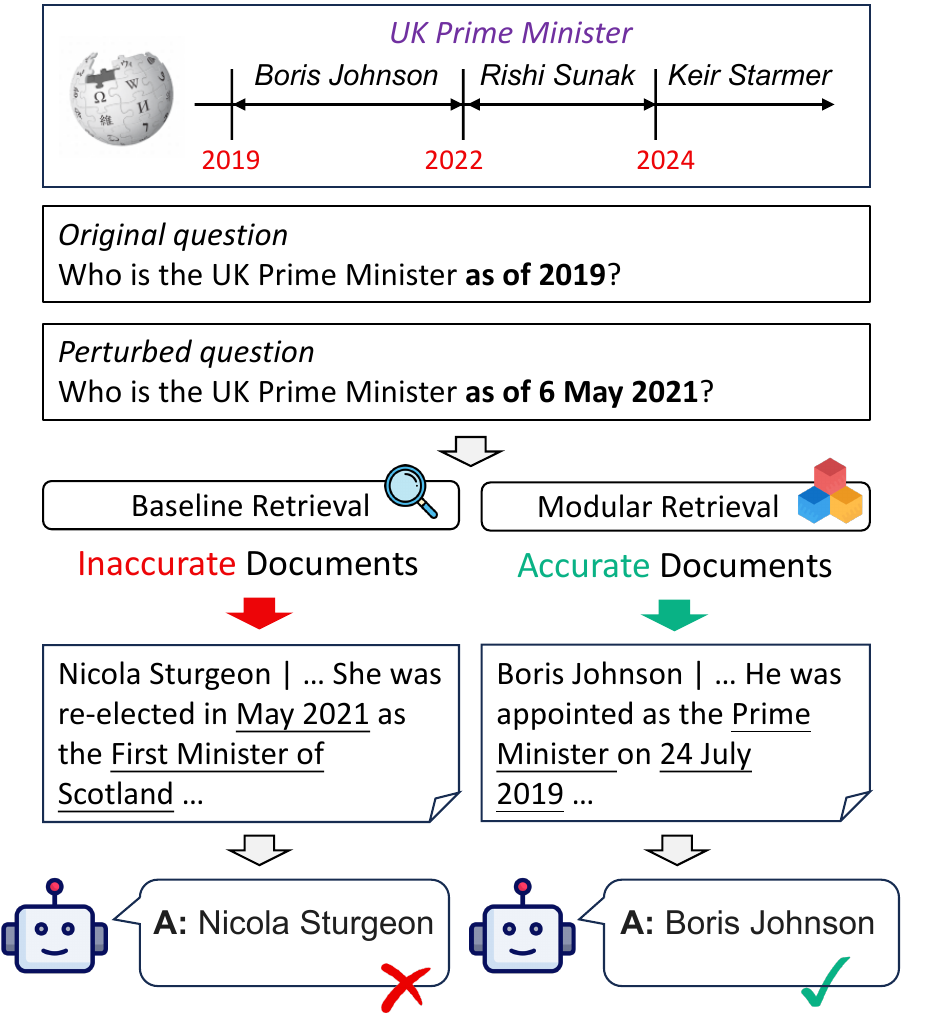}
    \captionsetup{justification=justified, singlelinecheck=false}  %
    \caption{A time-sensitive question example that requires temporal reasoning (\underline{as of 6 May 2021} $\rightarrow$ \underline{2019 - 2022}) to both retrieve documents and generate answers. State-of-the-art retrieval systems struggle to conduct in-depth reasoning to identify relevant documents. We provide a new diagnostic benchmark \temprageval, and propose a new modular framework to tackle this challenge.}
    
    \label{fig:idea}
\end{figure}

Facts are constantly evolving in our ever-changing world. This dynamic nature highlights the need for natural language processing (NLP) systems capable of updating information \citep{streamingqa2022, syntqa, realtimeqa} and providing accurate responses to time-sensitive questions \citep{timeqa, timebench}. For instance, a common query like ``Who is the UK Prime Minister?'' sees the answer transition from ``\texttt{Boris Johnson}'' to ``\texttt{Rishi Sunak}'' in 2022 (Figure~\ref{fig:idea}).

With developments of large language models (LLMs), existing approaches rely on the parametric knowledge of LLMs to answer time-sensitive questions directly, and constantly update the parametric knowledge on new facts \citep{knowledge, akew, easyedit}. However, updating LLM parameters are often resource-intensive. 
An alternative line of research explores Retrieval-Augmented Generation (RAG), which integrates LLMs with external knowledge (\eg Wikipedia) through information retrieval \citep{fid}. While RAG allows for the incorporation of new facts with minimal effort, its performance heavily relies on off-the-shelf retrieval systems, which are often limited to keywords or semantic matching. Time-sensitive questions, however, often require intensive temporal reasoning to identify relevant documents, \ie reasoning-intensive retrieval \citep{bright}. For example, in Figure~\ref{fig:idea}, retrievers should infer the date ``\texttt{24 July 2019}'' as relevant to the constraint ``\texttt{as of 6 May 2021}'', rather than only match with date like ``\texttt{May 2021}''. Although temporal QA is a well-established domain, there remains a lack of research for temporal reasoning-intensive retrieval systems. We fill that lacuna.

We begin by conducting a diagnostic evaluation of existing retrieval approaches for temporal reasoning-intensive retrieval. Following the idea of systematic evaluation with contrast set \citep{contrast_set}, we repurpose two existing datasets, \timeqa \citep{timeqa} and \situatedqa \citep{situatedqa}, to introduce the \textbf{Temp}oral QA for \textbf{RAG} \textbf{Eval}uation benchmark (\temprageval). We manually augment the test questions with complex temporal perturbations (\eg modifying the time period to avoid textual overlap). In addition, we annotate gold evidence on Wikipedia for more accurate retrieval evaluation. We observe remarkable degradation of retrieval performance, which points out a serious robustness issue for existing retrievers.

To address this issue, we propose a training-free Modular Retrieval framework (\mrag) to enhance temporal reasoning-intensive retrieval. Complementary to recent agentic RAG methods such as R1-Searcher and Search-o1 \citep{r1_searcher,search_o1}, \mrag provides an improved retrieval component and reduces the number of invocations for retrieval when handling complex temporal questions. Specifically, \mrag contains three key modules: (1) \textbf{Question Processing}, which decomposes each question into a main content and a temporal constraint; (2) \textbf{Retrieval and Summarization}, which utilizes off-the-shelf retrievers to find evidence passages based on the main content, segments them into independent sentences, and guides LLMs to condense the most relevant passages into query-specific sentences.(3) \textbf{Semantic-Temporal Hybrid Ranking}, which ranks each evidence sentence using a combination of a semantic score measuring semantic similarity, and a temporal score, a novel symbolic component that assesses temporal relevance to the query's temporal constraint.

On \temprageval, our proposed \mrag framework achieves substantial improvements in performance, with 9.3\% top-1 answer recall and 11\% top-1 evidence recall. We also incorporate state-of-the-art (SOTA) answer generators \citep{selfrag, crag, cot}, and demonstrate that the improvements in retrieval from \mrag propagate to enhanced final QA accuracy, with 4.5\% for both exact match and F1. Detailed case studies further confirms \mrag's robustness to temporal perturbations qualitatively.

Our contributions can be summarized as follows.
\begin{itemize}[leftmargin=*]
    \item We introduce \temprageval, a \textbf{time-sensitive RAG benchmark} to diagnostically evaluate each component of existing retrieval-augmented generation systems.
    \item We propose \mrag, a \textbf{modular retrieval framework} to separately determine semantic and temporal relevance.
    \item On \temprageval, \mrag significantly outperforms all baseline retrieval systems, and the improvements lead to better answer generation.
\end{itemize}

\section{Background}

In this section, we first define the time-sensitive question answering task (\S\ref{sec:problem}), and then introduce the baseline retrieval-augmented generation QA based systems (\S\ref{sec:rag}).

\subsection{Temporal Question Answering} 
\label{sec:problem}

There has been extensive research on Temporal Question Answering over Knowledge Graphs (TKGQA). Prior works have developed knowledge graph-based benchmarks (e.g., CronQuestions \citep{cronquestions} and TimeQuestions \citep{timequestions}), graph neural network models \citep{subgtr, local, twi}, and TKG-augmented LLMs \citep{timer4}. However, these methods heavily rely on knowledge graphs, which are costly to build, domain-specific, and prone to becoming outdated \citep{kg}. Therefore, we focus on the free-text setting, which is more universal and up-to-date. Instead of building temporal-specific graph models, we improve general-purpose text retrievers for temporal queries. We use Wikipedia as the main knowledge corpus $D$.\footnote{December 2021 Wikipedia dump, comprising 33.1 million text chunks.} Our approach is broadly applicable to other collections, such as the New York Times Annotated Archive \citep{nyt} and ClueWeb \citep{ClueWeb}.

\subsection{Retrieval-Augmented Generation} 
\label{sec:rag}

The goal of RAG is to address the limitations in the parametric knowledge of LLMs by incorporating external knowledge.

\paragraph{Passage retrieval and reranking.} Retrieval methods are typically categorized into sparse retrieval and dense retrieval. Sparse retrieval methods like BM25 \citep{bm25} rely on lexical matching. In contrast, dense retrieval models \citep{dpr, distant, SBERT, contriever} encode the question $q$ and passage $p$ into low-dimension vectors separately. The semantic similarity is computed using a scoring function (\eg dot product) as:
\begin{equation}
    f(q,p) = \text{sim}(\text{Enc}_{Q}(q), \text{Enc}_{P}(p)), p \in \mathcal{D}.
\end{equation}

However, these bi-encoder models lack the ability to capture fine-grained interactions between the query and passage. A common optional\footnote{Note that reranking is not always adopted, as it adds additional computational cost.} approach is to have another cross-encoder model to rerank top passages. Cross-encoder models \citep{ColBERT, minilm_hg, gemma} jointly encode the query $q$ and the passage $p$ together by concatenating them as input into a single model as:
\begin{equation}
    f(q,p) = \text{sim}(\text{Enc}([q;p])), p \in \mathcal{D}.
\end{equation}

\paragraph{Answer generation.}

Recent reader systems are mainly powered by LLMs with strong reasoning capabilities. With recent advancements in long-context LLMs \citep{llama3, longcontext}, top documents are typically concatenated with the query as reader input:
\begin{equation}
   y = \text{Dec}(\text{Enc}([p_1;\dots;p_k;q])).
\end{equation}
To unlock the reasoning capabilities of LLMs, Chain-of-Thought (CoT) prompting \citep{cot} introduces intermediate reasoning steps for improved performance. Self-RAG \citep{selfrag} critiques the retrieved passages and its own generations. Recent agentic RAG systems dynamically acquire and integrate external knowledge (using either retriever models or search APIs) during the reasoning process \citep{search_o1,r1_searcher,search_r1}.

\section{\temprageval Benchmark}
\label{sec:temprageval}

In this section, we first present existing time-sensitive QA datasets (\S\ref{sec:ext_data}), then introduce our diagnostic benchmark dataset \temprageval (\S\ref{sec:tempeval}), finally we evaluate existing retrieval approaches on \temprageval  (\S\ref{sec:rageval}).

\begin{table}[t]
\centering
\renewcommand{\arraystretch}{1}
\begin{tabular}{p{2cm}cccccc} 
\toprule
Dataset        & \# Eval.    & Evid.     & Natu.     & Comp.  \\
\midrule
ComplexTQA & 10M      &    &             &      $\checkmark$         \\
StreamingQA & 40K  &   $\checkmark$         & $\checkmark$       &         \\
TempLAMA     & 35K       &          &             &                      \\
SituatedQA    & 2K         &          & $\checkmark$           &                      \\
TimeQA       & 3K         &          &             &  $\checkmark$   \\
MenatQA     & 2K         &          &             &   $\checkmark$ \\
\hline
\textbf{TempRAGEval}   & 1K      & $\checkmark$   & $\checkmark$    &    $\checkmark$  \\
\bottomrule
\end{tabular}
\captionsetup{justification=justified,singlelinecheck=false}
\caption{Comparison of temporal QA datasets. \temprageval is featured by manual evidence annotations, human-written question (\ie Naturalness), and higher complexity in temporal reasoning.}
\label{table: comparison}
\end{table}

\subsection{Existing Time-Sensitive QA Datasets}
\label{sec:ext_data}

There are several existing QA datasets that focus on temporal reasoning. The most representative ones are the following:\footnote{We mainly focus on these two datasets, while others, such as \citet{streamingqa2022, templama, complextempqa}, serve as alternatives. Datasets such as \citet{tempreason, durationqa} that are not knowledge-intensive, are excluded from this work (\Cref{exclusion}).}

\begin{itemize}[leftmargin=*]
    \itemsep0em 

    \item \textbf{\situatedqa} \citep{situatedqa} is a time-sensitive QA dataset where the answer to an information-seeking question varies based on temporal context. These questions contain a single type of temporal constraint (\eg ``\texttt{as of}'') that directly align with the answers. Retrievers with surface-form date matching often exploit these shortcuts to bypass the need for temporal reasoning.

    \item \textbf{\timeqa} \citep{timeqa} is another time-sensitive QA dataset. Unlike \situatedqa, the questions in hard split include complex temporal constraints (\eg ``\texttt{between 2012 to 2018}''). However, question-answer pairs are \emph{synthetically} generated from time-evolving WikiData facts using templates. In addition, \timeqa does not include evidence annotations, making it imprecise to evaluate retrieval results.

\end{itemize}

According to \Cref{table: comparison}, we observe that none of existing datasets include these key factors for systematically evaluating current retrieval (and answer generation) systems: (1) Evidence annotation; (2) Natural questions from users;  (3) Complex temporal constraints. 
Therefore, as we will show in the
following section, we aim to address this limitation by
creating \temprageval.

\subsection{\temprageval Construction}
\label{sec:tempeval}

We create \temprageval, a time-sensitive QA benchmark for rigorously evaluating temporal reasoning in both retrieval and answer generation.

\paragraph{Perturbed question-answer pair generation.} Annotators first select question-answer pairs from the \situatedqa and \timeqa datasets\footnote{Since both datasets lack questions about knowledge beyond the cutoff date of existing LLMs, we primarily focus on historical knowledge and discuss potential future directions on recent knowledge in the Limitations section.} that can be grounded in Wikipedia facts with key timestamps or durations. They then revise temporal perturbations by selecting implicit conditions, temporal relations, and alternative dates to include complex temporal reasoning, without changing the final answer\footnote{Questions with the same content but different temporal constraints and answers are considered different samples. Perturbations are introduced for each sample.}. Annotators are also required to edit the question text to improve naturalness. We include detailed guidelines in \Cref{guidelines}.

\paragraph{Evidence annotation.} To better evaluate the performance of retrieval systems, we supplement question-answer pairs with up to two annotated gold evidence passages. A passage is relevant to the question if annotators can obtain the correct answer based on the passage. Specifically, for each question, annotators are asked to manually review top-20 passages retrieved by \contriever \citep{contriever} and reranked by the best \textsc{Gemma} \citep{bge_gemma} reranker. If there is no relevant passage, annotators are required to search Wikipedia pages related to the query entities to locate the gold evidence (around 12.7\% of questions). We create 1,000 test examples with human-annotated evidence. \Cref{statistics} presents sample statistics, revealing that \situatedqa questions include popular entities while the entities for \timeqa questions are long-tailed.

\subsection{Preliminary Evaluation on \temprageval}
\label{sec:rageval}

In \temprageval, we first evaluate the performance on SOTA retrieval systems as a sanity check. 

\paragraph{Experimental Setup.} We follow the popular retrieve-then-rerank pipeline, using the dense retriever \contriever \citep{contriever} and the LLM-based reranker \bgegemma \citep{gemma}. The retriever finds top 1,000 passages, and among them, the reranker reorders the top 100 passages. We use two evaluation metrics: \textbf{Answer Recall (AR@k)} that measures the proportion of samples where at least one answer appears within the top-\textit{k} retrieved passages, and \textbf{Gold Evidence Recall (ER@k)} that assesses the percentage of samples where at least one gold evidence document is included in the top-\textit{k} passages.

\begin{figure}[t!]
    \centering
    \includegraphics[width=0.4\textwidth]{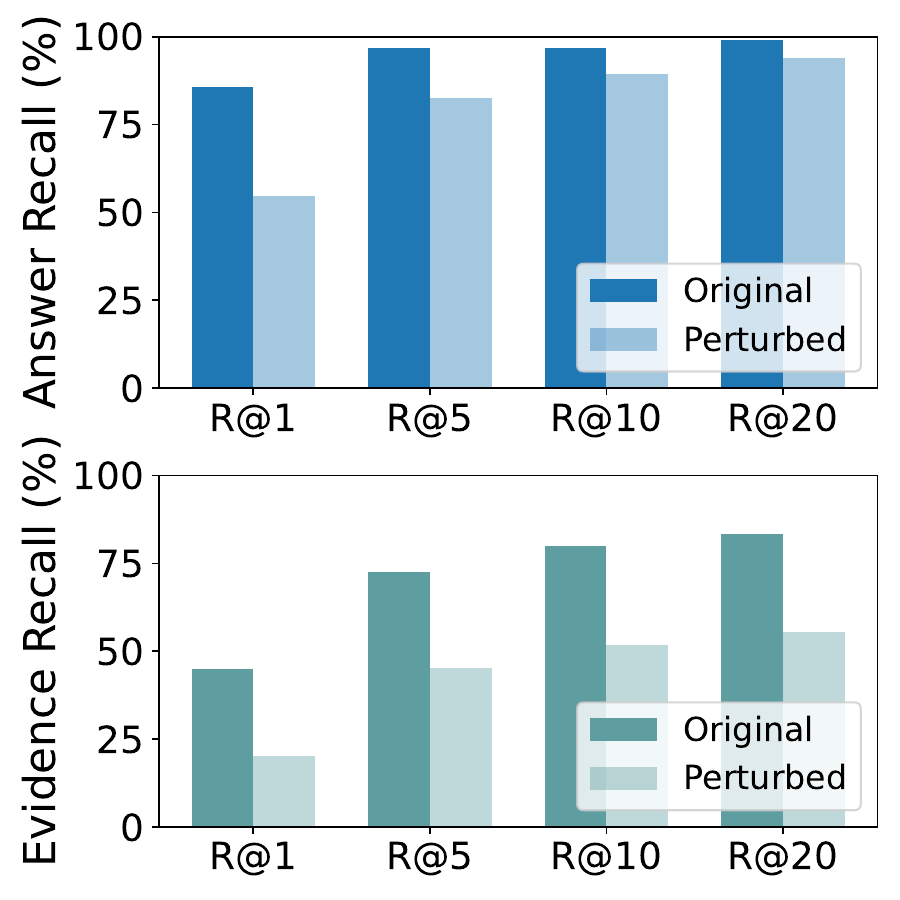}
    \captionsetup{justification=justified, singlelinecheck=false}  %
    \caption{
        The retrieval performance degradation of the \bgegemma baseline on \temprageval-\situatedqa, comparing original and perturbed questions (see \temprageval-\timeqa in \Cref{sec: degrade}).}
    \label{fig:preliminary}
\end{figure}

\paragraph{Performance degradation on perturbed questions.} As shown in \Cref{fig:preliminary}, we observe a significant degradation in retrieval performance caused by temporal perturbations. For instance, for the \bgegemma baseline, the top-1 answer recall and evidence recall drop from 85.8\% to 54.7\% and from 45.0\% to 20.3\% on \temprageval-\situatedqa. This is  because the perturbed temporal constraints avoid matching between timestamps in the questions and the passages. Consequently, retrievers must conduct in-depth temporal reasoning to identify the relevant passages. 

We further conduct a controlled experiment to reveal the temporal reasoning capabilities of existing retrieval methods. Specifically, we compute the similarity scores for query-evidence pairs by varying the temporal relation in the query, \eg ``\texttt{before}'', ``\texttt{after}'', and ``\texttt{as of}'', and the timestamp in the evidence, \eg from ``\texttt{1958}'' to ``\texttt{1965}''. Experiments confirm that all methods prioritize matching \emph{exact} dates indicating a shortcut for temporal reasoning in retrieval.  We present full results in \Cref{fig:control} in Appendix.

\section{\mrag: Modular Retrieval}
\label{sec:modular retrieval}

\begin{figure*}[t!]
    \centering
    \includegraphics[width=0.95\textwidth]{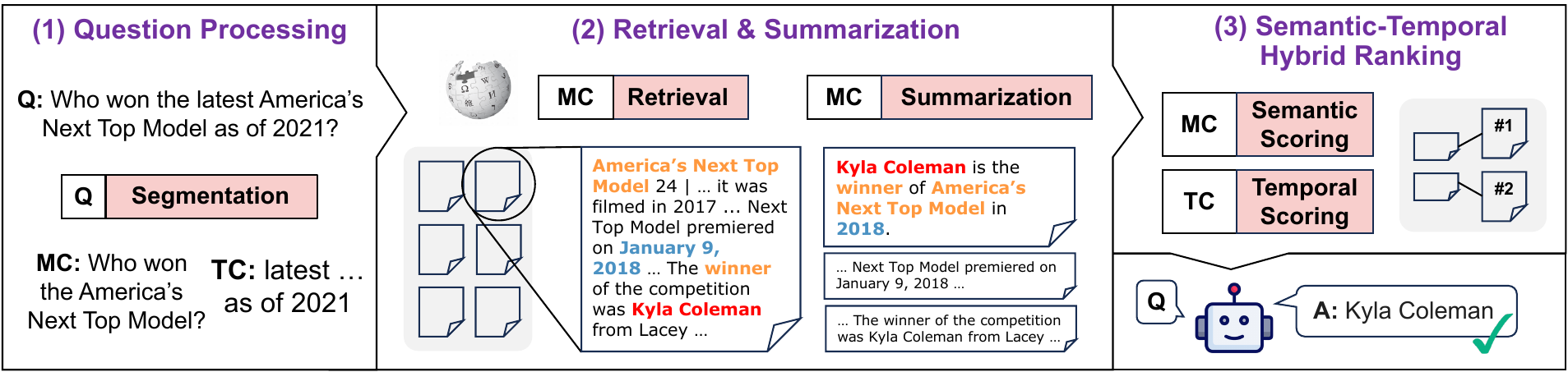}
    \captionsetup{justification=justified, singlelinecheck=false}  %
    \caption{An overview of the \mrag framework, consisting of three key modules: question processing, retrieval and summarization, and semantic-temporal hybrid ranking. The question processing module separates each query into the main content (\ie MC) and the temporal constraint (\ie TC). The retrieval and summarization module finds the most relevant evidence based on the main content and summarizes or splits these evidence into fine-grained sentences. The hybrid ranking module combines symbolic temporal scoring and dense embedding-based semantic scoring at a fine-grained level to determine the final evidence ranking.}
    \label{fig:system}
\end{figure*}

Motivated by the performance degradation of the existing retrieval methods in the preliminary evaluation, we propose a Modular Retrieval (\mrag) framework (as shown in \Cref{fig:system}) to enhance temporal reasoning-intensive retrieval. At a high level, \mrag disentangles relevance-based retrieval from temporal reasoning, leveraging a dense embedding model for semantic scoring and a set of symbolic heuristics for temporal scoring. Specifically, \mrag has three key modules: question processing, retrieval and summarization, and semantic-temporal hybrid ranking.

\paragraph{Question processing.} We prompt LLMs to decompose each time-sensitive question into a main content (MC) and a temporal constraint (TC). This approach disentangles temporal relevance from semantic relevance: MC measures the semantic relevance of the evidence, while TC determines its temporal relevance.

\paragraph{Retrieval and summarization.} We apply off-the-shelf retrievers (\eg \contriever) to find relevant passages to MC in Wikipedia. Then we employ reranker models to reorder these passages by semantic similarity to MC.

It is common for a passage to contain multiple pieces of temporal information, most of which are unrelated to the question and can distract the temporal scoring component introduced next. For example, the relevant passage in \Cref{fig:system} includes the sentence, ``\texttt{it was filmed in 2017}'', which satisfies the TC but is irrelevant. Therefore, we split passages into individual sentences to eliminate temporal distractors. However, as shown in \Cref{fig:system}, critical information from the most relevant passages—such as ``\texttt{America's Next Top Model}'', ``\texttt{The winner of the competition}'', and ''\texttt{January 9, 2018}''—can be scattered across different sentences. Relying solely on sentence splitting would miss key details. To overcome this challenge, we additionally employ LLMs to summarize each of the top-\textit{k} passages into a single sentence condensing relevant phrases and temporal information, as analyzed in \S\ref{sec: aba}.\footnote{While using LLMs to summarize passages can introduce hallucinations, we mitigate this by summarizing only the top-\textit{k} passages.}

\paragraph{Semantic-temporal hybrid ranking.} 
We rerank each sentence (summarized from a LLM or segmented from the original passage) with two distinct scores: a semantic score and a temporal score. The semantic score is calculated from the similarity between the evidence sentence and MC. For temporal score, we first extract the timestamp from each sentence (\eg ``\texttt{2018}''). Based on the timestamp and TC (\eg ``\texttt{as of 2021}''), we compute a temporal score using symbolic functions similar to temporal activation functions in \citet{subgtr}. The final score for each sentence is obtained by multiplying the semantic score and the temporal score.  Finally, we select the passages that contains the highest-scoring sentences. We include the details of symbolic functions in \Cref{spline}.

\section{Experiments}
\label{sec:exp}

In this section, we evaluate \mrag and baseline systems on \temprageval.

\subsection{Experimental Setup} 
\label{sec: setup}

\paragraph{Baselines.} For retrieval, we include BM25, \contriever, and a hybrid method \citep{hybrid}. Reranking methods include ELECTRA \citep{electra}, \mini \citep{minilm_hg}, Jina\citep{jina}, \bgereranker\citep{bge_reranker}, NV-Embed \citep{nv_embed}, and \bgegemma\citep{gemma}. We follow state-of-the-art answer generation approaches based on prompting LLMs (\S\ref{sec:rag}). We evaluate four approaches, \textbf{Direct Prompt} that adopts question-answer pairs as few-shot examples; \textbf{Direct CoT} that adds rationals into prompts \citep{cot}; \textbf{RAG-Concat}, where passages are concatenated into a LLM; and \textbf{Self-RAG}, which processes each passage independently and selects the best answer \citep{selfrag}.

\paragraph{Metrics.} We use the same setup in \S\ref{sec:rageval} for retrieval evaluation.  For answer evaluation, we use \textbf{Exact Match (EM)} that measures the exact match to the gold answer, and \textbf{F1 score (F1)} that measures the word overlap to the gold answer.

\paragraph{Implementation details.}  Due to limited budget, we evaluate GPT-4o mini \citep{gpt} with direct prompting, and three open-source LLMs: TIMO, a LLaMA2-13B model fine-tuned for temporal reasoning \citep{timo}, and two general-purpose models, Llama3.1-8B-Instruct and Llama3.1-70B-Instruct \cite{llama3}. We use 10 examples in prompts. To eliminate the impact of input length constraints across models, we conduct parametric studies as described in \Cref{sec: concat} and report each LLM's performance with its optimal number of input passages in \Cref{fig:reader_compare}.

\begin{table*}[t!]
\centering
\renewcommand{\arraystretch}{1}
\setlength{\tabcolsep}{12pt}
\arrayrulecolor{black}
\resizebox{\linewidth}{!}{
\begin{tabular}{ccccccccccc} 
\toprule
\multicolumn{3}{c}{\multirow{2}{*}{\textbf{Method}}}    & \multicolumn{4}{c}{\textbf{\temprageval-\timeqa}} & \multicolumn{4}{c}{\textbf{\temprageval-\situatedqa}}  \\ 
\cline{4-11}
\multicolumn{3}{c}{} & \multicolumn{2}{c}{\textbf{AR @}}    & \multicolumn{2}{c}{\textbf{ER @}}      & \multicolumn{2}{c}{\textbf{AR @}}    & \multicolumn{2}{c}{\textbf{ER @}}             \\ 
\hline
\textbf{1st}       &      \textbf{2nd}      &  \textbf{\# QFS} & \textbf{1}   & \textbf{5}    & \textbf{1}   & \textbf{5}                             & \textbf{1}    & \textbf{5}    & \textbf{1}   & \textbf{5} \\
\hline
BM25      & -   & -  &   17.5&39.0&4.2&14.1 &27.6&58.2&6.8&18.4 \\ 
Cont.     & -      & -   & 18.8 & 49.9 & 9.6 & 28.7   &  22.6 & 51.1 & 6.8  & 17.1 \\ 
Hybrid   & -  & -&  18.8&51.2&9.6&28.1 &22.6&55.8&6.8&19.7\\
\hline
Cont.      & ELECTRA   & -        &  40.1&76.9&21.8&58.6&35.5&71.3&15.3&37.1  \\
Cont.     & MiniLM   & -     & 34.0&76.1&16.2&57.3 &36.8&73.4&20.0&40.3  \\ 
Cont.     & Jina   & -     & 42.4&77.2&23.6&58.6&47.9&78.4&19.5&41.1  \\ 
Cont.       & BGE   & -         &  40.3&80.9&23.3&61.3 &36.3&74.2&14.5&35.0 \\ 
Cont.      & NV-Embed   & -        &  49.9&81.2&33.4&62.9&47.4&81.3&23.4&46.1   \\ 
Cont.       & Gemma      & -       & 46.7  & 82.5 & 26.0 & 66.6  &  54.7 & 82.6 & 20.3 & 45.3  \\
\hline
Cont.      & MRAG    & -  & 57.6 & 89.4 & 32.4 & 73.5  & 61.1 & 88.2 & 27.4 & 56.3 \\ 
Cont.   & MRAG     & 5   & \textbf{58.6} & \textbf{90.0} & \textbf{37.1} & \textbf{74.3}  & 61.3 & \textbf{89.0} & \textbf{31.1} & \textbf{59.2} \\ 
Cont.  & MRAG    & 10   &  56.0  & 88.1  & 35.5 & 73.2 & \textbf{62.1} & 87.9 &30.8& 57.9
\\ 
\arrayrulecolor{black}
\bottomrule
\end{tabular}
}
\captionsetup{justification=justified,singlelinecheck=false}
\caption{The answer recall (AR@k) and gold evidence recall (ER@k) of each retrieval method on perturbed temporal queries in \timeqa and \situatedqa subsets of \temprageval. 1st means the first-stage retrieving method; 2nd means the second-stage reranking method; \# QFS means the number of top passages to be summarized. \textbf{Bold} numbers indicate the best performance. We include complete results in \Cref{full_eval}.}
\label{compare_eval}
\end{table*}

\subsection{Main Results}

\label{sec: results}

\begin{table*}[t!]
\centering
\renewcommand{\arraystretch}{1}
\arrayrulecolor{black}
\setlength{\tabcolsep}{10pt}
\resizebox{0.9\linewidth}{!}{
\begin{tabular}{p{2.5cm}>{\centering\arraybackslash}p{1.2cm}>{\centering\arraybackslash}p{1.4cm}>{\centering\arraybackslash}p{1.4cm}>{\centering\arraybackslash}p{1.2cm}>{\centering\arraybackslash}p{1.4cm}>{\centering\arraybackslash}p{1.4cm}} 
\toprule
\multirow{2}{*}{\textbf{Method}}  & \multicolumn{3}{c}{\textbf{\temprageval-TimeQA}} & \multicolumn{3}{c}{\textbf{\temprageval-SituatedQA}}  \\ 
\cline{2-7}
& \textbf{\# Docs} & \textbf{EM} & \textbf{F1} & \textbf{\# Docs} & \textbf{EM} & \textbf{F1}
\\
\hline
\multicolumn{7}{c}{\textit{GPT4o-mini}}\\
Direct Prompt & - & 19.6 & 30.6 & - & 54.2 & 58.6 \\
\hline

\multicolumn{7}{c}{\textit{TIMO}}
\\
Direct Prompt & - & 16.2 & 24.8 & - & 50.6 & 53.1 \\
Direct CoT & - & 15.8 & 28.2& - & 49.4 & 53.9\\
RAG-Concat & 3 & 43.4 & \underline{55.2}& 3 & 55.8&58.1 \\
\mrag-Concat  & 3 & \textbf{48.2} & \textbf{57.2}& 3 & \underline{61.4} & \underline{63.6}\\
Self-\mrag  & 3 & \underline{44.6} & 54.9 & 3 & \textbf{62.4} & \textbf{64.5}\\
\hline
\multicolumn{7}{c}{\textit{Llama3.1-8B-Instruct}}\\
Direct Prompt & - & 16.0 & 23.9 & - & 42.8 & 45.0 \\
Direct CoT & - & 16.8 & 27.8 & - & 49.6 & 54.5\\
RAG-Concat & 5 & 44.0 & 52.8 & 5 & 60.0 & 62.7\\
\mrag-Concat & 5 & \underline{49.2} & \underline{59.2} & 5 & \underline{65.8} & \underline{68.0}\\
Self-\mrag & 5 & \textbf{54.2} &\textbf{65.6} & 5 & \textbf{66.4} &\textbf{68.2}\\
\hline
\multicolumn{7}{c}{\textit{Llama3.1-70B-Instruct}}\\
Direct Prompt & - & 31.0 & 42.3& - & 59.0 &62.1 \\
Direct CoT & - & 33.2 &45.8 & - & 69.0 &72.6 \\
RAG-Concat & 5 &54.4 &63.2 & 20 &67.0 &69.8  \\
\mrag-Concat & 5 & \underline{58.0} &\underline{68.4} & 20 & \underline{69.2} &\underline{72.5} \\
Self-\mrag & 5 & \textbf{61.2} & \textbf{75.3} & 20 & \textbf{72.2} & \textbf{76.0}\\
\arrayrulecolor{black}
\bottomrule
\end{tabular}
}
\caption{End-to-end QA performance comparison for various generation strategies and LLMs on \temprageval. \textbf{Bold} numbers indicate the best performance the each backbone LLM. The second best is \underline{underlined}. TIMO has a limited input length with up to three passages. We report the best number of passages for Llama models, and provide ablation on different numbers in \Cref{sec: concat}.}
\label{fig:reader_compare}
\end{table*}

\paragraph{\mrag enhances retrieval performance for time-sensitive questions.} According to \Cref{compare_eval}, \mrag significantly outperform all retrieve then rerank baselines, which highlight the superior temporal reasoning capabilities. For example, \mrag improves the best baselines Contriver + \bgegemma significantly, with 7.7\% top-5 evidence recall in \temprageval-\timeqa and 13.9\% top-5 evidence recall in \temprageval-\situatedqa.

\paragraph{Retrieval augmentation improves time-sensitive QA performance.} According to \Cref{fig:reader_compare}, we observe that LLMs relying solely on their parametric knowledge struggle to accurately answer time-sensitive questions, with limited QA accuracy. Incorporating retrieval-augmented generation significantly improves QA accuracy. Notably, we observe larger improvements on \temprageval-\timeqa, which primarily focuses on less frequent entities that pose greater challenges to the parametric knowledge of LLMs \citep{kandpal2023large}.

\paragraph{Enhanced retrieval contributes to improved time-sensitive QA performance.} As shown in \Cref{fig:reader_compare}, \mrag outperforms baseline RAG approaches in QA accuracy, for instance 49.2\% EM (MRAG) over 44.0\% EM (RAG) in \temprageval-\timeqa for Llama3.1-8B. Incorporating a self-reflection strategy improves performance for Llama3.1 models but not for TIMO, likely due to the limited reasoning capacity of its backbone model, Llama 2.

\section{Analysis}
This section presents a detailed analysis of the results and the contribution of each \mrag module.

\subsection{Ablation Study}
\label{sec: aba}

\paragraph{The impact of the number of passages for summarization.} 
Our LLM based summarization removes irrelevant temporal information but may also introduce hallucinations (details in \Cref{summarize_errors}). The parametric study at the bottom of \Cref{compare_eval} shows that summarizing top-five passages achieves the best balance. Additionally, we compare the RAG setup with the long-document QA setup bypassing retrievers in \Cref{longdoc}. RAG achieves a better accuracy, as retrievers exclude irrelevant passages and reduce noise.

\paragraph{The impact of the number of passages for answer generation.}
Our experiments show that the optimal number of passages depends on the LLMs. TIMO can handle a maximum of 3 passages, while the optimal number for LLaMA 3.1-8B is five, and for LLaMA 3.1-70B, it is twenty. Full results are presented in \Cref{sec: concat}.

\paragraph{The impact of key retrieval steps in \mrag.} In the retrieval and summarization module, with retrieved passages, \mrag conducts two key steps: (1) \textit{Passage Keyword Ranking} reduces the passages from 1,000 to 100 based on the keyword presence; (2) \textit{Passage Semantic Ranking} reorders the top-100 passages and splits them into $\sim$500 sentences, including chunk summaries. Similarly, in the semantic-temporal hybrid ranking module, another two key steps are conducted: (1) \textit{Sentence Keyword Ranking} further narrows the scope to 200 sentences; (2) \textit{Hybrid Ranking} ranks these sentences by semantic-temporal hybrid scoring. As shown in \Cref{ablation}, keyword-based ranking steps effectively reduce ranking candidates without significant performance loss, while semantic and hybrid ranking steps markedly enhance retrieval performance. An efficiency-accuracy trade-off can be made by adjusting the number of targeted passages or sentences at each step.

\paragraph{Computational Overhead.} As MRAG involves retrieval, summarization, and re-ranking, it incurs approximately twice the computational overhead of standard RAG pipelines, which is manageable. We provide a detailed analysis in \Cref{overhead}.

\subsection{Human Evaluation}
\label{sec: human_ana}
One limitation of the retrieval metrics is that AR@k overestimates performance, as a passage might incidentally contain an answer without directly supporting it. Conversely, ER@k acts as a conservative lower bound, potentially overlooking other relevant but unannotated passages. To address this, we conduct a human evaluation on a random subset of 200 examples to assess actual retrieval performance. The results validate the advantages of \mrag over \bgegemma in retrieval accuracy with full results presented in \Cref{sec: human}.

\subsection{Case Study}
\label{sec: case_ana}

\begin{figure}[t!]
    \centering
    \includegraphics[width=0.48\textwidth]{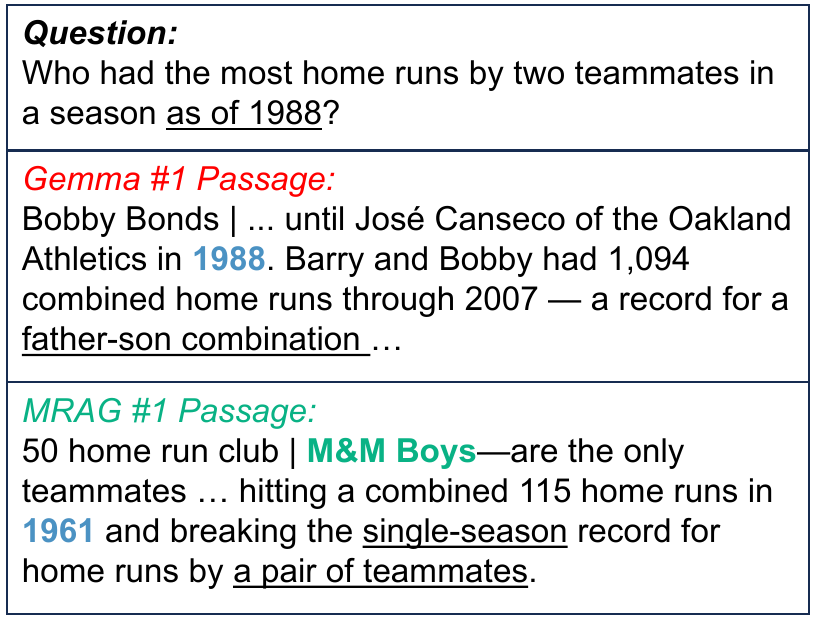}
    \captionsetup{justification=justified, singlelinecheck=false}  %
    \caption{A case study for top-1 passage retrieved by \bgegemma and \mrag from \temprageval.}
    \label{fig:case}
\end{figure}

We analyze retrieval errors qualitatively to highlight the advantages of \mrag over the \bgegemma baseline. As shown in \Cref{fig:case}, the top-1 passage by \bgegemma matches the query date ``\texttt{1988}'' but discusses a father-son record set in 2007. In contrast, \mrag retrieves a passage about a teammate combination record from the same season, despite the differing date (``\texttt{1961}'' vs. ``\texttt{1988}''). Additional cases and answer generation error cases are provided in \Cref{sec: cases_retrieval} and \Cref{sec: cases_qa}.

\section{Other Related Works}

\paragraph{LLM embeddings.} Recent research has explored LLM embeddings for retrieval. Some studies focus on distilling or fine-tuning LLM embeddings for reranking tasks, such as \bgegemma \citep{gemma} and MiniCPM \citep{hu2024minicpm}. Others aim to develop generalist embedding models capable of performing a wide range of tasks including retrieval and reranking, \eg gte-Qwen \citep{qwen} and NV-Embed \citep{nv_embed}. These LLM-based methods have demonstrated unprecedented performance in benchmarks such as MTEB \citep{mteb} and our \temprageval.

\paragraph{Reasoning intensive retrieval.} Existing retrieval benchmarks primarily target keyword-based or semantic-based retrieval. \citet{bright} introduces BRIGHT, a new retrieval task emphasizing intensive reasoning. We focus on temporal reasoning, one aspect of a broader class of reasoning-intensive retrieval. \mrag is expected to generalize to other forms of symbolic reasoning, such as numeric ranges and geospatial constraints. It mitigates direct numeric matching in retrieval and enhances reasoning capabilities.

\section{Conclusion}
This study focuses on time-sensitive QA, a task that challenges LLM based QA systems. We first present \temprageval, a diagnostic benchmark featuring natural questions, evidence annotations, and temporal complexity. We further propose a training-free \mrag framework, which disentangles relevance-based retrieval from temporal reasoning and introduces a symbolic temporal scoring mechanism. While existing systems struggle on \temprageval due to limited temporal reasoning capacities in retrieval, \mrag shows significant improvements. We hope this work advances future research on reasoning-intensive retrieval.

\section*{Limitations}
There are still some limitations in our work: (1) Our proposed benchmark is designed to evaluate time-sensitive questions with explicit temporal constraints. However, addressing questions with implicit temporal constraints presents a more complex challenge for retrieval systems. We could extend to implicit temporal reasoning by associating explicit information to the implicit one using LLM common sense and background knowledge like \citep{subgtr}. (2) Our dataset does not include time-sensitive questions that fall outside the LLM knowledge cutoff. We could extend our dataset with questions from RealTime QA \citep{realtimeqa} and AntiLeak-Bench \citep{wu2024antileak}. (3) Our main objective is to improve temporal reasoning in retrieval, which has not been tackled by previous works. More complex scenarios like multi-hop and recursive reasoning require further research efforts. (4) The proposed framework introduces computational overhead for improved performance as detailed in \Cref{overhead}. (5) We analyze knowledge conflicts between LLMs and passages in \Cref{sec: conflict} and leave conflicts among passages for future work.

\section*{Ethics Statement}
\temprageval were constructed upon the test set of \timeqa \citep{timeqa} and \situatedqa \citep{situatedqa} datasets, which are publicly available under the licenses BSD-3-Clause license\footnote{\url{https://opensource.org/licenses/BSD-3-Clause}} and Apache-2.0 license\footnote{\url{https://www.apache.org/licenses/LICENSE-2.0}}. These licenses all permit us to compose, modify, publish, and distribute additional annotations upon the original dataset. All the experiments in this paper were conducted on 4 NVIDIA L40S 46G GPUs. We hired 3 graduate students in STEM majors as annotators. We recommended that annotators spend at most 2 hours per day for annotation in order to reduce pressure and maintain a comfortable pace. The whole annotation work lasted about 5 days.

\bibliography{main.bbl}

\begin{thebibliography}{63}
\providecommand{\natexlab}[1]{#1}

\bibitem[{Asai et~al.(2024)Asai, Wu, Wang, Sil, and Hajishirzi}]{selfrag}
Akari Asai, Zeqiu Wu, Yizhong Wang, Avirup Sil, and Hannaneh Hajishirzi. 2024.
\newblock \href {https://arxiv.org/abs/2310.11511} {Self-rag: Learning to retrieve, generate, and critique through self-reflection}.
\newblock In \emph{International Conference on Learning Representations}.

\bibitem[{Bird and Loper(2004)}]{nltk}
Steven Bird and Edward Loper. 2004.
\newblock \href {https://aclanthology.org/P04-3031} {{NLTK}: The natural language toolkit}.
\newblock In \emph{Proceedings of the {ACL} Interactive Poster and Demonstration Sessions}.

\bibitem[{Chen et~al.(2021)Chen, Wang, and Wang}]{timeqa}
Wenhu Chen, Xinyi Wang, and William~Yang Wang. 2021.
\newblock \href {https://openreview.net/forum?id=9-LSfSU74n-} {A dataset for answering time-sensitive questions}.
\newblock \emph{2021 Conference on Neural Information Processing Systems Track on Datasets and Benchmarks}.

\bibitem[{Chen et~al.(2022)Chen, Zhao, Liao, Li, and Kanoulas}]{subgtr}
Ziyang Chen, Xiang Zhao, Jinzhi Liao, Xinyi Li, and Evangelos Kanoulas. 2022.
\newblock \href {https://doi.org/10.1016/j.knosys.2022.109134} {Temporal knowledge graph question answering via subgraph reasoning}.
\newblock \emph{Know.-Based Syst.}

\bibitem[{Chu et~al.(2024)Chu, Chen, Chen, Yu, Wang, Liu, and Qin}]{timebench}
Zheng Chu, Jingchang Chen, Qianglong Chen, Weijiang Yu, Haotian Wang, Ming Liu, and Bing Qin. 2024.
\newblock \href {https://aclanthology.org/2024.acl-long.66} {{T}ime{B}ench: A comprehensive evaluation of temporal reasoning abilities in large language models}.
\newblock In \emph{Proceedings of the 62nd Annual Meeting of the Association for Computational Linguistics}.

\bibitem[{Clark et~al.(2020)Clark, Luong, Le, and Manning}]{electra}
Kevin Clark, Minh-Thang Luong, Quoc~V. Le, and Christopher~D. Manning. 2020.
\newblock \href {https://openreview.net/pdf?id=r1xMH1BtvB} {{ELECTRA}: Pre-training text encoders as discriminators rather than generators}.
\newblock In \emph{International Conference on Learning Representations}.

\bibitem[{Dhingra et~al.(2022)Dhingra, Cole, Eisenschlos, Gillick, Eisenstein, and Cohen}]{templama}
Bhuwan Dhingra, Jeremy~R. Cole, Julian~Martin Eisenschlos, Daniel Gillick, Jacob Eisenstein, and William~W. Cohen. 2022.
\newblock \href {https://aclanthology.org/2022.tacl-1.15} {Time-aware language models as temporal knowledge bases}.
\newblock \emph{Transactions of the Association for Computational Linguistics}.

\bibitem[{Dubey et~al.(2024)Dubey, Jauhri, Pandey, Kadian, Al-Dahle, Letman, Mathur, Schelten, Yang, Fan, Goyal, Hartshorn, Yang, Mitra, Sravankumar, Korenev, Hinsvark, Rao, Zhang, Rodriguez, Gregerson, Spataru, Roziere, Biron, Tang, Chern, Caucheteux, Nayak, Bi, Marra, McConnell, Keller, Touret, Wu, Wong, Ferrer, Nikolaidis, Allonsius, Song, Pintz, Livshits, Esiobu, Choudhary, Mahajan, Garcia-Olano, Perino, Hupkes, Lakomkin, AlBadawy, Lobanova, Dinan, Smith, Radenovic, Zhang, Synnaeve, Lee, Anderson, Nail, Mialon, Pang, Cucurell, Nguyen, Korevaar, Xu, Touvron, Zarov, Ibarra, Kloumann, Misra, Evtimov, Copet, Lee, Geffert, Vranes, Park, Mahadeokar, Shah, van~der Linde, Billock, Hong, Lee, Fu, Chi, Huang, Liu, Wang, Yu, Bitton, Spisak, Park, Rocca, Johnstun, Saxe, Jia, Alwala, Upasani, Plawiak, Li, Heafield, Stone, El-Arini, Iyer, Malik, Chiu, Bhalla, Rantala-Yeary, van~der Maaten, Chen, Tan, Jenkins, Martin, Madaan, Malo, Blecher, Landzaat, de~Oliveira, Muzzi, Pasupuleti, Singh, Paluri, Kardas, Oldham, Rita,
  Pavlova, Kambadur, Lewis, Si, Singh, Hassan, Goyal, Torabi, Bashlykov, Bogoychev, Chatterji, Duchenne, Çelebi, Alrassy, Zhang, Li, Vasic, Weng, Bhargava, Dubal, Krishnan, Koura, Xu, He, Dong, Srinivasan, Ganapathy, Calderer, Cabral, Stojnic, Raileanu, Girdhar, Patel, Sauvestre, Polidoro, Sumbaly, Taylor, Silva, Hou, Wang, Hosseini, Chennabasappa, Singh, Bell, Kim, Edunov, Nie, Narang, Raparthy, Shen, Wan, Bhosale, Zhang, Vandenhende, Batra, Whitman, Sootla, Collot, Gururangan, Borodinsky, Herman, Fowler, Sheasha, Georgiou, Scialom, Speckbacher, Mihaylov, Xiao, Karn, Goswami, Gupta, Ramanathan, Kerkez, Gonguet, Do, Vogeti, Petrovic, Chu, Xiong, Fu, Meers, Martinet, Wang, Tan, Xie, Jia, Wang, Goldschlag, Gaur, Babaei, Wen, Song, Zhang, Li, Mao, Coudert, Yan, Chen, Papakipos, Singh, Grattafiori, Jain, Kelsey, Shajnfeld, Gangidi, Victoria, Goldstand, Menon, Sharma, Boesenberg, Vaughan, Baevski, Feinstein, Kallet, Sangani, Yunus, Lupu, Alvarado, Caples, Gu, Ho, Poulton, Ryan, Ramchandani, Franco, Saraf,
  Chowdhury, Gabriel, Bharambe, Eisenman, Yazdan, James, Maurer, Leonhardi, Huang, Loyd, Paola, Paranjape, Liu, Wu, Ni, Hancock, Wasti, Spence, Stojkovic, Gamido, Montalvo, Parker, Burton, Mejia, Wang, Kim, Zhou, Hu, Chu, Cai, Tindal, Feichtenhofer, Civin, Beaty, Kreymer, Li, Wyatt, Adkins, Xu, Testuggine, David, Parikh, Liskovich, Foss, Wang, Le, Holland, Dowling, Jamil, Montgomery, Presani, Hahn, Wood, Brinkman, Arcaute, Dunbar, Smothers, Sun, Kreuk, Tian, Ozgenel, Caggioni, Guzmán, Kanayet, Seide, Florez, Schwarz, Badeer, Swee, Halpern, Thattai, Herman, Sizov, Guangyi, Zhang, Lakshminarayanan, Shojanazeri, Zou, Wang, Zha, Habeeb, Rudolph, Suk, Aspegren, Goldman, Damlaj, Molybog, Tufanov, Veliche, Gat, Weissman, Geboski, Kohli, Asher, Gaya, Marcus, Tang, Chan, Zhen, Reizenstein, Teboul, Zhong, Jin, Yang, Cummings, Carvill, Shepard, McPhie, Torres, Ginsburg, Wang, Wu, U, Saxena, Prasad, Khandelwal, Zand, Matosich, Veeraraghavan, Michelena, Li, Huang, Chawla, Lakhotia, Huang, Chen, Garg, A, Silva, Bell,
  Zhang, Guo, Yu, Moshkovich, Wehrstedt, Khabsa, Avalani, Bhatt, Tsimpoukelli, Mankus, Hasson, Lennie, Reso, Groshev, Naumov, Lathi, Keneally, Seltzer, Valko, Restrepo, Patel, Vyatskov, Samvelyan, Clark, Macey, Wang, Hermoso, Metanat, Rastegari, Bansal, Santhanam, Parks, White, Bawa, Singhal, Egebo, Usunier, Laptev, Dong, Zhang, Cheng, Chernoguz, Hart, Salpekar, Kalinli, Kent, Parekh, Saab, Balaji, Rittner, Bontrager, Roux, Dollar, Zvyagina, Ratanchandani, Yuvraj, Liang, Alao, Rodriguez, Ayub, Murthy, Nayani, Mitra, Li, Hogan, Battey, Wang, Maheswari, Howes, Rinott, Bondu, Datta, Chugh, Hunt, Dhillon, Sidorov, Pan, Verma, Yamamoto, Ramaswamy, Lindsay, Lindsay, Feng, Lin, Zha, Shankar, Zhang, Zhang, Wang, Agarwal, Sajuyigbe, Chintala, Max, Chen, Kehoe, Satterfield, Govindaprasad, Gupta, Cho, Virk, Subramanian, Choudhury, Goldman, Remez, Glaser, Best, Kohler, Robinson, Li, Zhang, Matthews, Chou, Shaked, Vontimitta, Ajayi, Montanez, Mohan, Kumar, Mangla, Albiero, Ionescu, Poenaru, Mihailescu, Ivanov, Li, Wang,
  Jiang, Bouaziz, Constable, Tang, Wang, Wu, Wang, Xia, Wu, Gao, Chen, Hu, Jia, Qi, Li, Zhang, Zhang, Adi, Nam, Yu, Wang, Hao, Qian, He, Rait, DeVito, Rosnbrick, Wen, Yang, and Zhao}]{llama3}
Abhimanyu Dubey, Abhinav Jauhri, Abhinav Pandey, Abhishek Kadian, Ahmad Al-Dahle, Aiesha Letman, Akhil Mathur, Alan Schelten, Amy Yang, Angela Fan, Anirudh Goyal, Anthony Hartshorn, Aobo Yang, Archi Mitra, Archie Sravankumar, Artem Korenev, Arthur Hinsvark, Arun Rao, Aston Zhang, Aurelien Rodriguez, Austen Gregerson, Ava Spataru, Baptiste Roziere, Bethany Biron, Binh Tang, Bobbie Chern, Charlotte Caucheteux, Chaya Nayak, Chloe Bi, Chris Marra, Chris McConnell, Christian Keller, Christophe Touret, Chunyang Wu, Corinne Wong, Cristian~Canton Ferrer, Cyrus Nikolaidis, Damien Allonsius, Daniel Song, Danielle Pintz, Danny Livshits, David Esiobu, Dhruv Choudhary, Dhruv Mahajan, Diego Garcia-Olano, Diego Perino, Dieuwke Hupkes, Egor Lakomkin, Ehab AlBadawy, Elina Lobanova, Emily Dinan, Eric~Michael Smith, Filip Radenovic, Frank Zhang, Gabriel Synnaeve, Gabrielle Lee, Georgia~Lewis Anderson, Graeme Nail, Gregoire Mialon, Guan Pang, Guillem Cucurell, Hailey Nguyen, Hannah Korevaar, Hu~Xu, Hugo Touvron, Iliyan Zarov,
  Imanol~Arrieta Ibarra, Isabel Kloumann, Ishan Misra, Ivan Evtimov, Jade Copet, Jaewon Lee, Jan Geffert, Jana Vranes, Jason Park, Jay Mahadeokar, Jeet Shah, Jelmer van~der Linde, Jennifer Billock, Jenny Hong, Jenya Lee, Jeremy Fu, Jianfeng Chi, Jianyu Huang, Jiawen Liu, Jie Wang, Jiecao Yu, Joanna Bitton, Joe Spisak, Jongsoo Park, Joseph Rocca, Joshua Johnstun, Joshua Saxe, Junteng Jia, Kalyan~Vasuden Alwala, Kartikeya Upasani, Kate Plawiak, Ke~Li, Kenneth Heafield, Kevin Stone, Khalid El-Arini, Krithika Iyer, Kshitiz Malik, Kuenley Chiu, Kunal Bhalla, Lauren Rantala-Yeary, Laurens van~der Maaten, Lawrence Chen, Liang Tan, Liz Jenkins, Louis Martin, Lovish Madaan, Lubo Malo, Lukas Blecher, Lukas Landzaat, Luke de~Oliveira, Madeline Muzzi, Mahesh Pasupuleti, Mannat Singh, Manohar Paluri, Marcin Kardas, Mathew Oldham, Mathieu Rita, Maya Pavlova, Melanie Kambadur, Mike Lewis, Min Si, Mitesh~Kumar Singh, Mona Hassan, Naman Goyal, Narjes Torabi, Nikolay Bashlykov, Nikolay Bogoychev, Niladri Chatterji, Olivier
  Duchenne, Onur Çelebi, Patrick Alrassy, Pengchuan Zhang, Pengwei Li, Petar Vasic, Peter Weng, Prajjwal Bhargava, Pratik Dubal, Praveen Krishnan, Punit~Singh Koura, Puxin Xu, Qing He, Qingxiao Dong, Ragavan Srinivasan, Raj Ganapathy, Ramon Calderer, Ricardo~Silveira Cabral, Robert Stojnic, Roberta Raileanu, Rohit Girdhar, Rohit Patel, Romain Sauvestre, Ronnie Polidoro, Roshan Sumbaly, Ross Taylor, Ruan Silva, Rui Hou, Rui Wang, Saghar Hosseini, Sahana Chennabasappa, Sanjay Singh, Sean Bell, Seohyun~Sonia Kim, Sergey Edunov, Shaoliang Nie, Sharan Narang, Sharath Raparthy, Sheng Shen, Shengye Wan, Shruti Bhosale, Shun Zhang, Simon Vandenhende, Soumya Batra, Spencer Whitman, Sten Sootla, Stephane Collot, Suchin Gururangan, Sydney Borodinsky, Tamar Herman, Tara Fowler, Tarek Sheasha, Thomas Georgiou, Thomas Scialom, Tobias Speckbacher, Todor Mihaylov, Tong Xiao, Ujjwal Karn, Vedanuj Goswami, Vibhor Gupta, Vignesh Ramanathan, Viktor Kerkez, Vincent Gonguet, Virginie Do, Vish Vogeti, Vladan Petrovic, Weiwei Chu,
  Wenhan Xiong, Wenyin Fu, Whitney Meers, Xavier Martinet, Xiaodong Wang, Xiaoqing~Ellen Tan, Xinfeng Xie, Xuchao Jia, Xuewei Wang, Yaelle Goldschlag, Yashesh Gaur, Yasmine Babaei, Yi~Wen, Yiwen Song, Yuchen Zhang, Yue Li, Yuning Mao, Zacharie~Delpierre Coudert, Zheng Yan, Zhengxing Chen, Zoe Papakipos, Aaditya Singh, Aaron Grattafiori, Abha Jain, Adam Kelsey, Adam Shajnfeld, Adithya Gangidi, Adolfo Victoria, Ahuva Goldstand, Ajay Menon, Ajay Sharma, Alex Boesenberg, Alex Vaughan, Alexei Baevski, Allie Feinstein, Amanda Kallet, Amit Sangani, Anam Yunus, Andrei Lupu, Andres Alvarado, Andrew Caples, Andrew Gu, Andrew Ho, Andrew Poulton, Andrew Ryan, Ankit Ramchandani, Annie Franco, Aparajita Saraf, Arkabandhu Chowdhury, Ashley Gabriel, Ashwin Bharambe, Assaf Eisenman, Azadeh Yazdan, Beau James, Ben Maurer, Benjamin Leonhardi, Bernie Huang, Beth Loyd, Beto~De Paola, Bhargavi Paranjape, Bing Liu, Bo~Wu, Boyu Ni, Braden Hancock, Bram Wasti, Brandon Spence, Brani Stojkovic, Brian Gamido, Britt Montalvo, Carl
  Parker, Carly Burton, Catalina Mejia, Changhan Wang, Changkyu Kim, Chao Zhou, Chester Hu, Ching-Hsiang Chu, Chris Cai, Chris Tindal, Christoph Feichtenhofer, Damon Civin, Dana Beaty, Daniel Kreymer, Daniel Li, Danny Wyatt, David Adkins, David Xu, Davide Testuggine, Delia David, Devi Parikh, Diana Liskovich, Didem Foss, Dingkang Wang, Duc Le, Dustin Holland, Edward Dowling, Eissa Jamil, Elaine Montgomery, Eleonora Presani, Emily Hahn, Emily Wood, Erik Brinkman, Esteban Arcaute, Evan Dunbar, Evan Smothers, Fei Sun, Felix Kreuk, Feng Tian, Firat Ozgenel, Francesco Caggioni, Francisco Guzmán, Frank Kanayet, Frank Seide, Gabriela~Medina Florez, Gabriella Schwarz, Gada Badeer, Georgia Swee, Gil Halpern, Govind Thattai, Grant Herman, Grigory Sizov, Guangyi, Zhang, Guna Lakshminarayanan, Hamid Shojanazeri, Han Zou, Hannah Wang, Hanwen Zha, Haroun Habeeb, Harrison Rudolph, Helen Suk, Henry Aspegren, Hunter Goldman, Ibrahim Damlaj, Igor Molybog, Igor Tufanov, Irina-Elena Veliche, Itai Gat, Jake Weissman, James
  Geboski, James Kohli, Japhet Asher, Jean-Baptiste Gaya, Jeff Marcus, Jeff Tang, Jennifer Chan, Jenny Zhen, Jeremy Reizenstein, Jeremy Teboul, Jessica Zhong, Jian Jin, Jingyi Yang, Joe Cummings, Jon Carvill, Jon Shepard, Jonathan McPhie, Jonathan Torres, Josh Ginsburg, Junjie Wang, Kai Wu, Kam~Hou U, Karan Saxena, Karthik Prasad, Kartikay Khandelwal, Katayoun Zand, Kathy Matosich, Kaushik Veeraraghavan, Kelly Michelena, Keqian Li, Kun Huang, Kunal Chawla, Kushal Lakhotia, Kyle Huang, Lailin Chen, Lakshya Garg, Lavender A, Leandro Silva, Lee Bell, Lei Zhang, Liangpeng Guo, Licheng Yu, Liron Moshkovich, Luca Wehrstedt, Madian Khabsa, Manav Avalani, Manish Bhatt, Maria Tsimpoukelli, Martynas Mankus, Matan Hasson, Matthew Lennie, Matthias Reso, Maxim Groshev, Maxim Naumov, Maya Lathi, Meghan Keneally, Michael~L. Seltzer, Michal Valko, Michelle Restrepo, Mihir Patel, Mik Vyatskov, Mikayel Samvelyan, Mike Clark, Mike Macey, Mike Wang, Miquel~Jubert Hermoso, Mo~Metanat, Mohammad Rastegari, Munish Bansal, Nandhini
  Santhanam, Natascha Parks, Natasha White, Navyata Bawa, Nayan Singhal, Nick Egebo, Nicolas Usunier, Nikolay~Pavlovich Laptev, Ning Dong, Ning Zhang, Norman Cheng, Oleg Chernoguz, Olivia Hart, Omkar Salpekar, Ozlem Kalinli, Parkin Kent, Parth Parekh, Paul Saab, Pavan Balaji, Pedro Rittner, Philip Bontrager, Pierre Roux, Piotr Dollar, Polina Zvyagina, Prashant Ratanchandani, Pritish Yuvraj, Qian Liang, Rachad Alao, Rachel Rodriguez, Rafi Ayub, Raghotham Murthy, Raghu Nayani, Rahul Mitra, Raymond Li, Rebekkah Hogan, Robin Battey, Rocky Wang, Rohan Maheswari, Russ Howes, Ruty Rinott, Sai~Jayesh Bondu, Samyak Datta, Sara Chugh, Sara Hunt, Sargun Dhillon, Sasha Sidorov, Satadru Pan, Saurabh Verma, Seiji Yamamoto, Sharadh Ramaswamy, Shaun Lindsay, Shaun Lindsay, Sheng Feng, Shenghao Lin, Shengxin~Cindy Zha, Shiva Shankar, Shuqiang Zhang, Shuqiang Zhang, Sinong Wang, Sneha Agarwal, Soji Sajuyigbe, Soumith Chintala, Stephanie Max, Stephen Chen, Steve Kehoe, Steve Satterfield, Sudarshan Govindaprasad, Sumit Gupta,
  Sungmin Cho, Sunny Virk, Suraj Subramanian, Sy~Choudhury, Sydney Goldman, Tal Remez, Tamar Glaser, Tamara Best, Thilo Kohler, Thomas Robinson, Tianhe Li, Tianjun Zhang, Tim Matthews, Timothy Chou, Tzook Shaked, Varun Vontimitta, Victoria Ajayi, Victoria Montanez, Vijai Mohan, Vinay~Satish Kumar, Vishal Mangla, Vítor Albiero, Vlad Ionescu, Vlad Poenaru, Vlad~Tiberiu Mihailescu, Vladimir Ivanov, Wei Li, Wenchen Wang, Wenwen Jiang, Wes Bouaziz, Will Constable, Xiaocheng Tang, Xiaofang Wang, Xiaojian Wu, Xiaolan Wang, Xide Xia, Xilun Wu, Xinbo Gao, Yanjun Chen, Ye~Hu, Ye~Jia, Ye~Qi, Yenda Li, Yilin Zhang, Ying Zhang, Yossi Adi, Youngjin Nam, Yu, Wang, Yuchen Hao, Yundi Qian, Yuzi He, Zach Rait, Zachary DeVito, Zef Rosnbrick, Zhaoduo Wen, Zhenyu Yang, and Zhiwei Zhao. 2024.
\newblock \href {https://arxiv.org/abs/2407.21783} {The llama 3 herd of models}.

\bibitem[{Gardner et~al.(2020)Gardner, Artzi, Basmov, Berant, Bogin, Chen, Dasigi, Dua, Elazar, Gottumukkala, Gupta, Hajishirzi, Ilharco, Khashabi, Lin, Liu, Liu, Mulcaire, Ning, Singh, Smith, Subramanian, Tsarfaty, Wallace, Zhang, and Zhou}]{contrast_set}
Matt Gardner, Yoav Artzi, Victoria Basmov, Jonathan Berant, Ben Bogin, Sihao Chen, Pradeep Dasigi, Dheeru Dua, Yanai Elazar, Ananth Gottumukkala, Nitish Gupta, Hannaneh Hajishirzi, Gabriel Ilharco, Daniel Khashabi, Kevin Lin, Jiangming Liu, Nelson~F. Liu, Phoebe Mulcaire, Qiang Ning, Sameer Singh, Noah~A. Smith, Sanjay Subramanian, Reut Tsarfaty, Eric Wallace, Ally Zhang, and Ben Zhou. 2020.
\newblock \href {https://aclanthology.org/2020.findings-emnlp.117} {Evaluating models{'} local decision boundaries via contrast sets}.
\newblock In \emph{Findings of the Association for Computational Linguistics: EMNLP 2020}.

\bibitem[{Gemma et~al.(2024)Gemma, Mesnard, Hardin, Dadashi, Bhupatiraju, Pathak, Sifre, Rivière, Kale, Love, Tafti, Hussenot, Sessa, Chowdhery, Roberts, Barua, Botev, Castro-Ros, Slone, Héliou, Tacchetti, Bulanova, Paterson, Tsai, Shahriari, Lan, Choquette-Choo, Crepy, Cer, Ippolito, Reid, Buchatskaya, Ni, Noland, Yan, Tucker, Muraru, Rozhdestvenskiy, Michalewski, Tenney, Grishchenko, Austin, Keeling, Labanowski, Lespiau, Stanway, Brennan, Chen, Ferret, Chiu, Mao-Jones, Lee, Yu, Millican, Sjoesund, Lee, Dixon, Reid, Mikuła, Wirth, Sharman, Chinaev, Thain, Bachem, Chang, Wahltinez, Bailey, Michel, Yotov, Chaabouni, Comanescu, Jana, Anil, McIlroy, Liu, Mullins, Smith, Borgeaud, Girgin, Douglas, Pandya, Shakeri, De, Klimenko, Hennigan, Feinberg, Stokowiec, hui Chen, Ahmed, Gong, Warkentin, Peran, Giang, Farabet, Vinyals, Dean, Kavukcuoglu, Hassabis, Ghahramani, Eck, Barral, Pereira, Collins, Joulin, Fiedel, Senter, Andreev, and Kenealy}]{gemma}
Gemma, Thomas Mesnard, Cassidy Hardin, Robert Dadashi, Surya Bhupatiraju, Shreya Pathak, Laurent Sifre, Morgane Rivière, Mihir~Sanjay Kale, Juliette Love, Pouya Tafti, Léonard Hussenot, Pier~Giuseppe Sessa, Aakanksha Chowdhery, Adam Roberts, Aditya Barua, Alex Botev, Alex Castro-Ros, Ambrose Slone, Amélie Héliou, Andrea Tacchetti, Anna Bulanova, Antonia Paterson, Beth Tsai, Bobak Shahriari, Charline~Le Lan, Christopher~A. Choquette-Choo, Clément Crepy, Daniel Cer, Daphne Ippolito, David Reid, Elena Buchatskaya, Eric Ni, Eric Noland, Geng Yan, George Tucker, George-Christian Muraru, Grigory Rozhdestvenskiy, Henryk Michalewski, Ian Tenney, Ivan Grishchenko, Jacob Austin, James Keeling, Jane Labanowski, Jean-Baptiste Lespiau, Jeff Stanway, Jenny Brennan, Jeremy Chen, Johan Ferret, Justin Chiu, Justin Mao-Jones, Katherine Lee, Kathy Yu, Katie Millican, Lars~Lowe Sjoesund, Lisa Lee, Lucas Dixon, Machel Reid, Maciej Mikuła, Mateo Wirth, Michael Sharman, Nikolai Chinaev, Nithum Thain, Olivier Bachem, Oscar
  Chang, Oscar Wahltinez, Paige Bailey, Paul Michel, Petko Yotov, Rahma Chaabouni, Ramona Comanescu, Reena Jana, Rohan Anil, Ross McIlroy, Ruibo Liu, Ryan Mullins, Samuel~L Smith, Sebastian Borgeaud, Sertan Girgin, Sholto Douglas, Shree Pandya, Siamak Shakeri, Soham De, Ted Klimenko, Tom Hennigan, Vlad Feinberg, Wojciech Stokowiec, Yu~hui Chen, Zafarali Ahmed, Zhitao Gong, Tris Warkentin, Ludovic Peran, Minh Giang, Clément Farabet, Oriol Vinyals, Jeff Dean, Koray Kavukcuoglu, Demis Hassabis, Zoubin Ghahramani, Douglas Eck, Joelle Barral, Fernando Pereira, Eli Collins, Armand Joulin, Noah Fiedel, Evan Senter, Alek Andreev, and Kathleen Kenealy. 2024.
\newblock \href {https://arxiv.org/abs/2403.08295} {Gemma: Open models based on gemini research and technology}.

\bibitem[{Gruber et~al.(2024)Gruber, Abdallah, F{\"a}rber, and Jatowt}]{complextempqa}
Raphael Gruber, Abdelrahman Abdallah, Michael F{\"a}rber, and Adam Jatowt. 2024.
\newblock \href {https://arxiv.org/abs/2406.04866} {Complextempqa: A large-scale dataset for complex temporal question answering}.
\newblock \emph{arXiv preprint arXiv:2406.04866}.

\bibitem[{Hu et~al.(2024)Hu, Tu, Han, He, Cui, Long, Zheng, Fang, Huang, Zhao, Zhang, Thai, Zhang, Wang, Yao, Zhao, Zhou, Cai, Zhai, Ding, Jia, Zeng, Li, Liu, and Sun}]{hu2024minicpm}
Shengding Hu, Yuge Tu, Xu~Han, Chaoqun He, Ganqu Cui, Xiang Long, Zhi Zheng, Yewei Fang, Yuxiang Huang, Weilin Zhao, Xinrong Zhang, Zheng~Leng Thai, Kaihuo Zhang, Chongyi Wang, Yuan Yao, Chenyang Zhao, Jie Zhou, Jie Cai, Zhongwu Zhai, Ning Ding, Chao Jia, Guoyang Zeng, Dahai Li, Zhiyuan Liu, and Maosong Sun. 2024.
\newblock \href {https://arxiv.org/pdf/2404.06395} {Minicpm: Unveiling the potential of small language models with scalable training strategies}.

\bibitem[{Izacard et~al.(2021)Izacard, Caron, Hosseini, Riedel, Bojanowski, Joulin, and Grave}]{contriever}
Gautier Izacard, Mathilde Caron, Lucas Hosseini, Sebastian Riedel, Piotr Bojanowski, Armand Joulin, and Edouard Grave. 2021.
\newblock \href {https://openreview.net/pdf?id=jKN1pXi7b0} {Unsupervised dense information retrieval with contrastive learning}.
\newblock \emph{Transactions on Machine Learning Research}.

\bibitem[{Izacard et~al.(2020)Izacard, Petroni, Hosseini, De~Cao, Riedel, and Grave}]{fid}
Gautier Izacard, Fabio Petroni, Lucas Hosseini, Nicola De~Cao, Sebastian Riedel, and Edouard Grave. 2020.
\newblock \href {https://hal.science/hal-03463488/document} {A memory efficient baseline for open domain question answering}.
\newblock \emph{arXiv preprint arXiv:2012.15156}.

\bibitem[{Jedidi et~al.(2024)Jedidi, Chuang, Shing, and Glass}]{hybrid}
Nour Jedidi, Yung-Sung Chuang, Leslie Shing, and James Glass. 2024.
\newblock \href {https://arxiv.org/abs/2410.21242} {Zero-shot dense retrieval with embeddings from relevance feedback}.

\bibitem[{Jia et~al.(2021)Jia, Pramanik, Saha~Roy, and Weikum}]{timequestions}
Zhen Jia, Soumajit Pramanik, Rishiraj Saha~Roy, and Gerhard Weikum. 2021.
\newblock \href {http://dx.doi.org/10.1145/3459637.3482416} {Complex temporal question answering on knowledge graphs}.
\newblock In \emph{Proceedings of the 30th ACM International Conference on Information and Knowledge Management}.

\bibitem[{Jin et~al.(2025)Jin, Zeng, Yue, Yoon, Arik, Wang, Zamani, and Han}]{search_r1}
Bowen Jin, Hansi Zeng, Zhenrui Yue, Jinsung Yoon, Sercan Arik, Dong Wang, Hamed Zamani, and Jiawei Han. 2025.
\newblock Search-r1: Training llms to reason and leverage search engines with reinforcement learning.
\newblock \emph{arXiv preprint arXiv:2503.09516}.

\bibitem[{Jin et~al.(2024)Jin, Cao, Chen, Liu, Jiang, Xu, Qiuxia, and Zhao}]{conflict}
Zhuoran Jin, Pengfei Cao, Yubo Chen, Kang Liu, Xiaojian Jiang, Jiexin Xu, Li~Qiuxia, and Jun Zhao. 2024.
\newblock \href {https://aclanthology.org/2024.lrec-main.1466} {Tug-of-war between knowledge: Exploring and resolving knowledge conflicts in retrieval-augmented language models}.
\newblock In \emph{Proceedings of the 2024 Joint International Conference on Computational Linguistics, Language Resources and Evaluation (LREC-COLING 2024)}.

\bibitem[{Jina(2024)}]{jina}
Jina. 2024.
\newblock Jina reranker v2 for agentic rag: Ultra-fast, multilingual, function-calling and code search.
\newblock \url{https://jina.ai/news/jina-reranker-v2-for-agentic-rag-ultra-fast-multilingual-function-calling-and-code-search/}.
\newblock Accessed: 2024-11-26.

\bibitem[{Kandpal et~al.(2023)Kandpal, Deng, Roberts, Wallace, and Raffel}]{kandpal2023large}
Nikhil Kandpal, Haikang Deng, Adam Roberts, Eric Wallace, and Colin Raffel. 2023.
\newblock Large language models struggle to learn long-tail knowledge.
\newblock In \emph{International Conference on Machine Learning}, pages 15696--15707. PMLR.

\bibitem[{Karpukhin et~al.(2020)Karpukhin, Oguz, Min, Lewis, Wu, Edunov, Chen, and Yih}]{dpr}
Vladimir Karpukhin, Barlas Oguz, Sewon Min, Patrick Lewis, Ledell Wu, Sergey Edunov, Danqi Chen, and Wen-tau Yih. 2020.
\newblock \href {https://aclanthology.org/2020.emnlp-main.550} {Dense passage retrieval for open-domain question answering}.
\newblock In \emph{Proceedings of the 2020 Conference on Empirical Methods in Natural Language Processing (EMNLP)}.

\bibitem[{Kasai et~al.(2024)Kasai, Sakaguchi, Takahashi, Bras, Asai, Yu, Radev, Smith, Choi, and Inui}]{realtimeqa}
Jungo Kasai, Keisuke Sakaguchi, Yoichi Takahashi, Ronan~Le Bras, Akari Asai, Xinyan Yu, Dragomir Radev, Noah~A. Smith, Yejin Choi, and Kentaro Inui. 2024.
\newblock \href {https://arxiv.org/abs/2207.13332} {Realtime qa: What's the answer right now?}

\bibitem[{Kau et~al.(2024)Kau, He, Nambissan, Astudillo, Yin, and Aryani}]{kg}
Amanda Kau, Xuzeng He, Aishwarya Nambissan, Aland Astudillo, Hui Yin, and Amir Aryani. 2024.
\newblock \href {https://arxiv.org/abs/2407.06564} {Combining knowledge graphs and large language models}.

\bibitem[{Khattab and Zaharia(2020)}]{ColBERT}
Omar Khattab and Matei Zaharia. 2020.
\newblock \href {https://doi.org/10.1145/3397271.3401075} {Colbert: Efficient and effective passage search via contextualized late interaction over bert}.
\newblock In \emph{Proceedings of the 43rd International ACM SIGIR Conference on Research and Development in Information Retrieval}.

\bibitem[{Lee et~al.(2024{\natexlab{a}})Lee, Roy, Xu, Raiman, Shoeybi, Catanzaro, and Ping}]{nv_embed}
Chankyu Lee, Rajarshi Roy, Mengyao Xu, Jonathan Raiman, Mohammad Shoeybi, Bryan Catanzaro, and Wei Ping. 2024{\natexlab{a}}.
\newblock \href {https://openreview.net/forum?id=lgsyLSsDRe} {Nv-embed: Improved techniques for training llms as generalist embedding models}.

\bibitem[{Lee et~al.(2024{\natexlab{b}})Lee, Chen, Dai, Dua, Sachan, Boratko, Luan, Arnold, Perot, Dalmia, Hu, Lin, Pasupat, Amini, Cole, Riedel, Naim, Chang, and Guu}]{longcontext}
Jinhyuk Lee, Anthony Chen, Zhuyun Dai, Dheeru Dua, Devendra~Singh Sachan, Michael Boratko, Yi~Luan, Sébastien M.~R. Arnold, Vincent Perot, Siddharth Dalmia, Hexiang Hu, Xudong Lin, Panupong Pasupat, Aida Amini, Jeremy~R. Cole, Sebastian Riedel, Iftekhar Naim, Ming-Wei Chang, and Kelvin Guu. 2024{\natexlab{b}}.
\newblock \href {https://arxiv.org/abs/2406.13121} {Can long-context language models subsume retrieval, rag, sql, and more?}

\bibitem[{Li et~al.(2023)Li, Liu, Xiao, and Shao}]{bge_gemma}
Chaofan Li, Zheng Liu, Shitao Xiao, and Yingxia Shao. 2023.
\newblock \href {https://arxiv.org/abs/2312.15503} {Making large language models a better foundation for dense retrieval}.

\bibitem[{Li et~al.(2025)Li, Dong, Jin, Zhang, Zhou, Zhu, Zhang, and Dou}]{search_o1}
Xiaoxi Li, Guanting Dong, Jiajie Jin, Yuyao Zhang, Yujia Zhou, Yutao Zhu, Peitian Zhang, and Zhicheng Dou. 2025.
\newblock \href {https://doi.org/10.48550/arXiv.2501.05366} {Search-o1: Agentic search-enhanced large reasoning models}.
\newblock \emph{CoRR}.

\bibitem[{Li{\v{s}}ka et~al.(2022)Li{\v{s}}ka, Ko{\v{c}}isk{\'y}, Gribovskaya, Terzi, Sezener, Agrawal, de~Masson~d'Autume, Scholtes, Zaheer, Young, Austin, Blunsom, and Lazaridou}]{streamingqa2022}
Adam Li{\v{s}}ka, Tom{\'a}{\v{s}} Ko{\v{c}}isk{\'y}, Elena Gribovskaya, Tayfun Terzi, Eren Sezener, Devang Agrawal, Cyprien de~Masson~d'Autume, Tim Scholtes, Manzil Zaheer, Susannah Young, Ellen Gilsenan-McMahon~Sophia Austin, Phil Blunsom, and Angeliki Lazaridou. 2022.
\newblock \href {https://proceedings.mlr.press/v162/liska22a/liska22a.pdf} {Streamingqa: A benchmark for adaptation to new knowledge over time in question answering models}.
\newblock \emph{arXiv preprint arXiv:2205.11388}.

\bibitem[{Liu et~al.(2024)Liu, Lin, Hewitt, Paranjape, Bevilacqua, Petroni, and Liang}]{lostinmiddle}
Nelson~F Liu, Kevin Lin, John Hewitt, Ashwin Paranjape, Michele Bevilacqua, Fabio Petroni, and Percy Liang. 2024.
\newblock \href {https://aclanthology.org/2024.tacl-1.9} {Lost in the middle: How language models use long contexts}.
\newblock \emph{Transactions of the Association for Computational Linguistics}.

\bibitem[{Liu et~al.(2023)Liu, Liang, Li, Giunchiglia, Li, Wang, Wu, Huang, Feng, and Guan}]{local}
Yonghao Liu, Di~Liang, Mengyu Li, Fausto Giunchiglia, Ximing Li, Sirui Wang, Wei Wu, Lan Huang, Xiaoyue Feng, and Renchu Guan. 2023.
\newblock \href {https://doi.org/10.24963/ijcai.2023/571} {Local and global: temporal question answering via information fusion}.
\newblock In \emph{Proceedings of the Thirty-Second International Joint Conference on Artificial Intelligence}.

\bibitem[{MacAvaney et~al.(2020)MacAvaney, Nardini, Perego, Tonellotto, Goharian, and Frieder}]{bm25}
Sean MacAvaney, Franco~Maria Nardini, Raffaele Perego, Nicola Tonellotto, Nazli Goharian, and Ophir Frieder. 2020.
\newblock \href {https://dl.acm.org/doi/abs/10.1145/3397271.3401093} {Efficient document re-ranking for transformers by precomputing term representations}.
\newblock In \emph{Proceedings of the 43rd International ACM SIGIR Conference on Research and Development in Information Retrieval}.

\bibitem[{Muennighoff et~al.(2023)Muennighoff, Tazi, Magne, and Reimers}]{mteb}
Niklas Muennighoff, Nouamane Tazi, Loic Magne, and Nils Reimers. 2023.
\newblock \href {https://aclanthology.org/2023.eacl-main.148} {{MTEB}: Massive text embedding benchmark}.
\newblock In \emph{Proceedings of the 17th Conference of the European Chapter of the Association for Computational Linguistics}.

\bibitem[{OpenAI et~al.(2024)OpenAI, Achiam, Adler, Agarwal, Ahmad, Akkaya, Aleman, Almeida, Altenschmidt, Altman, Anadkat, Avila, Babuschkin, Balaji, Balcom, Baltescu, Bao, Bavarian, Belgum, Bello, Berdine, Bernadett-Shapiro, Berner, Bogdonoff, Boiko, Boyd, Brakman, Brockman, Brooks, Brundage, Button, Cai, Campbell, Cann, Carey, Carlson, Carmichael, Chan, Chang, Chantzis, Chen, Chen, Chen, Chen, Chen, Chess, Cho, Chu, Chung, Cummings, Currier, Dai, Decareaux, Degry, Deutsch, Deville, Dhar, Dohan, Dowling, Dunning, Ecoffet, Eleti, Eloundou, Farhi, Fedus, Felix, Fishman, Forte, Fulford, Gao, Georges, Gibson, Goel, Gogineni, Goh, Gontijo-Lopes, Gordon, Grafstein, Gray, Greene, Gross, Gu, Guo, Hallacy, Han, Harris, He, Heaton, Heidecke, Hesse, Hickey, Hickey, Hoeschele, Houghton, Hsu, Hu, Hu, Huizinga, Jain, Jain, Jang, Jiang, Jiang, Jin, Jin, Jomoto, Jonn, Jun, Kaftan, Łukasz Kaiser, Kamali, Kanitscheider, Keskar, Khan, Kilpatrick, Kim, Kim, Kim, Kirchner, Kiros, Knight, Kokotajlo, Łukasz Kondraciuk,
  Kondrich, Konstantinidis, Kosic, Krueger, Kuo, Lampe, Lan, Lee, Leike, Leung, Levy, Li, Lim, Lin, Lin, Litwin, Lopez, Lowe, Lue, Makanju, Malfacini, Manning, Markov, Markovski, Martin, Mayer, Mayne, McGrew, McKinney, McLeavey, McMillan, McNeil, Medina, Mehta, Menick, Metz, Mishchenko, Mishkin, Monaco, Morikawa, Mossing, Mu, Murati, Murk, Mély, Nair, Nakano, Nayak, Neelakantan, Ngo, Noh, Ouyang, O'Keefe, Pachocki, Paino, Palermo, Pantuliano, Parascandolo, Parish, Parparita, Passos, Pavlov, Peng, Perelman, de~Avila Belbute~Peres, Petrov, de~Oliveira~Pinto, Michael, Pokorny, Pokrass, Pong, Powell, Power, Power, Proehl, Puri, Radford, Rae, Ramesh, Raymond, Real, Rimbach, Ross, Rotsted, Roussez, Ryder, Saltarelli, Sanders, Santurkar, Sastry, Schmidt, Schnurr, Schulman, Selsam, Sheppard, Sherbakov, Shieh, Shoker, Shyam, Sidor, Sigler, Simens, Sitkin, Slama, Sohl, Sokolowsky, Song, Staudacher, Such, Summers, Sutskever, Tang, Tezak, Thompson, Tillet, Tootoonchian, Tseng, Tuggle, Turley, Tworek, Uribe, Vallone,
  Vijayvergiya, Voss, Wainwright, Wang, Wang, Wang, Ward, Wei, Weinmann, Welihinda, Welinder, Weng, Weng, Wiethoff, Willner, Winter, Wolrich, Wong, Workman, Wu, Wu, Wu, Xiao, Xu, Yoo, Yu, Yuan, Zaremba, Zellers, Zhang, Zhang, Zhao, Zheng, Zhuang, Zhuk, and Zoph}]{gpt}
OpenAI, Josh Achiam, Steven Adler, Sandhini Agarwal, Lama Ahmad, Ilge Akkaya, Florencia~Leoni Aleman, Diogo Almeida, Janko Altenschmidt, Sam Altman, Shyamal Anadkat, Red Avila, Igor Babuschkin, Suchir Balaji, Valerie Balcom, Paul Baltescu, Haiming Bao, Mohammad Bavarian, Jeff Belgum, Irwan Bello, Jake Berdine, Gabriel Bernadett-Shapiro, Christopher Berner, Lenny Bogdonoff, Oleg Boiko, Madelaine Boyd, Anna-Luisa Brakman, Greg Brockman, Tim Brooks, Miles Brundage, Kevin Button, Trevor Cai, Rosie Campbell, Andrew Cann, Brittany Carey, Chelsea Carlson, Rory Carmichael, Brooke Chan, Che Chang, Fotis Chantzis, Derek Chen, Sully Chen, Ruby Chen, Jason Chen, Mark Chen, Ben Chess, Chester Cho, Casey Chu, Hyung~Won Chung, Dave Cummings, Jeremiah Currier, Yunxing Dai, Cory Decareaux, Thomas Degry, Noah Deutsch, Damien Deville, Arka Dhar, David Dohan, Steve Dowling, Sheila Dunning, Adrien Ecoffet, Atty Eleti, Tyna Eloundou, David Farhi, Liam Fedus, Niko Felix, Simón~Posada Fishman, Juston Forte, Isabella Fulford, Leo
  Gao, Elie Georges, Christian Gibson, Vik Goel, Tarun Gogineni, Gabriel Goh, Rapha Gontijo-Lopes, Jonathan Gordon, Morgan Grafstein, Scott Gray, Ryan Greene, Joshua Gross, Shixiang~Shane Gu, Yufei Guo, Chris Hallacy, Jesse Han, Jeff Harris, Yuchen He, Mike Heaton, Johannes Heidecke, Chris Hesse, Alan Hickey, Wade Hickey, Peter Hoeschele, Brandon Houghton, Kenny Hsu, Shengli Hu, Xin Hu, Joost Huizinga, Shantanu Jain, Shawn Jain, Joanne Jang, Angela Jiang, Roger Jiang, Haozhun Jin, Denny Jin, Shino Jomoto, Billie Jonn, Heewoo Jun, Tomer Kaftan, Łukasz Kaiser, Ali Kamali, Ingmar Kanitscheider, Nitish~Shirish Keskar, Tabarak Khan, Logan Kilpatrick, Jong~Wook Kim, Christina Kim, Yongjik Kim, Jan~Hendrik Kirchner, Jamie Kiros, Matt Knight, Daniel Kokotajlo, Łukasz Kondraciuk, Andrew Kondrich, Aris Konstantinidis, Kyle Kosic, Gretchen Krueger, Vishal Kuo, Michael Lampe, Ikai Lan, Teddy Lee, Jan Leike, Jade Leung, Daniel Levy, Chak~Ming Li, Rachel Lim, Molly Lin, Stephanie Lin, Mateusz Litwin, Theresa Lopez, Ryan
  Lowe, Patricia Lue, Anna Makanju, Kim Malfacini, Sam Manning, Todor Markov, Yaniv Markovski, Bianca Martin, Katie Mayer, Andrew Mayne, Bob McGrew, Scott~Mayer McKinney, Christine McLeavey, Paul McMillan, Jake McNeil, David Medina, Aalok Mehta, Jacob Menick, Luke Metz, Andrey Mishchenko, Pamela Mishkin, Vinnie Monaco, Evan Morikawa, Daniel Mossing, Tong Mu, Mira Murati, Oleg Murk, David Mély, Ashvin Nair, Reiichiro Nakano, Rajeev Nayak, Arvind Neelakantan, Richard Ngo, Hyeonwoo Noh, Long Ouyang, Cullen O'Keefe, Jakub Pachocki, Alex Paino, Joe Palermo, Ashley Pantuliano, Giambattista Parascandolo, Joel Parish, Emy Parparita, Alex Passos, Mikhail Pavlov, Andrew Peng, Adam Perelman, Filipe de~Avila Belbute~Peres, Michael Petrov, Henrique~Ponde de~Oliveira~Pinto, Michael, Pokorny, Michelle Pokrass, Vitchyr~H. Pong, Tolly Powell, Alethea Power, Boris Power, Elizabeth Proehl, Raul Puri, Alec Radford, Jack Rae, Aditya Ramesh, Cameron Raymond, Francis Real, Kendra Rimbach, Carl Ross, Bob Rotsted, Henri Roussez,
  Nick Ryder, Mario Saltarelli, Ted Sanders, Shibani Santurkar, Girish Sastry, Heather Schmidt, David Schnurr, John Schulman, Daniel Selsam, Kyla Sheppard, Toki Sherbakov, Jessica Shieh, Sarah Shoker, Pranav Shyam, Szymon Sidor, Eric Sigler, Maddie Simens, Jordan Sitkin, Katarina Slama, Ian Sohl, Benjamin Sokolowsky, Yang Song, Natalie Staudacher, Felipe~Petroski Such, Natalie Summers, Ilya Sutskever, Jie Tang, Nikolas Tezak, Madeleine~B. Thompson, Phil Tillet, Amin Tootoonchian, Elizabeth Tseng, Preston Tuggle, Nick Turley, Jerry Tworek, Juan Felipe~Cerón Uribe, Andrea Vallone, Arun Vijayvergiya, Chelsea Voss, Carroll Wainwright, Justin~Jay Wang, Alvin Wang, Ben Wang, Jonathan Ward, Jason Wei, CJ~Weinmann, Akila Welihinda, Peter Welinder, Jiayi Weng, Lilian Weng, Matt Wiethoff, Dave Willner, Clemens Winter, Samuel Wolrich, Hannah Wong, Lauren Workman, Sherwin Wu, Jeff Wu, Michael Wu, Kai Xiao, Tao Xu, Sarah Yoo, Kevin Yu, Qiming Yuan, Wojciech Zaremba, Rowan Zellers, Chong Zhang, Marvin Zhang, Shengjia
  Zhao, Tianhao Zheng, Juntang Zhuang, William Zhuk, and Barret Zoph. 2024.
\newblock \href {https://arxiv.org/abs/2303.08774} {Gpt-4 technical report}.

\bibitem[{Overwijk et~al.(2022)Overwijk, Xiong, and Callan}]{ClueWeb}
Arnold Overwijk, Chenyan Xiong, and Jamie Callan. 2022.
\newblock \href {https://doi.org/10.1145/3477495.3536321} {Clueweb22: 10 billion web documents with rich information}.
\newblock In \emph{Proceedings of the 45th International ACM SIGIR Conference on Research and Development in Information Retrieval}.

\bibitem[{Pageviews(2024)}]{pageview}
Pageviews. 2024.
\newblock Pageviews analysis.
\newblock \url{https://pageviews.wmcloud.org/}.
\newblock Accessed: 2024-11-26.

\bibitem[{Pham et~al.(2024)Pham, Ngo, Luu, and Nguyen}]{whos}
Quang~Hieu Pham, Hoang Ngo, Anh~Tuan Luu, and Dat~Quoc Nguyen. 2024.
\newblock \href {https://aclanthology.org/2024.findings-emnlp.593/} {Who`s who: Large language models meet knowledge conflicts in practice}.
\newblock In \emph{Findings of the Association for Computational Linguistics: EMNLP 2024}.

\bibitem[{Qian et~al.(2024)Qian, Zhang, Zhao, Zhou, Sui, Zhang, and Song}]{timer4}
Xinying Qian, Ying Zhang, Yu~Zhao, Baohang Zhou, Xuhui Sui, Li~Zhang, and Kehui Song. 2024.
\newblock \href {https://aclanthology.org/2024.emnlp-main.394/} {Timer4: Time-aware retrieval-augmented large language models for temporal knowledge graph question answering}.
\newblock In \emph{Proceedings of the 2024 Conference on Empirical Methods in Natural Language Processing}.

\bibitem[{Radford et~al.(2019)Radford, Wu, Child, Luan, Amodei, and Sutskever}]{radford2019language}
Alec Radford, Jeff Wu, Rewon Child, David Luan, Dario Amodei, and Ilya Sutskever. 2019.
\newblock Language models are unsupervised multitask learners.

\bibitem[{Rozner et~al.(2024)Rozner, Battash, Wolf, and Lindenbaum}]{knowledge}
Amit Rozner, Barak Battash, Lior Wolf, and Ofir Lindenbaum. 2024.
\newblock \href {https://aclanthology.org/2024.findings-emnlp.273} {Knowledge editing in language models via adapted direct preference optimization}.
\newblock In \emph{Findings of the Association for Computational Linguistics: EMNLP 2024}.

\bibitem[{Sandhaus(2008)}]{nyt}
Evan Sandhaus. 2008.
\newblock The new york times annotated corpus.
\newblock \emph{Linguistic Data Consortium, Philadelphia}.

\bibitem[{Saxena et~al.(2021)Saxena, Chakrabarti, and Talukdar}]{cronquestions}
Apoorv Saxena, Soumen Chakrabarti, and Partha Talukdar. 2021.
\newblock \href {https://github.com/apoorvumang/CronKGQA} {Question answering over temporal knowledge graphs}.
\newblock In \emph{Proceedings of the 59th Annual Meeting of the Association for Computational Linguistics}.

\bibitem[{Sharma et~al.(2023)Sharma, Saxena, Gupta, Kazemi, Talukdar, and Chakrabarti}]{twi}
Aditya Sharma, Apoorv Saxena, Chitrank Gupta, Mehran Kazemi, Partha Talukdar, and Soumen Chakrabarti. 2023.
\newblock \href {https://aclanthology.org/2023.eacl-main.150/} {{T}wi{RGCN}: Temporally weighted graph convolution for question answering over temporal knowledge graphs}.
\newblock In \emph{Proceedings of the 17th Conference of the European Chapter of the Association for Computational Linguistics}.

\bibitem[{Song et~al.(2025)Song, Jiang, Min, Chen, Chen, Zhao, Wen, Lu, and Miu}]{r1_searcher}
Huatong Song, Jinhao Jiang, Yingqian Min, Jie Chen, Zhipeng Chen, Wayne~Xin Zhao, Ji-Rong Wen, Yang Lu, and Xu~Miu. 2025.
\newblock R1-searcher: Incentivizing the search capability in llms via reinforcement learning.
\newblock \url{https://github.com/RUCAIBox/R1-searcher}.
\newblock Accessed: 2025-05-19.

\bibitem[{Su et~al.(2024{\natexlab{a}})Su, Yen, Xia, Shi, Muennighoff, yu~Wang, Liu, Shi, Siegel, Tang, Sun, Yoon, Arik, Chen, and Yu}]{bright}
Hongjin Su, Howard Yen, Mengzhou Xia, Weijia Shi, Niklas Muennighoff, Han yu~Wang, Haisu Liu, Quan Shi, Zachary~S. Siegel, Michael Tang, Ruoxi Sun, Jinsung Yoon, Sercan~O. Arik, Danqi Chen, and Tao Yu. 2024{\natexlab{a}}.
\newblock \href {https://arxiv.org/abs/2407.12883} {Bright: A realistic and challenging benchmark for reasoning-intensive retrieval}.

\bibitem[{Su et~al.(2024{\natexlab{b}})Su, Zhang, Zhu, Qu, Li, Zhang, and Cheng}]{timo}
Zhaochen Su, Jun Zhang, Tong Zhu, Xiaoye Qu, Juntao Li, Min Zhang, and Yu~Cheng. 2024{\natexlab{b}}.
\newblock Timo: Towards better temporal reasoning for language models.
\newblock \emph{Proceedings of the Conference on Language Model}.

\bibitem[{Tan et~al.(2023)Tan, Ng, and Bing}]{tempreason}
Qingyu Tan, Hwee~Tou Ng, and Lidong Bing. 2023.
\newblock \href {https://doi.org/10.18653/v1/2023.acl-long.828} {Towards benchmarking and improving the temporal reasoning capability of large language models}.
\newblock In \emph{Proceedings of the 61st Annual Meeting of the Association for Computational Linguistics}.

\bibitem[{Thakur et~al.(2021)Thakur, Reimers, Daxenberger, and Gurevych}]{SBERT}
Nandan Thakur, Nils Reimers, Johannes Daxenberger, and Iryna Gurevych. 2021.
\newblock \href {https://www.aclweb.org/anthology/2021.naacl-main.28} {Augmented {SBERT}: Data augmentation method for improving bi-encoders for pairwise sentence scoring tasks}.
\newblock In \emph{Proceedings of the 2021 Conference of the North American Chapter of the Association for Computational Linguistics: Human Language Technologies}.

\bibitem[{Virgo et~al.(2022)Virgo, Cheng, and Kurohashi}]{durationqa}
Felix Virgo, Fei Cheng, and Sadao Kurohashi. 2022.
\newblock \href {https://aclanthology.org/2022.lrec-1.473} {Improving event duration question answering by leveraging existing temporal information extraction data}.
\newblock In \emph{Proceedings of the Language Resources and Evaluation Conference}.

\bibitem[{Wang et~al.(2024)Wang, Zhang, Tian, Xi, Yao, Xu, Wang, Mao, Wang, Cheng, Liu, Ni, Zheng, and Chen}]{easyedit}
Peng Wang, Ningyu Zhang, Bozhong Tian, Zekun Xi, Yunzhi Yao, Ziwen Xu, Mengru Wang, Shengyu Mao, Xiaohan Wang, Siyuan Cheng, Kangwei Liu, Yuansheng Ni, Guozhou Zheng, and Huajun Chen. 2024.
\newblock \href {https://aclanthology.org/2024.acl-demos.9} {{E}asy{E}dit: An easy-to-use knowledge editing framework for large language models}.
\newblock In \emph{Proceedings of the 62nd Annual Meeting of the Association for Computational Linguistics (Volume 3: System Demonstrations)}.

\bibitem[{Wang et~al.(2020)Wang, Wei, Dong, Bao, Yang, and Zhou}]{minilm_hg}
Wenhui Wang, Furu Wei, Li~Dong, Hangbo Bao, Nan Yang, and Ming Zhou. 2020.
\newblock \href {https://arxiv.org/abs/2002.10957} {Minilm: Deep self-attention distillation for task-agnostic compression of pre-trained transformers}.

\bibitem[{Wang and Zhao(2024)}]{tram}
Yuqing Wang and Yun Zhao. 2024.
\newblock \href {https://aclanthology.org/2024.findings-acl.382} {{TRAM}: Benchmarking temporal reasoning for large language models}.
\newblock In \emph{Findings of the Association for Computational Linguistics: ACL 2024}.

\bibitem[{Wei et~al.(2022)Wei, Wang, Schuurmans, Bosma, ichter, Xia, Chi, Le, and Zhou}]{cot}
Jason Wei, Xuezhi Wang, Dale Schuurmans, Maarten Bosma, brian ichter, Fei Xia, Ed~Chi, Quoc~V Le, and Denny Zhou. 2022.
\newblock \href {https://proceedings.neurips.cc/paper_files/paper/2022/file/9d5609613524ecf4f15af0f7b31abca4-Paper-Conference.pdf} {Chain-of-thought prompting elicits reasoning in large language models}.
\newblock In \emph{Advances in Neural Information Processing Systems}.

\bibitem[{Wei et~al.(2023)Wei, Su, Ma, Yu, Lei, Zhang, Zhao, and Liu}]{menatqa}
Yifan Wei, Yisong Su, Huanhuan Ma, Xiaoyan Yu, Fangyu Lei, Yuanzhe Zhang, Jun Zhao, and Kang Liu. 2023.
\newblock \href {https://aclanthology.org/2023.findings-emnlp.100} {{M}enat{QA}: A new dataset for testing the temporal comprehension and reasoning abilities of large language models}.
\newblock In \emph{Findings of the Association for Computational Linguistics: EMNLP 2023}.

\bibitem[{Wu et~al.(2024{\natexlab{a}})Wu, Pan, Wang, and Luu}]{akew}
Xiaobao Wu, Liangming Pan, William~Yang Wang, and Anh~Tuan Luu. 2024{\natexlab{a}}.
\newblock \href {https://aclanthology.org/2024.emnlp-main.843} {{AKEW}: Assessing knowledge editing in the wild}.
\newblock In \emph{Proceedings of the 2024 Conference on Empirical Methods in Natural Language Processing}.

\bibitem[{Wu et~al.(2024{\natexlab{b}})Wu, Pan, Xie, Zhou, Zhao, Ma, Du, Mao, Luu, and Wang}]{wu2024antileak}
Xiaobao Wu, Liangming Pan, Yuxi Xie, Ruiwen Zhou, Shuai Zhao, Yubo Ma, Mingzhe Du, Rui Mao, Anh~Tuan Luu, and William~Yang Wang. 2024{\natexlab{b}}.
\newblock \href {https://arxiv.org/abs/2412.13670} {Antileak-bench: Preventing data contamination by automatically constructing benchmarks with updated real-world knowledge}.
\newblock \emph{arXiv preprint arXiv:2412.13670}.

\bibitem[{Xiao et~al.(2023)Xiao, Liu, Zhang, and Muennighoff}]{bge_reranker}
Shitao Xiao, Zheng Liu, Peitian Zhang, and Niklas Muennighoff. 2023.
\newblock \href {https://arxiv.org/abs/2309.07597} {C-pack: Packaged resources to advance general chinese embedding}.

\bibitem[{Yan et~al.(2024)Yan, Gu, Zhu, and Ling}]{crag}
Shi-Qi Yan, Jia-Chen Gu, Yun Zhu, and Zhen-Hua Ling. 2024.
\newblock \href {https://arxiv.org/abs/2401.15884} {Corrective retrieval augmented generation}.

\bibitem[{Yang et~al.(2024)Yang, Yang, Hui, Zheng, Yu, Zhou, Li, Li, Liu, Huang, Dong, Wei, Lin, Tang, Wang, Yang, Tu, Zhang, Ma, Yang, Xu, Zhou, Bai, He, Lin, Dang, Lu, Chen, Yang, Li, Xue, Ni, Zhang, Wang, Peng, Men, Gao, Lin, Wang, Bai, Tan, Zhu, Li, Liu, Ge, Deng, Zhou, Ren, Zhang, Wei, Ren, Liu, Fan, Yao, Zhang, Wan, Chu, Liu, Cui, Zhang, Guo, and Fan}]{qwen}
An~Yang, Baosong Yang, Binyuan Hui, Bo~Zheng, Bowen Yu, Chang Zhou, Chengpeng Li, Chengyuan Li, Dayiheng Liu, Fei Huang, Guanting Dong, Haoran Wei, Huan Lin, Jialong Tang, Jialin Wang, Jian Yang, Jianhong Tu, Jianwei Zhang, Jianxin Ma, Jianxin Yang, Jin Xu, Jingren Zhou, Jinze Bai, Jinzheng He, Junyang Lin, Kai Dang, Keming Lu, Keqin Chen, Kexin Yang, Mei Li, Mingfeng Xue, Na~Ni, Pei Zhang, Peng Wang, Ru~Peng, Rui Men, Ruize Gao, Runji Lin, Shijie Wang, Shuai Bai, Sinan Tan, Tianhang Zhu, Tianhao Li, Tianyu Liu, Wenbin Ge, Xiaodong Deng, Xiaohuan Zhou, Xingzhang Ren, Xinyu Zhang, Xipin Wei, Xuancheng Ren, Xuejing Liu, Yang Fan, Yang Yao, Yichang Zhang, Yu~Wan, Yunfei Chu, Yuqiong Liu, Zeyu Cui, Zhenru Zhang, Zhifang Guo, and Zhihao Fan. 2024.
\newblock \href {https://arxiv.org/abs/2407.10671} {Qwen2 technical report}.

\bibitem[{Yang(2023)}]{small2big}
Sophia Yang. 2023.
\newblock Advanced rag 01: Small-to-big retrieval.
\newblock \url{https://towardsdatascience.com/advanced-rag-01-small-to-big-retrieval-172181b396d4}.
\newblock Accessed: 2024-09-15.

\bibitem[{Zhang and Choi(2021)}]{situatedqa}
Michael Zhang and Eunsol Choi. 2021.
\newblock \href {https://aclanthology.org/2021.emnlp-main.586} {{S}ituated{QA}: Incorporating extra-linguistic contexts into {QA}}.
\newblock In \emph{Proceedings of the 2021 Conference on Empirical Methods in Natural Language Processing}.

\bibitem[{Zhang et~al.(2024)Zhang, Luu, and Zhao}]{syntqa}
Siyue Zhang, Anh~Tuan Luu, and Chen Zhao. 2024.
\newblock \href {https://aclanthology.org/2024.findings-emnlp.131} {{S}yn{TQA}: Synergistic table-based question answering via mixture of text-to-{SQL} and {E}2{E} {TQA}}.
\newblock In \emph{Findings of the Association for Computational Linguistics: EMNLP 2024}.

\bibitem[{Zhao et~al.(2021)Zhao, Xiong, Boyd-Graber, and {Daum\'{e} III}}]{distant}
Chen Zhao, Chenyan Xiong, Jordan Boyd-Graber, and Hal {Daum\'{e} III}. 2021.
\newblock Distantly-supervised dense retrieval enables open-domain question answering without evidence annotation.
\newblock In \emph{Emperical Methods in Natural Language Processing}.

\end{thebibliography}

\clearpage
\appendix

\section{Controlled Experiments}
\label{control}

\begin{figure}[t!]
    \centering
    \includegraphics[width=0.48\textwidth]{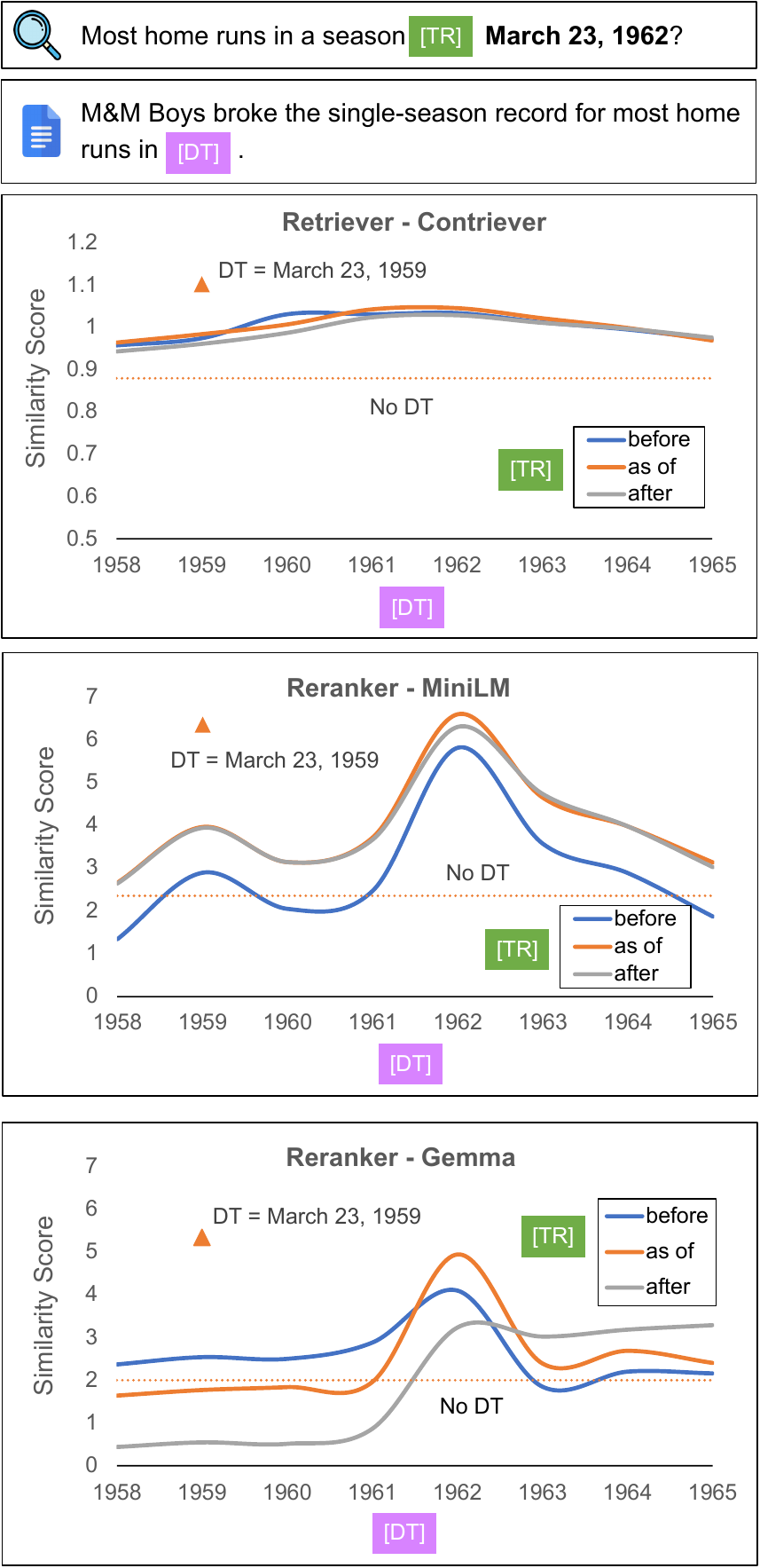}
\captionsetup{justification=justified,singlelinecheck=false}
    \caption{Similarity scores of query-document pairs by varying the temporal relation in the query and the date in the document.}
    \label{fig:control}
\end{figure}

We conduct controlled experiments to investigate the behaviors of retrieval methods on temporally constrained queries, including the bi-encoder retriever \contriever \citep{contriever}, the cross-encoder reranker MiniLM \citep{minilm_hg}, and the LLM embedding-based reranker \bgegemma \citep{bge_gemma}. As shown in \Cref{fig:control}, all methods prioritize date-matching, having the highest scores when the query and document share the same year. Besides, the score is unusually high when the query and document share the same month and day but differ in year, e.g., orange triangles in the diagrams.

The \contriever retriever is less sensitive to document dates than the \mini reranker. Both methods exhibit similar trends across varying temporal relations, indicating their inability to differentiate effectively between different relations, e.g., ``\texttt{before}'' and ``\texttt{after}''. Notably, documents without specific dates receive unusually low scores, even lower than those with irrelevant dates, e.g., orange dash lines in the diagrams.

The LLM embedding based reranker Gemma exhibits stronger temporal reasoning capabilities. For the ``\texttt{after}'' relation, documents with dates later than the query date are assigned relatively high and consistent scores. So all temporally relevant documents will be retained. However, for ``\texttt{before}'' and ``\texttt{as of}'', despite their temporal relevance, documents with earlier dates fail to achieve sufficiently high similarity scores, potentially leading to their exclusion from the retrieval process.

In summary, existing retrieval methods demonstrate limited temporal reasoning capabilities. The LLM embedding-based method shows better performance than others. Our proposed \mrag framework is retriever-agnostic, which aims to improve temporal reasoning capabilities for any type of retrieval models.

\section{Dataset Selection Criteria}
\label{exclusion}

Our benchmark focuses on time-sensitive question answering, which is knowledge-intensive. Therefore, datasets designed for only temporal reasoning (e.g., “What is the time 5 year and 5 month after Oct, 1444”) are not considered, such as TempReason \citep{tempreason} and DurationQA \citep{durationqa}. Aggregated benchmarks (\eg TimeBench \citep{timebench} and TRAM \citep{tram}) focus on evaluating diverse temporal reasoning capabilities, which have a broader scope than our focus. MenatQA \citep{menatqa} is built by adding counterfactual and order factors to TimeQA \citep{timeqa} questions, which is similar to our approach. Other knowledge-intensive temporal QA datasets can serve as alternative sample sources including StreamingQA \cite{streamingqa2022}, TempLAMA \citep{templama}, and concurrent dataset ComplexTQA \citep{complextempqa}. We select \situatedqa \citep{situatedqa} for its human-written questions, distinguishing it from other temporal QA datasets that are typically synthetic. Additionally, we opt for \timeqa \citep{timeqa} due to its hard split, which already includes complex temporal questions. Notably, both \situatedqa and \timeqa can be grounded in the Wikipedia corpus.

\section{Annotation Guidelines}
\label{guidelines}

\subsection{Annotating Perturbations}
Given a question-answer pair sourced from \timeqa or \situatedqa (\eg Q: ``Arnolfini Portrait was owned by whom between Jul 1842 and Nov 1842?'' A: ``National Gallery''), annotators should ground the pair to facts in Wikipedia (\eg ``The Arnolfini Wedding by Jan van Eyck, has been part of the National Gallery's collection in London since 1842.''). Then they identify the key timestamps or durations of Wikipedia facts (\eg 1842). To create temporal perturbations, annotators are asked to come up with combinations of implicit conditions, temporal relations, and alternative dates to form complex temporal constraints. The implicit condition can be ``\texttt{None}'' or selected from a list of 4 types: ``\texttt{first}'', ``\texttt{earliest}'', ``\texttt{last}'', and ``\texttt{latest}''. The temporal relation should be selected from a list of 11 types: ``\texttt{as of}'', ``\texttt{from to}'', ``\texttt{until}'', ``\texttt{before}'', ``\texttt{after}'', ``\texttt{around}'', ``\texttt{between}'', ``\texttt{by}'', ``\texttt{in}'', ``\texttt{on}'', and ``\texttt{since}''. Finally, annotators rewrite questions naturally (\eg ``Who is the last one owned Arnolfini Portrait after 1700?'') by introducing perturbed temporal constraints (\eg ``last ... after 1700'') and ensure that the answers (\eg ``National Gallery'') remain unchanged. After different annotators create perturbed question-answer pairs, they exchange these pairs with each other to validate the correctness of the answers. Only the perturbed samples validated by two annotators are kept.

\subsection{Annotating Gold Evidences}

For gold evidence annotations, annotators are assigned different perturbed question-answer pairs. For each pair, 20 context passages are provided to annotators, which are retrieved by the leading retriever \contriever \citep{contriever}, and the best reranker \bgegemma \citep{bge_gemma}. Annotators are asked to identify up to two gold evidence passages from these passages. A passage is regarded as relevant and annotated as gold evidence if annotators can obtain the correct answer from this passage. If there is no relevant passage among these 20 retrieved ones, annotators should search Wikipedia pages related to the query entities to locate the gold evidence passages manually. Lastly, annotators exchange samples to validate gold evidence annotated by others. Only the gold evidence annotations validated by two annotators are kept.

\section{Sample Statistics}
\label{statistics}

We gather a similar size of examples as previous temporal QA benchmarks (e.g., 3K for TimeQA \citep{timeqa} and 2K for MenatQA \citep{menatqa}), which is enough for an evaluation set (see examples in \Cref{examples}). As we manually annotate gold evidence passages in Wikipedia, it is time-consuming to scale up like other synthetic datasets \citep{templama, complextempqa}. To understand the difference between two subsets, we summarize the statistics in \Cref{statistics}. The average length
of questions is measured by the GPT-2 tokenizer \citep{radford2019language}. We assess the popularity of key entities in questions using the average monthly page view counts of the corresponding Wikipedia page in 2024 \citep{pageview}. As we can see, the main difference lies in the question entity popularity. \temprageval-\timeqa questions typically inquire about lesser-known individuals and are generally straightforward and clear. In contrast, \temprageval-\situatedqa questions commonly ask about a sports team and championship, which are more general and sometimes ambiguous. 
This difference may explain varying retrieval and QA performance across the two subsets.

\begin{table}[hbt!]
\centering
\renewcommand{\arraystretch}{1}
\begin{tabular}{p{3.4cm}cc} 
\toprule
                       & \textbf{TimeQA} & \textbf{SituatedQA} \\ 
\midrule
\# original questions  &  123 &   120     \\ 
\# perturbed questions & 377  &    380   \\ 
\# total questions     & 500&   500   \\ 
Temporal complexity   &  hard  &   hard  \\ 
Avg. question length   &  15.2   &   15.6 \\
Avg. entity popularity &     7,456   &  57,521     \\ 
\bottomrule
\end{tabular}
\captionsetup{justification=justified,singlelinecheck=false}
\caption{Sample statistics for \temprageval-\timeqa and \temprageval-\situatedqa. }
\label{table: statistics}
\end{table}

\begin{figure*}[h]
    \centering
    \includegraphics[width=0.7\textwidth]{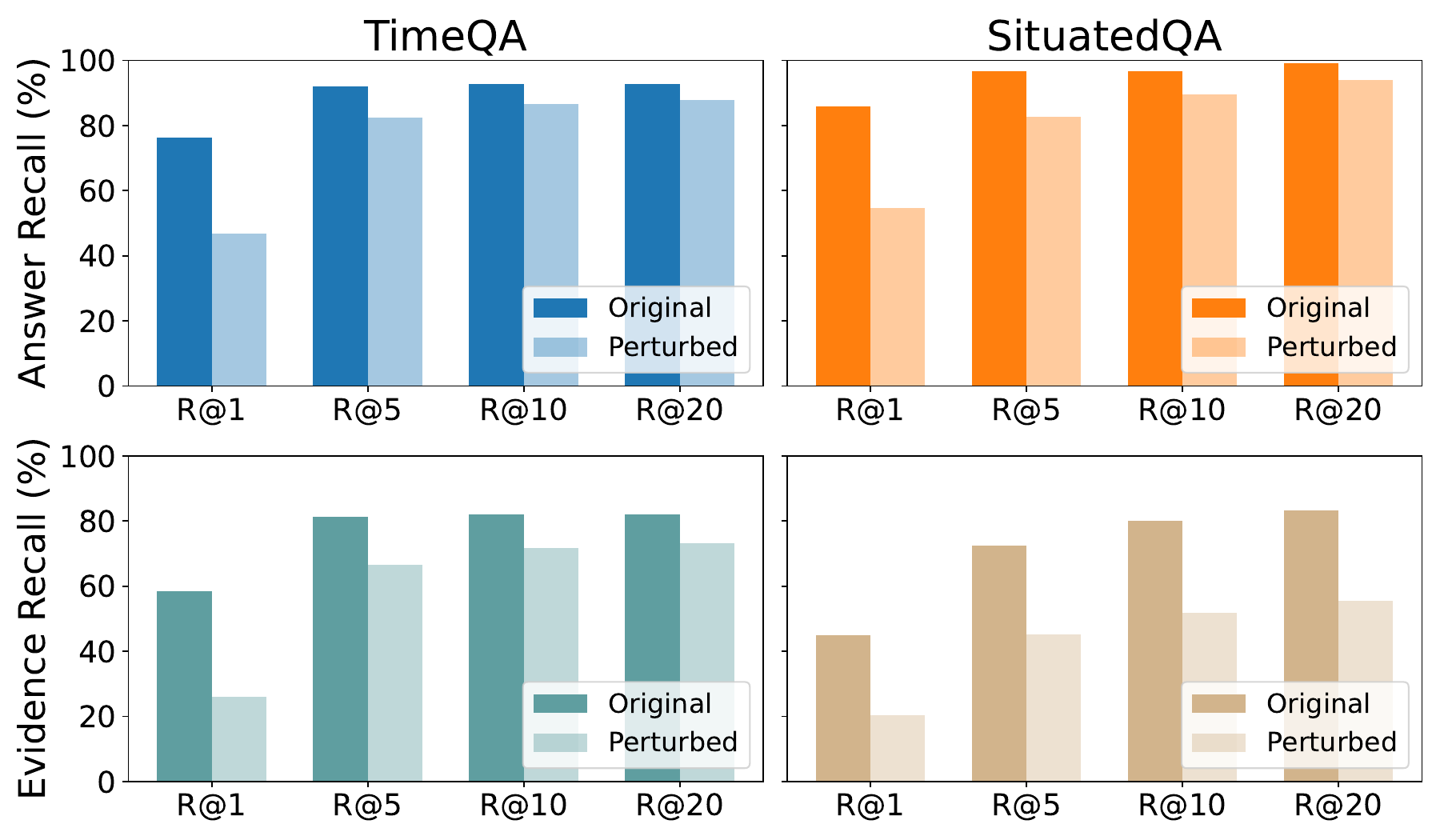}
\captionsetup{justification=justified,singlelinecheck=false}
    \caption{Retrieval performance difference between original queries and perturbed queries in \temprageval subsets for the baseline \bgegemma retrieval.}
    \label{fig:degrade}
\end{figure*}

\section{Retrieval Performance Degradation Due to Perturbations}
\label{sec: degrade}

As shown in \Cref{fig:degrade}, we compare the retrieval performance between original queries and perturbed queries using the same baseline retrieval system (\ie \contriever retriever and \bgegemma reranker). For both the \timeqa and \situatedqa subsets, the perturbed questions significantly increase the difficulty of retrieving relevant documents, particularly when evaluating the top-1 and top-5 ranked documents. This suggests that the introduction of perturbations introduces greater complexity. The existing retrieval method has limited temporal reasoning capabilities and is not robust to such variations.

\section{Evaluation Experiment Implementation Details}
\label{imp}

We conduct empirical evaluations for \mrag and SOTA retrieving-and-reranking systems on \temprageval. In baselines, due to limited computing resources, we use LLM-based embedding models as a reranking model, such as \bgegemma \citep{gemma} and NV-Embed \citep{nv_embed}. \mrag consists of functional modules, which can be based on algorithms, models, or prompting methods. In implementation, algorithm based modules include question normalization, keyword ranking, time extraction, and semantic-temporal hybrid ranking. Model based modules are retrieving, semantic ranking, sentence tokenization. To ensure fair comparison, we use the same retriever model, i.e., \contriever \citep{contriever}, as the first stage method for \mrag and two-stage systems. We use \bgegemma embeddings \citep{bge_gemma} as the main tool to measure semantic similarity for passages and sentences in \mrag. NLTK package is used for sentence tokenization \citep{nltk}. LLM prompting based modules are keyword extraction, query-focused summarization. As shown in \Cref{full_eval}, we have tested Llama3.1-8B-Instruct and Llama3.1-70B-Instruct models \citep{llama3} for LLM prompting based modules. Detailed prompts are listed in \Cref{sec: prompts}. The evaluation metrics are computed based on retrieved passages not sentences.

\begin{figure*}[hbt!]
    \centering
    \includegraphics[width=0.8\textwidth]{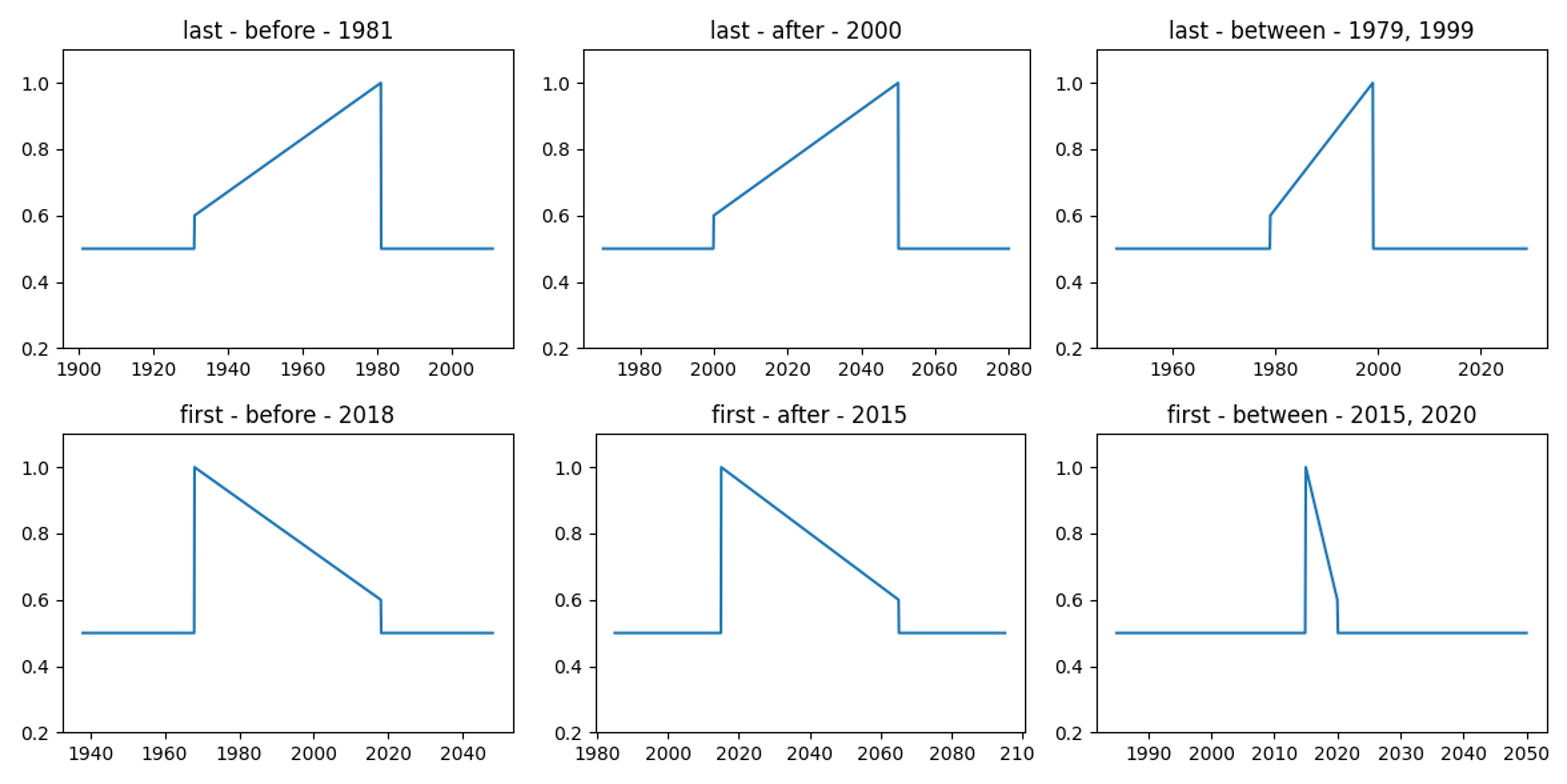}
\captionsetup{justification=justified,singlelinecheck=false}
    \caption{Pre-defined spline functions for temporal relevance scoring. The title of each subplot represents the type of query temporal constraint. The horizontal coordinate of each subplot is the date in the document sentence.}
    \label{fig:splines}
\end{figure*}

\section{Implementations for Semantic-Temporal Hybrid Scoring}
\label{spline}

The \textbf{Retrieval and Summarization} module in \mrag splits and summarizes top relevant passages into independent sentences for the downstream fine-grained reranking, which is inspired by \citet{small2big}. The \textbf{Semantic-Temporal Hybrid Scoring} module is designed to assess the semantic relevance and temporal relevance between the question and the evidence sentence. To quantify the semantic relevance, we apply an embedding model (\eg \bgegemma) to the question main content and the sentence.

For temporal relevance, we employ a symbolic scoring approach, wherein the module automatically generates scoring functions for each question and computes temporal scores for individual sentences.

To generate scoring functions, we classify question temporal constraints into six categories and define a template for each. A scoring function is instantiated using the corresponding template and extracted timestamp(s) from the question. The six constraint types include: ``\texttt{first - before}'', ``\texttt{first - after}'', ``\texttt{first - between}'', ``\texttt{last - before}'', ``\texttt{last - after}'', and ``\texttt{last - between}''. Here, ``\texttt{last}'' denotes that the question seeks the most recent event, while ``\texttt{before}'' indicates that the event must precede a specified date. For instance, the question “Who won the latest game as of 1981?” corresponds to ``\texttt{last - before - 1981}'', as illustrated in the top-left subplot of \Cref{fig:splines}.

Afterwards, the module extracts the timestamp(s) from the evidence sentence. It computes the temporal score based on the extracted timestamp and the corresponding scoring function. If multiple timestamps are present, the highest temporal score is selected. For example, as depicted in \Cref{fig:splines}, for the constraint ``\texttt{last - before - 1981}'', an evidence sentence mentioning ``\texttt{1970}'' would receive a temporal score around 0.9.

The final score of each evidence sentence for each question is obtained by multiplying the temporal score and the semantic score. Finally, we select the passages that contains the highest-scoring sentences. The passages are fed into the later answer generation stage rather than the sentences. The passages provide better background information, which leads to higher generation quality for reader systems.

\newpage
\section{LLM-based Summarization Case Study}
\label{summarize_errors}

LLM-based query-focused summarization enhances retrieval performance by distilling key information from passages while filtering out irrelevant context, as demonstrated in \Cref{sum_cases}. In the second success case, the summarization effectively converts structured data into natural language, benefiting retrievers that are primarily trained on free-text retrieval tasks. However, LLM-generated summaries may introduce hallucinations and errors, though their occurrence is infrequent. As shown in \Cref{sum_cases}, erroneous summaries can mislead the retriever with non-factual events or incorrect dates, resulting in irrelevant passages ranking higher. To balance retrieval improvements with the risk of errors, we summarize only the top-$k$ passages per query, which also reduces computational overhead. Furthermore, to prevent error propagation, we provide the reader model with original passages rather than their summaries.

\onecolumn

\begin{table*}[hbt!]
\centering
\renewcommand{\arraystretch}{1.1}
\begin{tabular}{p{2.8cm}p{12.2cm}} 
\toprule
   & \textcolor{ForestGreen}{\textbf{Success Cases}}\\ 
\midrule
Question &  Who won the latest America's Next Top Model by May 8, 2021? \\ 
Answer &  Kyla Coleman \\ 
Passage &  America's Next Top Model (season 24) | The twenty-fourth cycle of America's Next Top Model premiered on January 9, 2018 ... The winner of the competition was 20 year-old Kyla Coleman from Lacey, Washington with Jeana Turner placing as the runner up.\\
Summarization & Kyla Coleman, a 20-year-old from Lacey, Washington, won the competition in 2018.
\\
\midrule
Question & When was the last time Kentucky won NCAA in basketball after 2010? \\ 
Answer & 2012 \\
Passage &  Kentucky Wildcats | Men (8) ; Basketball (8): 1948, 1949, 1951, 1958, 1978, 1996, 1998, 2012 ; Women (2) ... List of NCAA schools with the most NCAA Division ... Kentucky has won 13 NCAA team national championships.  \\
Summarization & The Kentucky Wildcats won the NCAA basketball championship in 1948, 1949, 1951, 1958, 1978, 1996, 1998, and 2012. \\
\midrule
   & \textcolor{red}{\textbf{Failure Cases}} \\
   \midrule
Question & When was the last time the Ducks won the Stanley Cup as of 2010? \\ 
Answer & 2007 \\
Passage &  Anaheim Ducks | ... Despite the arenas being six hours away from each other, the teams have developed a strong rivalry, primarily from the 2009 and 2018 Stanley Cup playoffs. The Ducks won the series in 2009, but the Sharks came back in 2018.
\\
Summarization & The Anaheim Ducks won the Stanley Cup in 2009. \\
True fact & The Anaheim Ducks won the Stanley Cup in 2007. That was their first and only championship so far. \\
   \midrule 
Question & How many times has South Korea held the Winter Olympics as of 2018? \\ 
Answer & 1 | one \\
Passage &  2018 Winter Olympics |  The 2018 Winter Olympics ... This marked the second time that South Korea had hosted the Olympic Games (having previously hosted the 1988 Summer Olympics in Seoul) ...\\
Summarization & South Korea held the Winter Olympics in 2018 and previously in 1988. \\
True fact & South Korea held the Winter Olympics in 2018 and Summer Olympics in 1988. \\
\bottomrule

\end{tabular}
\captionsetup{justification=justified,singlelinecheck=false}
\caption{Success and error cases of LLM-based query-focused summarization using Llama3.1-8B-Instruct.}
\label{sum_cases}
\end{table*}

\twocolumn

\section{Complete Retrieval Evaluation Results}
\label{full_eval}

We evaluate \mrag on \temprageval with baseline retrieval methods, including ELECTRA\footnote{cross-encoder/ms-marco-electra-base}, \mini\footnote{cross-encoder/ms-marco-MiniLM-L-12-v2}, Jina\footnote{jinaai/jina-reranker-v2-base-multilingual}, BGE\footnote{BAAI/bge-reranker-large}, NV-Embed\footnote{nvidia/NV-Embed-v1}, and \bgegemma\footnote{BAAI/bge-reranker-v2-gemma}. Complete rsults are presented in \Cref{experiments_timeqa} and \Cref{experiments_situatedqa}.

\begin{table*}[h!]
\centering
\renewcommand{\arraystretch}{1.1}
\arrayrulecolor{black}
\begin{tabular}{cccc!{\color{black}\vrule}cccc!{\color{black}\vrule}cccc!} 
\arrayrulecolor{black}
\toprule
\multicolumn{4}{c|}{\textbf{Method}}     & \multicolumn{4}{c|}{\textbf{Answer Recall @}} & \multicolumn{4}{c}{\textbf{Gold Evidence Recall @}}  \\ 
\hline
\textbf{1st}                  & \textbf{2nd}                  & \textbf{LLM}                  & \textbf{\# QFS} & \textbf{1}   & \textbf{5}    & \textbf{10}   & \textbf{20}                             & \textbf{1}    & \textbf{5}    & \textbf{10}   & \textbf{20}                                     \\ 
\hline
BM25  & -  & -   & -   &  17.5  &  39.0 & 49.1  &  59.0 & 4.2 &  14.1 &  22.6  & 33.7  \\ 
\arrayrulecolor{black}
Cont.     & -   & -   & -   & 18.8 & 49.9 & 62.1 & 72.9  & 9.6  & 28.7 & 39.5 & 51.5 \\
Hybrid      & -   & -   & -   & 18.8  & 51.2  &  65.0 & 75.3 & 9.6 & 28.1  & 41.1  &  55.2  \\
\hline
Cont.      & ELECTRA  & -     & -     &   40.1    &   76.9    &   83.6    &   86.7   &  21.8     &   58.6  &    66.8   &    71.6    \\ 
Cont.   & MiniLM    & -   & -     & 34.0 & 76.1 & 84.4 & 87.0  & 16.2 & 57.3 & 68.2 & 72.4  \\ 
Cont.   & Jina   & -   & -     & 42.4&77.2&86.2&87.5&23.6&58.6&68.2&71.4  \\ 
Cont.    & BGE     & -   & -   &  40.3 & 80.9 & 85.7 &  87.0   & 23.3 & 61.3 & 68.7 & 72.2 \\ 
Cont.      & NV-Embed  & -     & -     &   49.9    &  81.2   &  85.7  &    87.5   &  33.4  &  62.9 &   70.6  &  72.7      \\ 
Cont. & Gemma  & -  & -   & 46.7  & 82.5 & 86.5 & 87.8  & 26.0 & 66.6 & 71.6 & 73.2 \\ 
\hline
Cont.   & \mrag   & Llama3.1    & -     & 57.6 & 89.4 & 93.6  & 94.2 &  32.4 & 73.5 & 82.8 & 84.1                                   \\ 
Cont.   & \mrag   & Llama3.1   & 5   & 58.6 & 90.0 & 93.4 & 94.2 & 37.1 & 74.3 & 82.5 & 84.1  \\ 
Cont.   & \mrag   & Llama3.1   & 10   & 56.0  & 88.1  & 93.6 & 94.2 & 35.5  & 73.2 & 82.2 & 84.4  \\ 
\hline
Cont.    & \mrag  & Llama3.1$^{\flat}$    & -   &  57.6 & 89.4 & 93.6 & 94.2 & 32.1 & 73.5 & 82.5 & 84.1 \\
Cont.    & \mrag   & Llama3.1$^{\flat}$     & 5    &  57.0&90.5&93.6&94.2&34.5&75.3&82.5&84.1 \\
Cont.     & \mrag  & Llama3.1$^{\flat}$     & 10   & 53.3&90.7&93.6&94.2&33.2&74.8&82.8&84.1  \\
\arrayrulecolor{black}
\bottomrule
\end{tabular}
\captionsetup{justification=justified,singlelinecheck=false}
\caption{The answer recall (AR@k) and gold evidence recall (ER@k) performance of each retrieval system on perturbed temporal queries in \temprageval $-$ \timeqa subset. $^{\flat}$Meta-Llama-3.1-70B-Instruct.}
\label{experiments_timeqa}
\end{table*}

\begin{table*}[h!]
\centering
\renewcommand{\arraystretch}{1.1}
\arrayrulecolor{black}
\begin{tabular}{cccc!{\color{black}\vrule}cccc!{\color{black}\vrule}cccc!} 
\arrayrulecolor{black}
\toprule
\multicolumn{4}{c|}{\textbf{Method}}     & \multicolumn{4}{c|}{\textbf{Answer Recall @}} & \multicolumn{4}{c}{\textbf{Gold Evidence Recall @}}  \\ 
\hline
\textbf{1st}                  & \textbf{2nd}                  & \textbf{LLM}                  & \textbf{\# QFS} & \textbf{1}   & \textbf{5}    & \textbf{10}   & \textbf{20}                             & \textbf{1}    & \textbf{5}    & \textbf{10}   & \textbf{20}                                     \\ 
\hline
BM25      & -   & -   & -   & 27.6  &  58.2  &  69.0 &  80.8 & 6.8 & 18.4  &  25.8  & 34.7  \\ 
\arrayrulecolor{black}
Cont.     & -    & -     & -   & 22.6 & 51.1 & 65.5 & 79.5  &  6.8 &  17.1 & 22.9  & 30.5 \\ 
Hybrid   & -  & - & - & 22.6  & 55.8 & 71.8 & 81.6  & 6.8  & 19.7 & 26.6 & 35.0  \\
\hline
Cont.      & ELECTRA  & -     & -     &    35.5   &     71.3  &   82.4    & 88.4   &   15.3    &    37.1   &   45.0 & 52.9   \\ 
Cont.     & MiniLM   & -    & -   & 36.8 & 73.4 & 86.3 & 90.8  & 20.0 & 40.3 & 50.5 & 54.2  \\ 
Cont.     & Jina   & -    & -   &  47.9&78.4&87.6&93.2&19.5&41.1&48.2&54.2 \\
Cont.       & BGE   & -   & -  &  36.3 & 74.2 & 86.3 & 92.9 & 14.5 & 35.0 & 44.7 & 54.2 \\ 
Cont.      & NV-Embed    & -     & -     &    47.4   &   81.3  &   88.7    &  92.4     &  23.4    &    46.1   &    50.5   &    55.0 \\
Cont.       & Gemma      & -   & -  & 54.7 & 82.6 & 89.5 & 94.0 & 20.3 & 45.3 & 51.8 & 55.5 \\  
\hline
Cont.      & MRAG   & Llama3.1     & -     & 61.1 & 88.2 & 92.1 & 93.7  & 27.4 & 56.3 & 64.0 & 68.7  \\ 
Cont.   & MRAG   & Llama3.1  & 5   &  61.3 & 89.0 & 93.4 & 94.2 & 31.1 & 59.2 & 65.8 & 69.0 \\ 
Cont.  & MRAG   & Llama3.1  & 10   &  62.1 & 87.9 & 92.6 & 94.2 & 30.8 & 57.9 & 66.1 & 70.3 \\ 
\hline
Cont.    & \mrag  & Llama3.1$^{\flat}$    & -   & 61.1&86.3&92.4&94.0&27.1&54.5&63.4&67.6 \\
Cont.    & \mrag   &  Llama3.1$^{\flat}$     & 5    &   63.2 &86.1&92.6&93.7&29.7&56.6&64.0&67.1 \\
Cont.     & \mrag  & Llama3.1$^{\flat}$     & 10   &  62.1&86.8&92.1&93.7&27.1&56.1&64.2&69.0  \\

\arrayrulecolor{black}
\bottomrule
\end{tabular}
\captionsetup{justification=justified,singlelinecheck=false}
\caption{The answer recall (AR@k) and gold evidence recall (ER@k) performance of each retrieval system on perturbed temporal queries in \temprageval $-$ \situatedqa subset. $^{\flat}$Meta-Llama-3.1-70B-Instruct.}
\label{experiments_situatedqa}
\end{table*}

\section{Human Evaluation of Retrieval}
\label{sec: human}
\begin{figure}[t!]
    \centering
    \includegraphics[width=0.48\textwidth]{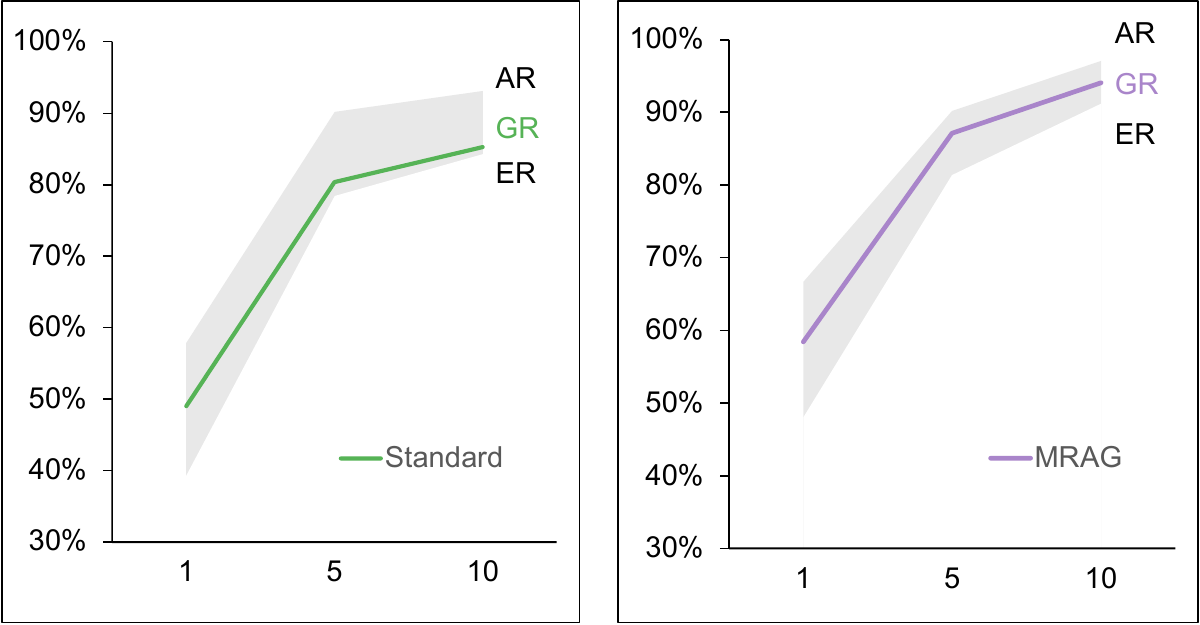}
    \captionsetup{justification=justified, singlelinecheck=false}  %
    \caption{Human annotated retrieval performance on 100 examples from \temprageval-\timeqa.}
    \label{fig:timeqa_human}
\end{figure}

\begin{figure}[t!]
    \centering
    \includegraphics[width=0.48\textwidth]{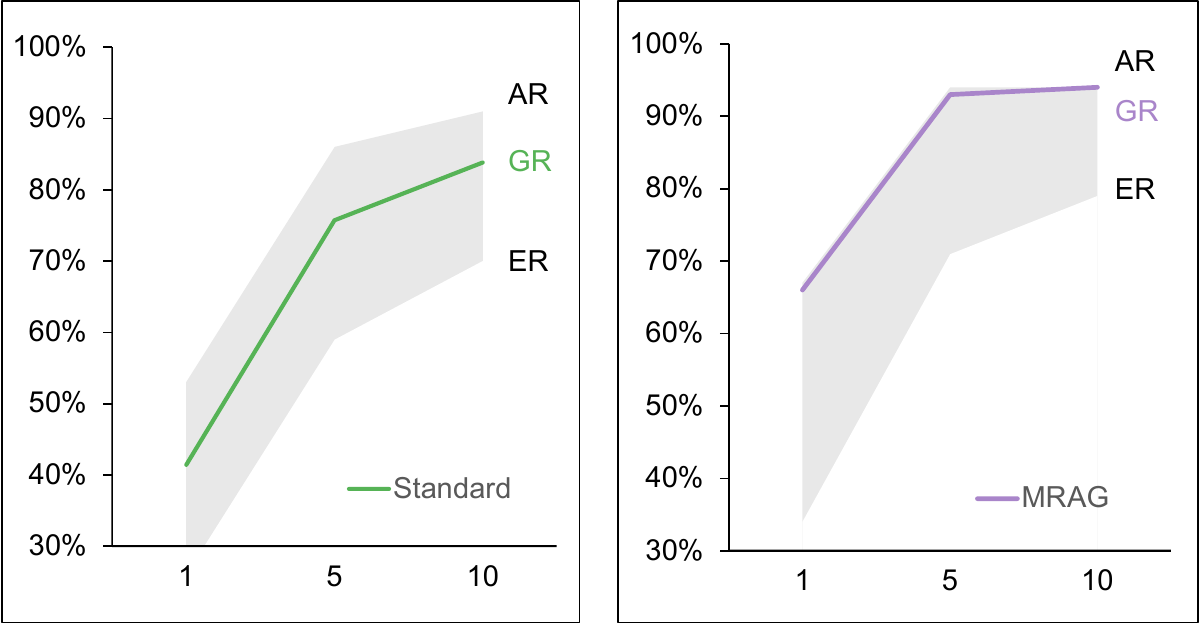}
    \captionsetup{justification=justified, singlelinecheck=false}  %
    \caption{Human annotated retrieval performance on 100 examples from \temprageval-\situatedqa.}
    \label{fig:situatedqa_human}
\end{figure}

The Answer Recall (AR@k) represents the upper bound of the retrieval performance, while the Evidence Recall (ER@k) signifies the lower bound. As shown in \Cref{fig:timeqa_human} and \Cref{fig:situatedqa_human}, the gray areas are delineated by the AR@k and ER@k lines, within which the actual performance remains uncertain. To address this, we conduct a human evaluation of ranked document passages retrieved by \mrag and \bgegemma (denoted as ``Standard'' in the figures) on a subset of 200 randomly selected examples from \temprageval.

The metric for the actual retrieval performance, termed \textbf{Ground Truth Recall (GR@k)}, is computed based on the annotations of the highest-ranking passages supporting the answers. As illustrated, the gray areas for \mrag are positioned higher in the plots than those for \bgegemma. Furthermore, the actual performance curves (purple lines) for \mrag are consistently closer to the upper boundaries compared to those for \bgegemma (green lines). These two observations demonstrate the superior performance of \mrag in temporal reasoning-intensive retrieval.

\newpage

\newpage
~\newpage
~\newpage
~\newpage

\section{Parametric Study on the Optimal Number of Passages for Concatenation}
\label{sec: concat}

The number of concatenated passages and their order significantly impact the accuracy of reader QA tasks. This is largely due to the inherent primacy and recency biases exhibited by LLMs, where information presented earlier or later in the input sequence tends to be weighted more heavily during processing \citep{lostinmiddle}. Therefore, the retrieval performance is of great importance. 

\begin{figure*}[h]
    \centering
    \begin{subfigure}[t]{0.48\textwidth}
        \centering
        \includegraphics[width=\textwidth]{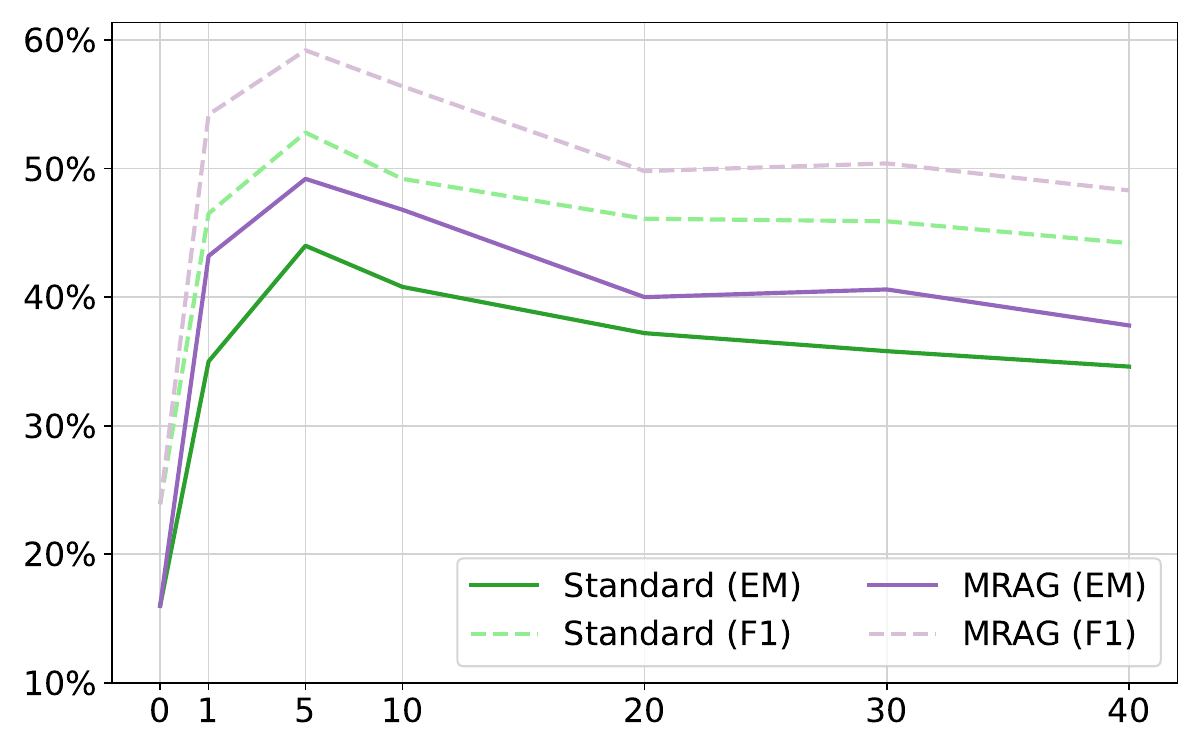}  %
        \caption{\temprageval-\timeqa}
        \label{fig:line_timeqa}
    \end{subfigure}
    \vspace{0.5cm}  %
    \begin{subfigure}[t]{0.48\textwidth}
        \centering
        \includegraphics[width=\textwidth]{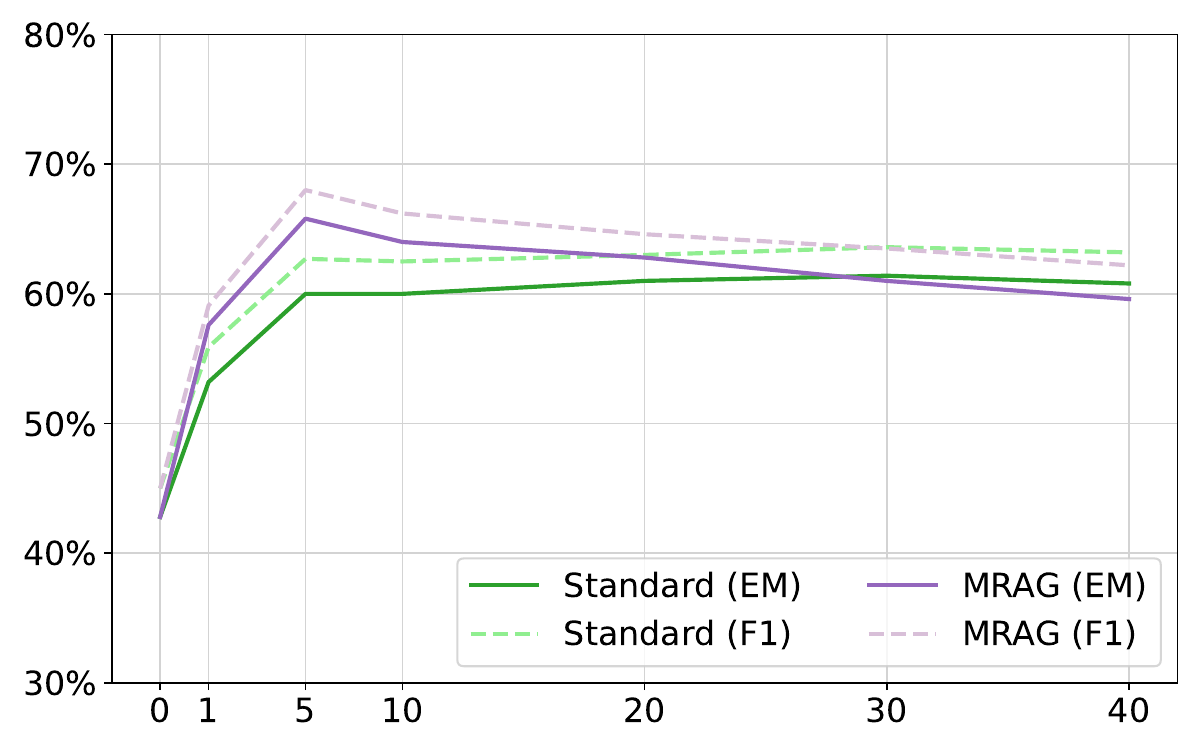}  %
        \caption{\temprageval-\situatedqa}
        \label{fig:line_situatedqa}
    \end{subfigure}
    \captionsetup{justification=justified, singlelinecheck=off}
    \caption{Llama3.1-8B-Instruct reader performance versus number of concatenated context passages retrieved by the \bgegemma and \mrag methods. Standard refers to RAG-Concat and \mrag refers to MRAG-Concat.}
    \label{fig:reader_chunks}
\end{figure*}

We evaluate the Llama reader accuracy with a varying number of concatenated documents retrieved by \bgegemma and \mrag in \Cref{fig:reader_chunks}. The rapid accuracy improvement within the first five passages highlights the effectiveness of RAG in enhancing LLMs' performance by supplementing their knowledge with external information. In both \temprageval subsets, the reader demonstrates higher accuracy with \mrag-retrieved documents in most cases. Notably, \mrag achieves peak accuracy with only the top 5 retrieved documents, whereas \bgegemma might require more, as illustrated in \Cref{fig:line_situatedqa}. For Llama3.1-8B, using 5 documents is optimal, as including more passages in the input may introduce noise and distractors, leading to errors made by the reader.

\section{RAG vs. Long-Document QA}
\label{longdoc}

\temprageval-\timeqa is derived from \timeqa, a dataset originally designed for long-document QA. \timeqa questions are constructed using Wikipedia pages as evidence, ensuring answer presence in the source. The page name is explicitly included in each question \citep{timeqa}. We compare retrieval-augmented generation (RAG) with \bgegemma and \mrag retrievers against the long-document QA setup (without retrieval) using two Llama3.1 models (\Cref{table: longdoc}). In the long-document QA setup, the entire Wikipedia page is provided as context, leading to longer inputs with numerous distractors. Our results show that RAG, when using a high-quality retriever, outperforms long-document QA, validating our hypothesis that supplying full Wikipedia pages introduces noise that degrades performance. Although these LLMs have strong long-context reasoning capabilities \citep{llama3}, they can still be misled by irrelevant passages. The RAG approach mitigates this issue by limiting input passages and excluding irrelevant passages, thereby improving QA accuracy.

\begin{table*}[h]
\centering
\renewcommand{\arraystretch}{1}
\arrayrulecolor{black}
\setlength{\tabcolsep}{10pt}
\begin{tabular}{p{3cm}>{\centering\arraybackslash}p{1.2cm}>{\centering\arraybackslash}p{1.4cm}>{\centering\arraybackslash}p{1.4cm}} 
\toprule
\multirow{2}{*}{\textbf{Method}}  & \multicolumn{3}{c}{\textbf{\temprageval-TimeQA}} \\ 
\cline{2-4}
& \textbf{\# Docs} & \textbf{EM} & \textbf{F1} 
\\
\hline
\multicolumn{4}{c}{\textit{Llama3.1-8B-Instruct}}\\
Direct Prompt & - & 16.0 & 23.9  \\
Direct CoT & - & 16.8 & 27.8\\
RAG-Concat & 5 & 44.0 & 52.8\\
MRAG-Concat & 5 & \textbf{49.2} &\textbf{59.2}   \\
Long-Doc QA & 16.4$^{\sharp}$ & \underline{45.2} & \underline{54.9} \\

\hline
\multicolumn{4}{c}{\textit{Llama3.1-70B-Instruct}}\\
Direct Prompt & - & 31.0 &  \\
Direct CoT & - & 33.2 &45.8  \\
RAG-Concat & 5 &\underline{54.4} &\underline{63.2}   \\
MRAG-Concat & 5 &\textbf{58.0} &\textbf{68.4}   \\
Long-Doc QA  & 16.4$^{\sharp}$ & 48.1 &  59.0  \\

\arrayrulecolor{black}
\bottomrule
\end{tabular}
\caption{End-to-end QA performance comparison for RAG and long-document QA setups. $^{\sharp}$The average number of passages in Wikipedia pages corresponding to \temprageval-\timeqa questions.}
\label{table: longdoc}
\end{table*}

\section{Computational Overhead Assessment}
\label{overhead}

As \mrag involves multiple processing steps, including retrieval, re-ranking, summarization, and hybrid ranking, which could introduce computational overhead compared to standard RAG pipelines. To assess real-world scalability, we assess the average processing (inference) time in seconds per query in comparison to baseline retrieval methods (\ie \mini and \bgegemma). Given retrieved passages by \contriever for each query, these methods rerank the top-100 passages. The processing time of \mrag is further broken down by each module in \Cref{table: overhead}. The assessment is conducted on a machine with one NVIDIA L40S 46G GPU and one AMD EPYC 9554 64-Core CPU. \mrag is implemented using \bgegemma for pure semantic scoring. The inference time can be significantly reduced by using \mini. Compared to \bgegemma, \mrag incurs approximately twice the runtime overhead.

\section{Ablation Study on \mrag Modules}

In the retrieval and summarization module, with retrieved passages, \mrag conducts two key steps: (1) \textit{Passage Keyword Ranking} reduces the passages from 1,000 to 100 based on the keyword presence; (2) \textit{Passage Semantic Ranking} reorders the top-100 passages and splits them into $\sim$500 sentences, including chunk summaries. Similarly, in the semantic-temporal hybrid ranking module, another two key steps are conducted: (1) \textit{Sentence Keyword Ranking} further narrows the scope to 200 sentences; (2) \textit{Hybrid Ranking} ranks these sentences by semantic-temporal hybrid scoring.

To validate the effectiveness of each step in each module, we conduct the ablation study. As shown in \Cref{ablation}, keyword-based ranking steps effectively reduce ranking candidates without significant performance loss, while semantic and hybrid ranking steps markedly enhance retrieval performance. An efficiency-accuracy trade-off can be made by adjusting the number of targeted passages or sentences at each step.

\clearpage
\newpage

\begin{table*}[h]
\centering
\renewcommand{\arraystretch}{1}
\begin{tabular}{p{2cm}ccccc} 
\toprule
         & \textbf{\makecell{Question \\ Processing}} & \textbf{\makecell{Retrieval \& \\ Summarization}} & \textbf{\makecell{Temporal-Semantic \\ Hybrid Ranking}} & \textbf{Total} \\ 
\midrule
\mini  & - & - & - &  0.14 \\ 
\bgegemma &-  &-  & - &  1.03  \\ 
\mrag    & 0.06 & 1.23 & 1.04 &   2.33  \\ 
\bottomrule
\end{tabular}
\captionsetup{justification=justified,singlelinecheck=false}
\caption{Latency assessment (in seconds) for \mrag and baseline retrieval methods.}
\label{table: overhead}
\end{table*}

\begin{table*}[h]
\centering
\renewcommand{\arraystretch}{1.1}
\begin{tabular}{lccccccccc} 
\toprule[1.5pt]
\multicolumn{2}{r}{\begin{tabular}[c]{@{}r@{}}\textbf{\# Chunk / Sent.}\end{tabular}} & \multicolumn{4}{c}{\textbf{Answer Recall @}}                                & \multicolumn{4}{c}{\textbf{Gold Evidence Recall @}}                          \\ 
\hline
\textbf{Steps}                      &                                                    & \textbf{1}            & \textbf{5}           & \textbf{10}           & \textbf{20}           & \textbf{1}            & \textbf{5}            & \textbf{10}          & \textbf{20}            \\ 
\hline
Chunk Retrieving      & 21M   &  22.6 & 51.1 & 65.5 & 79.5  &  6.8 &  17.1 & 22.9  & 30.5  \\
+ Chunk Keyword Ranking    & 1000  & 29.2  & 62.4  & 77.4  &  87.1 &  12.4  & 26.1 & 35.5  & 46.8  \\
+ Chunk Semantic Ranking   & 100   &  55.3  & 84.0  & 88.2  &  95.5 &  22.9 &  50.0 & 59.0  &  66.8  \\
+ Sent. Keyword Ranking & 500    &  55.3  & 81.8 & 85.5 & 91.6 &  23.4 &  49.0 & 55.3  &  62.4 \\
+ Sent. Hybrid Ranking  & 200    & \textbf{61.3}  & \textbf{89.0}  & \textbf{93.4} & \textbf{94.2} & \textbf{31.1}  & \textbf{59.2} & \textbf{65.8} & \textbf{69.0}  \\
\bottomrule[1.5pt]
\end{tabular}
\captionsetup{justification=justified,singlelinecheck=false}
\caption{The ablation study of key steps of \mrag on perturbed queries in \temprageval $-$ \situatedqa.}
\label{ablation}
\end{table*}

\onecolumn
\section{Case Studies}
\label{sec: cases}
\subsection{Retrieval Failure Case Study}
\label{sec: cases_retrieval}

We conducted five case studies to qualitatively evaluate the advantages of \mrag over the \bgegemma retriever as below. The results demonstrate \mrag's robustness to temporal perturbations and its ability to retrieve relevant context passages. For instance, in Case 1, the top-1 passage retrieved by \bgegemma matches the query date ``\texttt{1988}'' but discusses a father-son record set in 2007. In contrast, the first passage retrieved by \mrag focuses on a teammate combination record in the same season, despite the date ``\texttt{1961}'' differing from the query date ``\texttt{1988}''. Since semantic relevance outweighs strict date matching in this situation, \mrag provides more contextually appropriate results for the time-sensitive question.

\renewcommand{\arraystretch}{1.2}
\begin{longtable*}{|p{0.2\linewidth}|p{0.35\linewidth}|p{0.35\linewidth}|} \hline

\textbf{Question}   & \textbf{Gemma-based Retrieval} & \textbf{Modular Retrieval}  \\ 

\endfirsthead

\hline
\textbf{Question}   & \textbf{Gemma-based Retrieval} & \textbf{Modular Retrieval}  \\ 
\hline
\endhead

\hline
\multirow{2}{=}{(1) Who had the most home runs by two teammates in a season as of 1988?} &
  \#7 is the top true evidence &
  \#1 is the top true evidence \\
  \cline{2-3}
 & 
\#1 Bobby Bonds | ... until José Canseco of the Oakland Athletics in 1988. Barry and Bobby had 1,094 combined home runs through 2007 — a record for a father-son combination.\newline\newline \#2 1987 in baseball | ... With teammate Howard Johnson already having joined, it marks the first time that two teammates achieve 30–30 seasons in the same year.\newline\newline \#3 1988 Toronto Blue Jays season | April 4, 1988: George Bell set a major league record for the most home runs hit on Opening Day, with three ...\newline\newline \#7 50 home run club | \textcolor{ForestGreen}{\textbf{M\&M Boys}}—are the only teammates to reach the 50 home run club in the same season, hitting a combined 115 home runs in 1961 and breaking the single-season record for home runs by a pair of teammates.
 & 
\#1 50 home run club | \textcolor{ForestGreen}{\textbf{M\&M Boys}}—are the only teammates to reach the 50 home run club in the same season, hitting a combined 115 home runs in 1961 and breaking the single-season record for home runs by a pair of teammates.\newline\newline \#2 1987 Major League Baseball season | Cal Ripken, Jr. is lifted from the lineup and replaced by Ron Washington ... it marks the first time that two teammates achieve 30–30 seasons in the same year.\newline\newline \#3 1987 in baseball | Whitt connects on three of the home runs ... it marks the first time that two teammates achieve 30–30 seasons in the same year.
 \\
\hline
\pagebreak
\multirow{2}{=}{(2) Who had the most home runs by two teammates in a season by August 17, 1992?} &
  No true evidence retrieved &
  \#1 is the top true evidence \\
  \cline{2-3}
 & 
\#1 1992 in baseball | ... August 28 – The Milwaukee Brewers lash 31 hits in a 22-2 drubbing of the Toronto Blue Jays , setting a record for the most hits by a team in a single nine-inning game.\newline\newline\#2 1997 in baseball |  ... McGwire, who hit a major league-leading 52 homers for the Oakland Athletics last season, becomes the first player with back-to-back 50-homer seasons since Ruth did it ...
 & 
\#1 50 home run club | \textcolor{ForestGreen}{\textbf{M\&M Boys}}—are the only teammates to reach the 50 home run club in the same season, hitting a combined 115 home runs in 1961 and breaking the single-season record for home runs by a pair of teammates.\newline\newline \#2 List of career achievements by Babe Ruth | 1927 (Ruth 60, Lou Gehrig 47) ... Achieved by several other pairs of teammates since ... Two teammates with 40 or more home runs, season: Thrice Clubs with three consecutive home runs in inning ...
 \\
 \hline
\multirow{2}{=}{(3) Who won the latest America's Next Top Model as of 2021?} &
  No true evidence retrieved &
  \#3 is the top true evidence \\
  \cline{2-3}
 & 
\#1 America's Next Top Model (season 17) |  the final season for Andre Leon Talley as a judge. The winner of the competition was 30-year-old Lisa D'Amato from Los Angeles, California, who originally placed sixth on Cycle 5 making her the oldest winner at the age of 30. Allison Harvard, who originally placed second on cycle 12 ... \newline\newline\#2 Germany's Next Topmodel | that Soulin Omar who was the second runner up, should've won based on her performance throughout the season. German Magazine "OK!" and "Der Westen" stated ... \newline\newline\#3 America's Next Top Model (season 21) |  (Ages stated are at start of contest) Indicates that the contestant died after filming ended
 & 
\#1 America's Next Top Model (season 23) |  The twenty-third cycle of America's Next Top Model premiered on December 12, 2016 ... The winner of the competition was 20 year-old India Gants from Seattle ...\newline\newline
\#2 America's Next Top Model |  ... five contestants were featured modeling Oscar gowns: ... On May 12, 2010, Angelea Preston, Jessica Serfaty, and Simone Lewis (all cycle 14) appeared on a Jay Walking ... On February 24, 2012, Brittany Brower (cycle 4), Bre Scullark (cycle 5) (both cycle 17), and Lisa D'Amato (cycle 5 and cycle 17 winner) appeared on a Jay\newline\newline
\#3 America's Next Top Model (season 24) |  The twenty-fourth cycle of America's Next Top Model premiered on January 9, 2018 ... The winner of the competition was 20 year-old \textcolor{ForestGreen}{\textbf{Kyla Coleman}} from Lacey, Washington ...
\\
\hline
\pagebreak
\multirow{2}{=}{(4) When did Dwight Howard play for Los Angeles Lakers between 2000 and 2017?} &
  \#5 is the top true evidence &
  \#1 is the top true evidence \\
  \cline{2-3}
 & 
\#1 List of career achievements by Dwight Howard | Defensive rebounds, 5-game series: 58, Orlando Magic vs. Los Angeles Lakers, 2009\newline\newline\#2 Dwight Howard |  wanted". In a 2013 article titled "Is Dwight Howard the NBA's Worst Teammate?" ... When he was traded from the Atlanta Hawks to the Charlotte Hornets, some of his Hawks teammates reportedly cheered. After Charlotte traded Howard to the Washington Wizards, Charlotte player Brendan Haywood asserted ...\newline\newline\#5 \textcolor{ForestGreen}{\textbf{2012-13}} Los Angeles Lakers season | In a March 12, 2013 game against his former team, the Orlando Magic, Dwight Howard tied his own NBA record of 39 free throw attempts ...
 & 
\#1 Dwight Howard |  On August 10, \textcolor{ForestGreen}{\textbf{2012}}, Howard was traded from Orlando to the Los Angeles Lakers in a deal that also involved the Philadelphia 76ers and the Denver Nuggets ...\newline\newline\#2 Dwight Howard | ... In \textcolor{ForestGreen}{\textbf{2012}}, after eight seasons with Orlando, Howard was traded to the Los Angeles Lakers ... Howard returned to the Lakers in 2019 and won his first NBA championship in 2020.
\\
\hline
\pagebreak
\multirow{2}{=}{(5) Which political party did Clive Palmer belong to on Apr 20, 1976?} &
  No true evidence retrieved &
  \#3 is the top true evidence
  \\
  \cline{2-3}
 & 
\#1 Clive Palmer | ... On 25 April 2013, Palmer announced a "reformation" of the United Australia Party, which had been folded into the present-day Liberal Party in 1945, to stand candidates in the 2013 federal election, and had applied for its registration in Queensland ...\newline\newline\#2 Clive Palmer |  de-registering the party on 5 May 2017, Palmer revived his party as the United Australia Party, announcing that he would be running candidates for all 151 seats in the House of Representatives and later that he would run as a Queensland candidate for the Senate. In the 2019 federal election, despite extensive advertising ...\newline\newline\#3 United Australia Party (2013) |  Clive Palmer of bullying, swearing and yelling at people. Lazarus stated "I have a different view of team work. Given this, I felt it best that I resign from the party and pursue my senate role as an independent senator." ...
 & 
\#1 Clive Palmer |  Palmer deregistered the party's state branches in September 2016, initially intending to keep it active at the federal level. However, in April 2017, he announced that the party would be wound up. In February 2018, Palmer announced his intention to resurrect his party and return to federal politics. The party was revived in June under its original name, the United Australia Party ...\newline\newline\#2 Clive Palmer |  ... Palmer resigned his life membership of the Liberal National Party. His membership of the party had been suspended on 9 November 2012, following his comments on the actions of state government ministers. He was re-instated to the party on 22 November, but resigned the same day ...\newline\newline\#3 Clive Palmer |  Palmer was instrumental in the split of the South Australian conservatives in the 1970s, and was active in the Liberal Movement headed by former Premier of South Australia, Steele Hall. Palmer joined the \textcolor{ForestGreen}{\textbf{Queensland division of the Nationals}} in 1974 ...

\\
\hline
\end{longtable*}

\twocolumn

\subsection{Downstream QA Failure Case Study}
\label{sec: cases_qa}

\begin{figure}[h!]
    \centering
    \includegraphics[width=0.48\textwidth]{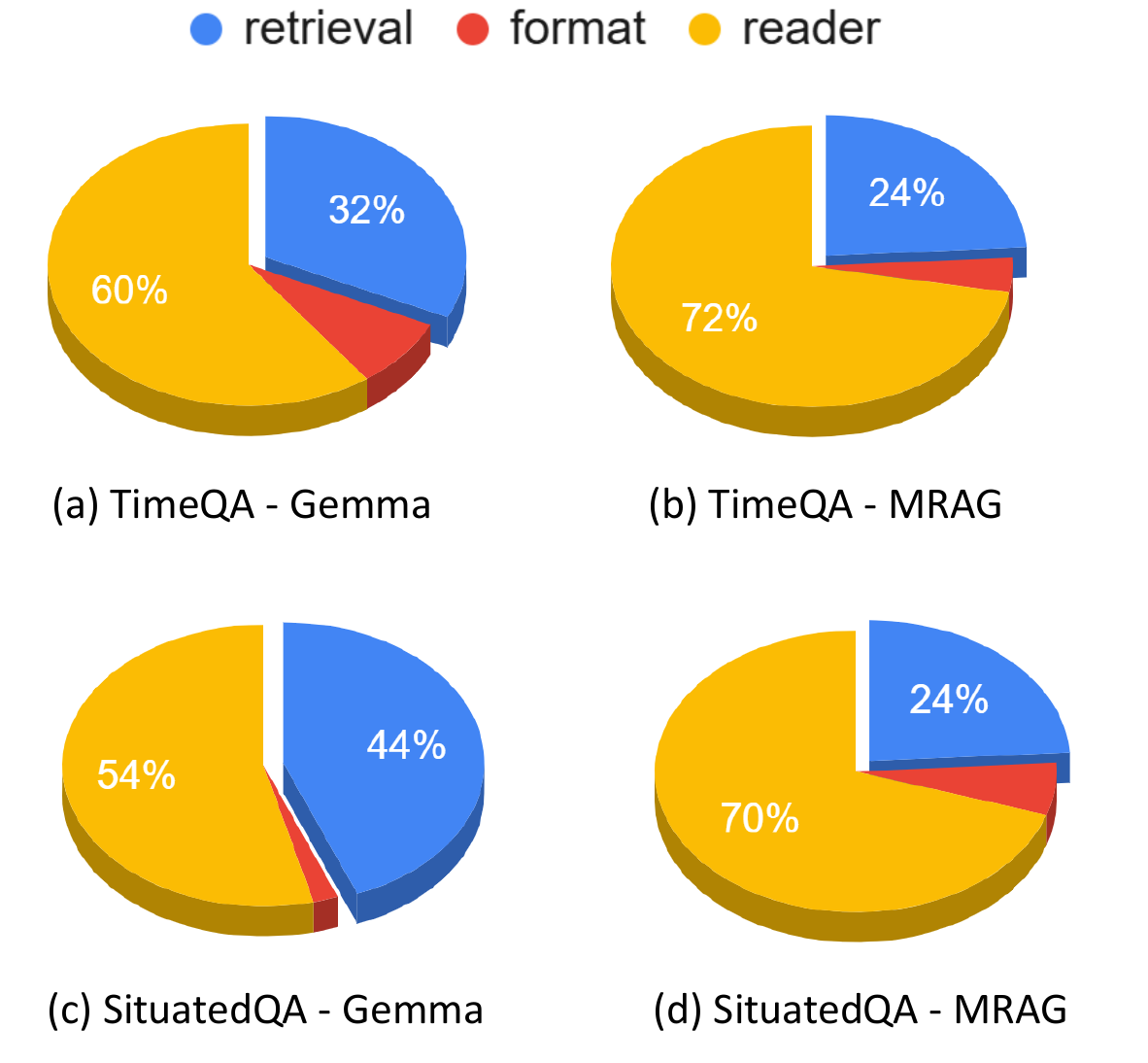}
    \captionsetup{justification=justified, singlelinecheck=false}  %
    \caption{Percentage distribution of error case root causes on \temprageval-\timeqa and \temprageval-\situatedqa. Gemma refers to RAG-Concat and \mrag refers to MRAG-Concat.}
    \label{QA_failure}
\end{figure}

To identify the most error-prone component (retrieval or generation), we manually analyze 50 random failure cases for \bgegemma and another 50 for \mrag from \temprageval-\timeqa. The same analysis is applied to \temprageval-\situatedqa, focusing on the RAG-Concat and \mrag-Concat methods using the Llama3.1-8B-Instruct model. We categorize each failure by root cause: retrieval, format, or reader. Failures are attributed to the retriever when it fails to find at least one relevant passage within the top-5 retrieved results. In cases where the reader model receives relevant passages, errors are classified as format if the generated answer is correct but in a different format, or as reader errors if the model fails to perform temporal reasoning correctly, despite having access to relevant knowledge. Our analysis, shown in \Cref{QA_failure}, reveals that the majority of errors stem from the reader, indicating that both retrievers perform well. Compared to \bgegemma, \mrag exhibits a lower percentage of retrieval errors, \eg \Cref{QA_failure}(b) vs. \Cref{QA_failure}(a), demonstrating the effectiveness of our proposed retrieval approach.

\newpage

\section{Knowledge Conflicts Between Parametric Knowledge and External Passages}
\label{sec: conflict}

RAG systems commonly confront knowledge conflicts either between LLM internal knowledge and external passage knowledge or across different passages \citep{conflict}. We illustrate two categories of examples where the LLM internal knowledge conflicts with the retrieved passage, using the \bgegemma retriever and the Llama3.1-8B-Instruct model. RAG systems typically prioritize external knowledge in the retrieved context. Therefore, we observe a significant amount of errors in parametric knowledge are avoided by providing relevant passages as examples in the top of \Cref{table: conflicts}. Besides, when only irrelevant passages are retrieved, the correct parametric knowledge can be misled by the the distracting context as examples in the bottom of \Cref{table: conflicts}. Thus, the retriever performance is of great importance.

External passages may have conflicting knowledge, requiring LLMs to make nuanced judgments in such cases \citep{whos}. In our experiments on time-sensitive question answering using the Wikipedia corpus, we rarely observe conflicting passages within Wikipedia. Different answers typically correspond to different temporal constraints. To gain a deeper understanding of conflicting passages, one approach would be to introduce counterfactual documents. However, as our research focuses on temporal reasoning in retrieval, we leave this direction for future work.

\onecolumn

\begin{table*}[hbt!]
\centering
\renewcommand{\arraystretch}{1.1}
\begin{tabular}{p{3.5cm}p{6.5cm}p{2.1cm}p{2.1cm}} 
\toprule
   \multicolumn{4}{l}{\textbf{Wrong parametric predictions with relevant passages.}} \\ 
   \midrule
\textbf{Query} & \textbf{Relevant passage} & \textbf{CoT pred.} & \textbf{RAG pred.} \\
\midrule
Who owned the Newton D. Baker House in Washington DC from 1978 to 1982? &  Newton D. Baker House |  … Straight and his wife lived in the home from until 1976. In 1976, Yolande Bebeze Fox, the former Miss America 1951, bought the home from Straight. Fox lived in the home until her death in February 2016. & \textcolor{red}{American Enterprise Institute} & \textcolor{ForestGreen}{Yolande Bebeze Fox} \\ 
\midrule
What was the last position of Homer Thornberry between 1941 to 1943? & Homer Thornberry |  Thornberry was born in Austin, Texas … He was district attorney of Travis County, Texas from 1941 to 1942. He was a United States Navy Lieutenant Commander from 1942 to 1946 … & \textcolor{red}{United States Senator from Texas} & \textcolor{ForestGreen}{United States Navy Lieutenant Commander} \\ 
\midrule
Who was the chair of National Council of French Women in Dec 1951? & National Council of French Women | … Marguerite Pichon-Landry (1878–1972) chaired the Legislation section of the CNFF from 1914 to 1927, and was secretary-general from 1929 to 1932. She was president from 1932 to 1952 … & \textcolor{red}{Éliane Brault} & \textcolor{ForestGreen}{Marguerite Pichon-Landry} \\
\midrule
Warlugulong was owned by whom in 1997? & Warlugulong | … the work was sold by art dealer Hank Ebes on 24 July 2007, setting a record price for a contemporary Indigenous Australian art work bought at auction when it was purchased by the National Gallery of Australia for A\$2.4 million. & \textcolor{red}{the Pritzker family} & \textcolor{ForestGreen}{Hank Ebes} \\
\midrule
   \multicolumn{4}{l}{\textbf{Correct parametric predictions with irrelevant (distracting) passages.}} \\ 
   \midrule
\textbf{Query} & \textbf{Irrelevant passage} & \textbf{CoT pred.} & \textbf{RAG pred.} \\
\midrule
What was the first U-boat unit Erich Topp commanded between 5 October 1937 and December 1941? & Erich Topp |  World War II commenced following the German invasion of Poland on 1 September 1939. U-46, under the command of Sohler, had already been at sea since 19 August, returning to port on 15 September. & \textcolor{ForestGreen}{1st U-boat Flotilla} & \textcolor{red}{U-46} \\
\midrule
Who is the first one owned Arnolfini Portrait after 1900? & Arnolfini Portrait |  The Arnolfini Portrait (or The Arnolfini Wedding, The Arnolfini Marriage, the Portrait of Giovanni Arnolfini and his Wife, or other titles) is a 1434 oil painting on oak panel by the Early Netherlandish painter Jan van Eyck … & \textcolor{ForestGreen}{The National Gallery} & \textcolor{red}{There is no information about who owned the Arnolfini Portrait after 1900 in the given context.} \\
\bottomrule
\end{tabular}
\captionsetup{justification=justified,singlelinecheck=false}
\caption{Examples of LLM parametric knowledge and retrieved passages.}
\label{table: conflicts}
\end{table*}

\section{\temprageval Examples}

\begin{table*}[hbt!]
\centering
\renewcommand{\arraystretch}{1.1}
\begin{tabular}{p{3cm}p{12cm}} 
\toprule
   & \textbf{\temprageval-\situatedqa} \\ 
\midrule
Question &  When did Dwight Howard play for Los Angeles Lakers between 2000 and 2017? \\ 
Answer &  2012 | 2013 | 2012-2013 \\ 
Gold evidence &  Dwight Howard | ... On August 10, 2012, Howard was traded from Orlando to the Los Angeles Lakers in a deal that also involved the Philadelphia 76ers and the Denver Nuggets ...\\
\midrule
Question & When was the earliest time Dwight Howard play for the Lakers after August 10, 2014? \\ 
Answer & 2019 | 2020 | 2019-2020 \\
Gold evidence &  Dwight Howard | ... On August 26, 2019, Howard signed a \$2.6 million veteran's minimum contract with the Los Angeles Lakers, reuniting him with his former team ...\\
\midrule
Question & When did the last season on The 100 come out between 2018 and 2021? \\ 
Answer & May 20, 2020 | 2020 \\
Gold evidence (1) &  The 100 (TV series) | ... The CW renewed the series for a seventh season, that would consist of 16 episodes and premiered on May 20, 2020 ...\\
Gold evidence (2) & The 100 season 7 | ... On March 4, 2020, it was revealed that the last season of The 100 would premiere on The CW on May 20, 2020 ...\\
\midrule
Question & Who was the leader of the Ontario PC Party after 2020? \\ 
Answer & Doug Ford \\
Gold evidence (1) &  Progressive Conservative Party of Ontario | ... On March 10, 2018, Doug Ford, former Toronto city councillor ... was elected as leader of the PC Party ...\\
Gold evidence (2) &  New Blue Party of Ontario | ... on March 10, 2018, Doug Ford was elected as leader of the Progressive Conservative Party of Ontario ...\\
\midrule
   & \textbf{\temprageval-\timeqa} \\ 
\midrule
Question & Oliver Bulleid was an employee for whom as of Oct 1905? \\ 
Answer & Great Northern Railway \\
Gold evidence &  Oliver Bulleid | ... In 1901 ... he joined the Great Northern Railway (GNR) at Doncaster at the age of 18, as an apprentice under H. A. Ivatt ...\\
\midrule
Question & Fred Hoiberg was the coach of which team between 2016 and 2017? \\ 
Answer & Chicago Bulls | Bulls \\
Gold evidence &  Fred Hoiberg | On June 2, 2015, the Chicago Bulls hired Hoiberg as head coach ... On December 3, 2018, the Bulls fired Hoiberg ...\\
\midrule
Question & Who was the first spouse of Merle Oberon since May 7, 1948? \\ 
Answer & Lucien Ballard \\
Gold evidence &  Merle Oberon | ... She divorced him in 1945, to marry cinematographer Lucien Ballard ...\\
\midrule
Question & When was the last airplane crashing for All Nippon Airways as of 1970? \\ 
Answer & 13 November 1966 | 1966 \\ 
Gold evidence &  All Nippon Airways | ... On 13 November 1966, Flight 533 operated by a NAMC YS-11, crashed in the Seto Inland Sea off ...\\

\bottomrule

\end{tabular}
\captionsetup{justification=justified,singlelinecheck=false}
\label{examples}
\end{table*}

\pagebreak
\section{Prompts List}
\label{sec: prompts}

\begin{table*}[hbt!]
\centering
\begin{tabular}{>{\hspace{0pt}}m{0.94\linewidth}} 
\toprule
\textbf{Keyword Extraction Prompting}\par{}Your task is to extract keywords from the question. Response by a list of keyword strings. Do not include pronouns, prepositions, articles.\par\null\par{}There are some examples for you to refer to:\par{}\textless{}Question\textgreater{}\par{}When was the last time the United States hosted the Olympics?\par{}\textless{}/Question\textgreater{}\par{}\textless{}Keywords\textgreater{}\par{}{[}"United States", "hosted", "Olympics"]\par{}\textless{}/Keywords\textgreater{}\par\null\par{}\textless{}Question\textgreater{}\par{}Who sang 1 national anthem for Super Bowl last year?\par{}\textless{}/Question\textgreater{}\par{}\textless{}Keywords\textgreater{}\par{}{[}"sang", "1", "national anthem", "Super Bowl"]\par{}\textless{}/Keywords\textgreater{}\par\null\par{}\textless{}Question\textgreater{}\par{}Who runs the fastest 40-yard dash in the NFL?\par{}\textless{}/Question\textgreater{}\par{}\textless{}Keywords\textgreater{}\par{}{[}"runs", "fastest", "40-yard", "dash", "NFL"]\par{}\textless{}/Keywords\textgreater{}\par\null\par{}\textless{}Question\textgreater{}\par{}When did Khalid write Young Dumb and Broke?\par{}\textless{}/Question\textgreater{}\par{}\textless{}Keywords\textgreater{}\par{}{[}"Khalid", "write", "Young Dumb and Broke"]\par{}\textless{}/Keywords\textgreater{}\par\null\par{}Now your question is\\\textless{}Question\textgreater{}\par{}\{normalized question\}\par{}\textless{}/Question\textgreater{}\par{}\textless{}Keywords\textgreater{}  \\
\toprule
\end{tabular}
\captionsetup{justification=justified,singlelinecheck=false}
\caption{Detailed prompts for Keyword Extraction.}
\label{keyword}
\end{table*}

\newpage

\begin{table*}
\centering
\begin{tabular}{>{\hspace{0pt}}m{0.94\linewidth}} 
\toprule
\textbf{Query-Focused Summarization Prompting}\par{}You are given a context paragraph and a specific question. Your goal is to summarize the context paragraph in one standalone sentence by answering the given question. If dates are mentioned in the paragraph, include them in your answer. If the question cannot be answered based on the paragraph, respond with "None". Ensure that the response is relevant, complete, concise and directly addressing the question.\par\null\par{}There are some examples for you to refer to:\par{}\textless{}Context\textgreater{}\par{}Houston Rockets \textbar{} The Houston Rockets have won the NBA championship twice in their history. Their first win came in 1994, when they defeated the New York Knicks in a seven-game series. The following year, in 1995, they claimed their second title by sweeping the Orlando Magic. Despite several playoff appearances in the 2000s and 2010s, the Rockets have not reached the NBA Finals since their last championship victory in 1995.\par{}\textless{}/Context\textgreater{}\par{}\textless{}Question\textgreater{}\par{}When did the Houston Rockets win the NBA championship?\par{}\textless{}/Question\textgreater{}\par{}\textless{}Summarization\textgreater{}\par{}The Houston Rockets have won the NBA championship in 1994 and 1995.\par{}\textless{}/Summarization\textgreater{}\par\null\par{}\textless{}Context\textgreater{}\par{}2019 Grand National \textbar{} The 2019 Grand National (officially known as the Randox Health 2019 Grand National for sponsorship reasons) was the 172nd annual running of the Grand National horse race at Aintree Racecourse near Liverpool, England. The showpiece steeplechase is the pinnacle of a three-day festival which began on 4 April, followed by Ladies' Day on 5 April.\par{}\textless{}/Context\textgreater{}\par{}\textless{}Question\textgreater{}\par{}Who won the Grand National?\par{}\textless{}/Question\textgreater{}\par{}\textless{}Summarization\textgreater{}\par{}None\par{}\textless{}/Summarization\textgreater{}\par\null\par{}Now your question and paragraph are\par{}\textless{}Context\textgreater{}\par{}\{title\} \textbar{} \{text\}\par{}\textless{}/Context\textgreater{}\par{}\textless{}Question\textgreater{}\par{}\{normalized question\}\par{}\textless{}/Question\textgreater{}\par{}\textless{}Summarization\textgreater{}  \\
\toprule
\end{tabular}
\captionsetup{justification=justified,singlelinecheck=false}
\caption{Detailed prompts for Query-Focused Summarization.}
\label{qfs}
\end{table*}

\begin{table*}
\centering
\begin{tabular}{>{\hspace{0pt}}m{0.94\linewidth}} 
\toprule
\textbf{Reader Direct Prompting}\par{}As an assistant, your task is to answer the question directly after \textless{}Question\textgreater{}. Your answer should be after \textless{}Answer\textgreater{}.\par{}\\There are some examples for you to refer to:\par{}\textless{}Question\textgreater{}\par{}When did England last get to the semi final of a World Cup before 2019?\par{}\textless{}/Question\textgreater{}\par{}\textless{}Answer\textgreater{}\par{}2018\par{}\textless{}/Answer\textgreater{}\par\null\par{}\textless{}Question\textgreater{}\par{}Who sang the national anthem in the last Super Bowl as of 2021?\par{}\textless{}/Question\textgreater{}\par{}\textless{}Answer\textgreater{}\par{}Eric Church and Jazmine Sullivan\par{}\textless{}/Answer\textgreater{}\par\null\par{}\textless{}Question\textgreater{}\par{}What's the name of the latest Pirates of the Caribbean by 2011?\par{}\textless{}/Question\textgreater{}\par{}\textless{}Answer\textgreater{}\par{}On Stranger Tides\par{}\textless{}/Answer\textgreater{}\par\null\par{}\textless{}Question\textgreater{}\par{}What was the last time France won World Cup between 2016 and 2019?\par{}\textless{}/Question\textgreater{}\par{}\textless{}Answer\textgreater{}\par{}Priscilla Joan Torres\par{}\textless{}/Answer\textgreater{}\par\null\par{}\textless{}Question\textgreater{}\par{}Which school did Marshall Sahlins go to from 1951 to 1952?\par{}\textless{}/Question\textgreater{}\par{}\textless{}Answer\textgreater{}\par{}Columbia University\par{}\textless{}/Answer\textgreater{}\par\null\par{}Now your Question is\par{}\textless{}Question\textgreater{}\par{}\{question\}\par{}\textless{}/Question\textgreater{}\par{}\textless{}Answer\textgreater{}  \\
\toprule
\end{tabular}
\captionsetup{justification=justified,singlelinecheck=false}
\caption{Detailed prompts for Reader Direct Question Answering.}
\label{dp}
\end{table*}

\begin{table*}
\centering
\begin{tabular}{>{\hspace{0pt}}m{0.94\linewidth}} 
\toprule
\textbf{Reader Chain-of-Thought Prompting}\par{}As an assistant, your task is to answer the question after \textless{}Question\textgreater{}. You should first think step by step about the question and give your thought and then answer the \textless{}Question\textgreater{} in the short form. Your thought should be after \textless{}Thought\textgreater{}. The direct answer should be after \textless{}Answer\textgreater{}.\par{}\\There are some examples for you to refer to:\par{}\textless{}Question\textgreater{}\par{}When did England last get to the semi final of a World Cup before 2019?\par{}\textless{}/Question\textgreater{}\par{}\textless{}Thought\textgreater{}\par{}England has reached the semi-finals of FIFA World Cup in 1966, 1990, 2018. The latest year before 2019 is 2018. So the answer is 2018.\par{}\textless{}/Thought\textgreater{}\par{}\textless{}Answer\textgreater{}\par{}2018\par{}\textless{}/Answer\textgreater{}\par\null\par{}\textless{}Question\textgreater{}\par{}Who sang the national anthem in the last Super Bowl as of 2021?\par{}\textless{}/Question\textgreater{}\par{}\textless{}Thought\textgreater{}\par{}The last Super Bowl as of 2021 is Super Bowl LV, which took place in February 2021. In Super Bowl LV, the national anthem was performed by Eric Church and Jazmine Sullivan. So the answer is Eric Church and Jazmine Sullivan.\par{}\textless{}/Thought\textgreater{}\par{}\textless{}Answer\textgreater{}\par{}Eric Church and Jazmine Sullivan\par{}\textless{}/Answer\textgreater{}\par\null\par{}\textless{}Question\textgreater{}\par{}Where was the last Rugby World Cup held between 2007 and 2016?\par{}\textless{}/Question\textgreater{}\par{}\textless{}Thought\textgreater{}\par{}The Rugby World Cup was held in 1987, 1991, 1995, 1999, 2003, 2007, 2011, 2015, 2019. The last Rugby World Cup held between 2007 and 2016 is in 2015. The IRB 2015 Rugby World Cup was hosted by England. So the answer is England.\par{}\textless{}/Thought\textgreater{}\par{}\textless{}Answer\textgreater{}\par{}England\par{}\textless{}/Answer\textgreater{}\par\null\par{}Now your Question is\par{}\textless{}Question\textgreater{}\par{}\{question\}\par{}\textless{}/Question\textgreater{}\par{}\textless{}Thought\textgreater{}  \\
\toprule
\end{tabular}
\captionsetup{justification=justified,singlelinecheck=false}
\caption{Detailed prompts for Reader Chain-of-Thought Question Answering.}
\label{cot}
\end{table*}

\begin{table*}
\centering
\begin{tabular}{>{\hspace{0pt}}m{0.94\linewidth}} 
\toprule
\textbf{Retrieval-Augmented Reader Prompting}\par{}As an assistant, your task is to answer the question based on the given knowledge. Your answer should be after \textless{}Answer\textgreater{}. The given knowledge will be after the \textless{}Context\textgreater{} tage. You can refer to the knowledge to answer the question. If the context knowledge does not contain the answer, answer the question directly.\par\null\par{}There are some examples for you to refer to:\par{}\textless{}Context\textgreater{}\par{}Sport in the United Kingdom Field \textbar{} hockey is the second most popular team recreational sport in the United Kingdom. The Great Britain men's hockey team won the hockey tournament at the 1988 Olympics, while the women's hockey team repeated the success in the 2016 Games.\par\null\par{}Three Lions (song) \textbar{} The song reached number one on the UK Singles Chart again in 2018 following England reaching the semi-finals of the 2018 FIFA World Cup, with the line "it's coming home" featuring heavily on social media.\par\null\par{}England national football team \textbar{} They have qualified for the World Cup sixteen times, with fourth-place finishes in the 1990 and 2018 editions.\par{}\textless{}/Context\textgreater{}\par{}\textless{}Question\textgreater{}\par{}When did England last get to the semi final of a World Cup before 2019?\par{}\textless{}/Question\textgreater{}\par{}\textless{}Answer\textgreater{}\par{}2018\par{}\textless{}/Answer\textgreater{}\par\null\par{}\textless{}Context\textgreater{}\par{}Bowl LV \textbar{} For Super Bowl LV, which took place in February 2021, the national anthem was performed by Eric Church and Jazmine Sullivan. They sang the anthem together as a duet.\par\null\par{}Super Bowl LVI \textbar{} For Super Bowl LVI, which took place in February 2022, the national anthem was performed by Mickey Guyton. She delivered a powerful rendition of the anthem.\par{}\textless{}/Context\textgreater{}\par{}\textless{}Question\textgreater{}\par{}Who sang the national anthem in the last Super Bowl as of 2021?\par{}\textless{}/Question\textgreater{}\par{}\textless{}Answer\textgreater{}\par{}Eric Church and Jazmine Sullivan\par{}\textless{}/Answer\textgreater{}\par\null\par{}Now your question and context knowledge are\par{}\textless{}Context\textgreater{}\par{}\{texts\}\par{}\textless{}/Context\textgreater{}\par{}\textless{}Question\textgreater{}\par{}\{question\}\par{}\textless{}/Question\textgreater{}\par{}\textless{}Answer\textgreater{}  \\
\toprule
\end{tabular}
\captionsetup{justification=justified,singlelinecheck=false}
\caption{Detailed prompts for Retrieval-Augmented Question Answering.}
\label{c_prompt}
\end{table*}

\begin{table*}[h!]
\centering
\begin{tabular}{>{\raggedright\arraybackslash}m{0.94\linewidth}} 
\toprule
\textbf{Relevance Checking Prompting} \\
You will be given a context paragraph and a question. Your task is to decide whether the context is relevant and contains the answer to the question. Requirements are as follows: \\
- First, read the paragraph after \textless{}Context\textgreater{} and the question after \textless{}Question\textgreater{} carefully. \\
- Then you should think step by step and give your thought after \textless{}Thought\textgreater{}. \\
- Finally, write the response as "Yes" or "No" after \textless{}Response\textgreater{}. \\\\
There are some examples for you to refer to: \\
\textless{}Context\textgreater{} \\
Petronas Towers \textbar{} From 1996 to 2004, they were officially designated as the tallest buildings in the world until they were surpassed by the completion of Taipei 101. The Petronas Towers remain the world's tallest twin skyscrapers, surpassing the World Trade Center towers in New York City, and were the tallest buildings in Malaysia until 2019, when they were surpassed by The Exchange 106. \\
\textless{}/Context\textgreater{} \\
\textless{}Question\textgreater{} \\
Tallest building in the world? \\
\textless{}/Question\textgreater{} \\
\textless{}Thought\textgreater{} \\
The question asks what the tallest building in the world is. The context paragraph talks about the Petronas Towers. The context paragraph states that Petronas Towers were officially designated as the tallest buildings in the world from 1996 to 2004. And the Taipei 101 became the tallest building in the world after 2004. This context paragraph contains two answers to the question. Therefore, the response is "Yes". \\
\textless{}/Thought\textgreater{} \\
\textless{}Response\textgreater{} \\
Yes \\
\textless{}/Response\textgreater{} \\\\

Now your context paragraph and question are: \\
\textless{}Context\textgreater{} \\
\{context\} \\
\textless{}/Context\textgreater{} \\
\textless{}Question\textgreater{} \\
\{normalized question\} \\
\textless{}/Question\textgreater{} \\
\textless{}Thought\textgreater{} \\
\toprule
\end{tabular}
\captionsetup{justification=justified,singlelinecheck=false}
\caption{Detailed prompts for relevance checking.}
\label{check_prompt}
\end{table*}

\begin{table}[h!]
\centering
\begin{tabular}{>{\raggedright\arraybackslash}m{0.94\linewidth}} 
\toprule
\textbf{Independent Reading Prompting} \\ 
You are a summarizer summarizing a retrieved document about a user question. Keep the key dates in the summarization. Write "None" if the document has no relevant content about the question. \\
\\
There are some examples for you to refer to: \\

\textless{}Document\textgreater{} \\
David Beckham \textbar{} As the summer 2003 transfer window approached, Manchester United appeared keen to sell Beckham to Barcelona and the two clubs even announced that they reached a deal for Beckham's transfer, but instead he joined reigning Spanish champions Real Madrid for €37 million on a four-year contract. Beckham made his Galaxy debut, coming on for Alan Gordon in the 78th minute of a 0--1 friendly loss to Chelsea as part of the World Series of Soccer on 21 July 2007. \\
\textless{}/Document\textgreater{} \\

\textless{}Question\textgreater{} \\
David Beckham played for which team? \\
\textless{}/Question\textgreater{} \\

\textless{}Summarization\textgreater{} \\
David Beckham played for Real Madrid from 2003 to 2007 and for LA Galaxy from July 21, 2007. \\
\textless{}/Summarization\textgreater{} \\
\\
\textless{}Document\textgreater{} \\
Houston Rockets \textbar{} The Houston Rockets have won the NBA championship twice in their history. Their first win came in 1994, when they defeated the New York Knicks in a seven-game series. The following year, in 1995, they claimed their second title by sweeping the Orlando Magic. Despite several playoff appearances in the 2000s and 2010s, the Rockets have not reached the NBA Finals since their last championship victory in 1995. \\
\textless{}/Document\textgreater{} \\

\textless{}Question\textgreater{} \\
When did the Houston Rockets win the NBA championship? \\
\textless{}/Question\textgreater{} \\

\textless{}Summarization\textgreater{} \\
The Houston Rockets won the NBA championship twice in 1994 and 1995. \\
\textless{}/Summarization\textgreater{} \\

\\
Now your document and question are: \\

\textless{}Document\textgreater{} \\
\{document\} \\
\textless{}/Document\textgreater{} \\

\textless{}Question\textgreater{} \\
\{normalized question\}? \\
\textless{}/Question\textgreater{} \\

\textless{}Summarization\textgreater{} \\
\bottomrule
\end{tabular}
\captionsetup{justification=justified,singlelinecheck=false}
\caption{Detailed prompts for Independent Reading.}
\label{indipendent_reader_prompt}
\end{table}

\begin{table}[h!]
\centering
\begin{tabular}{>{\raggedright\arraybackslash}m{0.94\linewidth}} 
\toprule
\textbf{Combined Reading Prompting} \\
As an assistant, your task is to answer the question based on the given knowledge. Answer the given question; you can refer to the document provided. Your answer should follow the \textless{}Answer\textgreater{} tag. The given knowledge will be after the \textless{}Context\textgreater{} tag. You can refer to the knowledge to answer the question. Answer only the name for 'Who' questions. If the knowledge does not contain the answer, answer the question directly. \\[1em]
\\
There are some examples for you to refer to: \\

\textless{}Context\textgreater{} \\
In 1977, Trump married Czech model Ivana Zelníčková. The couple divorced in 1990, following his affair with actress Marla Maples.
\\\\
Trump and Maples married in 1993 and divorced in 1999. 
\\\\
In 2005, Donald Trump married Slovenian model Melania Knauss. They have one son, Barron (born 2006). \\
\textless{}/Context\textgreater{} \\

\textless{}Question\textgreater{} \\
Who was the spouse of Donald Trump between 2010 and 2014? \\
\textless{}/Question\textgreater{} \\

\textless{}Thought\textgreater{} \\
According to the context, Donald Trump married Melania Knauss in 2005. The period between 2010 and 2014 is after 2005. Therefore, the answer is Melania Knauss. \\
\textless{}/Thought\textgreater{} \\

\textless{}Answer\textgreater{} \\
Melania Knauss \\
\textless{}/Answer\textgreater{} \\

\\
Now your question and context knowledge are: \\

\textless{}Context\textgreater{} \\
\{generations\} \\
\textless{}/Context\textgreater{} \\

\textless{}Question\textgreater{} \\
\{question\} \\
\textless{}/Question\textgreater{} \\

\textless{}Thought\textgreater{} \\
\toprule
\end{tabular}
\captionsetup{justification=justified,singlelinecheck=false}
\caption{Detailed prompts for Combined Reading.}
\label{combined_reader_prompt}
\end{table}

\end{document}